%% file: arxiv_v2_11-17-26.tex
\documentclass{article} 
\usepackage{iclr2027_conference,times}

\input{math_commands.tex}

\usepackage{hyperref}
\usepackage{url}

\usepackage{graphicx}
\usepackage{booktabs}
\usepackage{subcaption}
\usepackage{float}
\usepackage{placeins}
\usepackage{flafter}
\usepackage{mathtools}
\usepackage{multirow}
\usepackage{standalone}

\usepackage{amsmath}
\usepackage{pgfplots}
\pgfplotsset{compat=1.18}
\usetikzlibrary{arrows.meta}

\usepackage{algorithm}
\usepackage{algpseudocode}
\usepackage{amssymb}

\definecolor{barblue}{RGB}{31,119,180}
\definecolor{accorange}{RGB}{224,123,57}
\definecolor{gridgray}{RGB}{223,223,223}
\definecolor{framegray}{RGB}{178,178,178}
\definecolor{bluebg}{RGB}{242,246,250}
\definecolor{orangebg}{RGB}{253,247,243}

\pgfplotsset{
  panel/.style={
    scale only axis,
    anchor=south west,
    height=7.6cm,
    axis line style={framegray, line width=0.6pt},
    tick style={draw=none},
    grid=major,
    grid style={gridgray, line width=0.5pt},
    xmin=-0.8, xmax=15.8,
    xtick={0,1,2,3,4,5,6,7,8,9,10,11,12,13,14,15},
    xlabel={Token index},
    title style={yshift=4pt, font=\bfseries},
    label style={font=\normalsize},
    tick label style={font=\normalsize},
  },
  barstyle/.style={ybar, bar width=0.42cm, fill=barblue, draw=barblue},
}

\tikzset{
  keybox/.style={
    draw, line width=1.1pt, rounded corners=6pt,
    inner sep=6pt, align=center, text width=7.45cm,
  },
}

\usepackage{xcolor}
\usepackage{xurl}
\usepackage{array,tabularx,longtable}
\hypersetup{hidelinks}
\title{SANTA++: Sampling Attention through Representative Keys}
\author{%
\textbf{Kyle Lee \quad Christian Z. Pratt \quad
Ruoyu Fang \quad Heekyung Lee}\\[0.35em]
\textbf{Avinash Lohitsa \quad Ryan Modafe \quad
Kerem Y. Camsari}\\[0.65em]
Department of Electrical and Computer Engineering\\
University of California, Santa Barbara, CA, USA
}
\iclrfinalcopy
\begin{document}
\maketitle
\lhead{}

\begin{abstract}
Attention often concentrates on a small subset of tokens in the context, but which subset matters changes from one query to the next. To exploit this changing structure, we introduce SANTA++, a training-free stochastic attention method that uses representative keys for memory-efficient selection without scanning the entire key-value (KV) cache. Cached keys are organized into teams, and the query scores one representative from each team to decide which teams to sample. We compute exact attention scores within the sampled teams and reweight each team's contribution by the inverse of its inclusion probability. This importance sampling correction estimates attention over the full cache, with a sampling budget that lets us trade memory reads for accuracy. {With Qwen2.5-7B-Instruct at 32K context, contiguous SANTA++ at $S=1024$ retains 95--98\% of dense aggregate scores on LongBench v2, HELMET RAG, and RULER using 28--31\% logical KV access.} {On RULER, SANTA++ outperforms Quest across the tested accuracy--access sweep, using 48.7\% fewer KV reads at approximately 79\% accuracy.} {It also outperforms MagicPIG at low access.} {At $S=1024$, our Triton implementation gives a $1.38\times$
attention-operator speedup over the Flash backend of PyTorch scaled
dot-product attention (Flash SDPA) in FP16. At batch eight,
full-model decoding is $1.23\times$ faster than dense Flash SDPA
with matched INT4 model weights and FP16 KV-cache precision
on an RTX 5090 Laptop GPU.} {Our kernels are available at \url{https://github.com/OPUSLab/santapp-kernel-demo}.}
\end{abstract}

\section{Introduction}
\label{sec:introduction}

Which cached tokens make the largest contributions to attention depends on the current decoding query.
At long context lengths, attention can concentrate on a small subset of cached tokens~\citep{zhang2023h2o,tang2024quest}, but identifying that subset from exact attention scores still requires reading every key.
Selecting fewer tokens then saves value reads while leaving a full-key
scan. When decoding is limited by memory bandwidth, these key-value
(KV) cache reads are costly, and their cost grows with context length~\citep{ribar2024sparq}. {For example, with Qwen2.5-7B-Instruct at a 32K-token context, streaming FP16 keys and values amounts to about 64 MiB per layer for each generated token.} To reduce both selection and evaluation reads, a sparse method must
find useful entries without first scoring every key.

\begin{figure}[!htbp]
    \centering
    \includegraphics[width=1.0\linewidth]{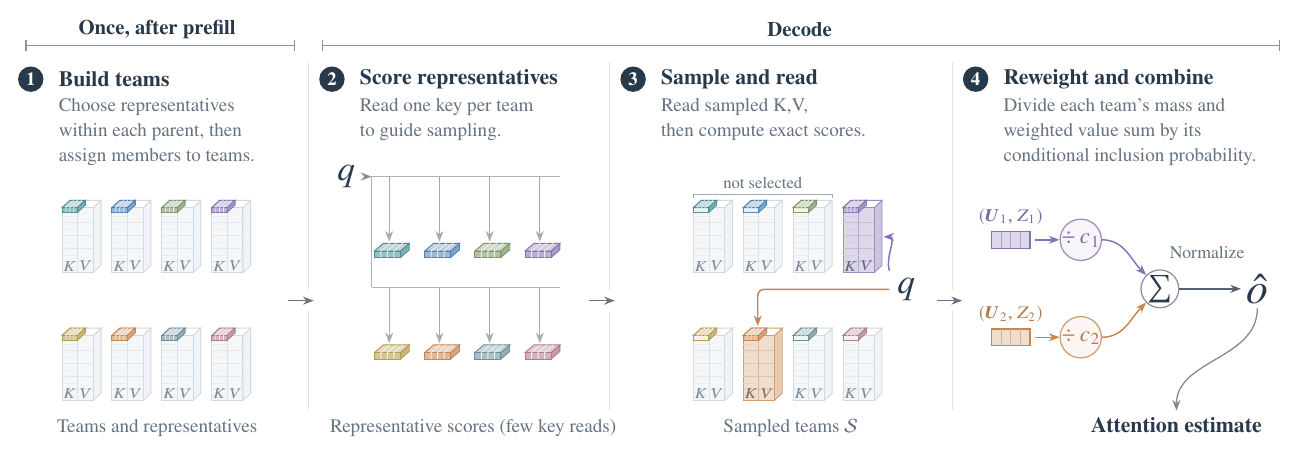}
    \caption{\textbf{Representative-guided attention sampling.}
After prefill, parent groups cover the prompt, and actual member keys
represent the teams built within each parent. For each decoding query, we form a routing weight for each team from its representative score and team size, and use these weights to sample teams $\mathcal S$. Selected teams' keys and values are
read in full, and their members' exact attention contributions are
computed. Each team's unnormalized attention mass and weighted value
sum are divided by its conditional inclusion probability $c_g$ before
the corrected sums are combined and normalized
(Section~\ref{sec:method}).}
    \label{fig:santapp algo}
\end{figure}

We introduce SANTA\texttt{++}, a training-free attention method that
routes queries through actual-key representatives
(Figure~\ref{fig:santapp algo}). After dense prefill, we divide prompt
keys into parent groups, choose representative keys within each parent,
and assign the remaining keys to their nearest representative. {Each team contains its selected representative, called its leader.} For each decoding query, we
use each representative score and team size to estimate the team's
attention mass, then sample teams according to these estimates. We then
read the selected teams' keys and values and evaluate every selected
member's exact query--key score. Newly generated tokens form a separate
suffix evaluated exactly.

{The prior method SANTA~\citep{lee2026santa} scores every cached
key and samples from the resulting exact attention distribution.
SANTA\texttt{++} replaces this full-key scan with approximate team-level
routing. When every key forms its own team, its routing weights reduce
to SANTA's exact attention weights.}
{SANTA\texttt{++} combines distance-based teams, inexpensive representative routing,
and corrected whole-team sampling to reduce both selection reads and
attention reads without retraining.}

{We use several actual keys to represent different parts of each parent
rather than assigning the entire parent a single mean-key score.}
{This lets the query distinguish different parts of a parent before
reading their members.} {Section~\ref{sec:representative-diagnostic}
compares leaders, centroids, and random members on the same teams.}
Representative scores guide selection, while exact member scores
determine the selected teams' contributions. Because
teams have unequal chances of selection, we divide each selected team's
unnormalized attention mass and weighted value sum by its conditional
inclusion probability before normalization (Section~\ref{sec:method}). This correction targets
full-cache contributions rather than simply renormalizing an
uncorrected selected subset.

The nominal budget $S$ controls how many teams are selected, providing
an adjustable accuracy--KV-access trade-off. Actual access also
depends on team sizes, routing and suffix reads, and overlap across
query heads. We form parent groups either with k-means or from contiguous token
spans. K-means groups keys by similarity. Contiguous spans instead
define parents directly from token positions, avoiding the k-means
preparation step. Both use the same distance-based team construction,
and every prompt team remains eligible for each query. For the Qwen2.5-7B-Instruct evaluations below, we fix the nominal parent
size to $P=16$ and use up to $R=4$ representatives per parent. {These
values were chosen heuristically and kept fixed across the main benchmark
sweeps.} {We also vary $P$ and $R$ in a separate sensitivity experiment
(Appendix~\ref{app:pr-sensitivity}).}

Our evaluation with Qwen2.5-7B-Instruct at 32K context shows that
SANTA\texttt{++} can retain most of dense attention's accuracy while
reading much less of the KV cache. {We use contiguous grouping at
$S=1024$ as a representative operating point: it retains 97.99\% of the
dense score on LongBench v2~\citep{bai2025longbenchv2}, 97.56\% on
HELMET's retrieval-augmented generation subset~\citep{yen2025helmet},
and 95.07\% on RULER~\citep{hsieh2024ruler}, at 28--31\% logical KV
access.} {Lower budgets offer further savings, whereas difficult retrieval
tasks benefit from more samples.}

{SANTA\texttt{++} improves the aggregate accuracy--access trade-off over
Quest across the tested RULER sweep.} {At the representative setting,
it reaches 79.40\% accuracy with 28.16\% access, compared with Quest's
78.89\% at 54.88\% access---48.7\% fewer KV reads.} {Against MagicPIG,
SANTA\texttt{++} has a clear low-access advantage and similar aggregate
accuracy around this setting.} {Section~\ref{sec:selection-baselines}
compares routing costs and auxiliary storage as well as task scores.}

{Our Triton implementation gives a $1.38\times$ attention-operator
speedup over Flash SDPA at $S=1024$, and $1.80\times$ at $S=128$.} {Integrating the method into all
28 model layers gives a $1.23\times$ decoding speedup with INT4 model
weights, FP16 KV caches, and batch size eight.} {Section~\ref{sec:kernels} reports full-request latency, preparation
costs, and memory use alongside decoding speed.}
{Appendix~\ref{app:mla} tests whether useful token sparsity remains
in DeepSeek-V2-Lite-Chat after latent compression and head sharing.}

\section{Method}
\label{sec:method}
{SANTA\texttt{++} approximates full-cache attention by routing through one representative key per team and evaluating all members of the sampled teams.}

For one decoding query $\boldsymbol q\in\mathbb R^d$, let $\mathcal C$
index the currently cached key-value pairs $(\boldsymbol k_i,\boldsymbol v_i)$.
The keys and query include the model's positional embedding. The
full-cache attention output is
\begin{equation}
\begin{aligned}
 s_i &= \frac{\boldsymbol q^{\mathsf T}\boldsymbol k_i}{\sqrt d},
 & Z &= \sum_{i\in\mathcal C} e^{s_i},
 & \boldsymbol U &= \sum_{i\in\mathcal C} e^{s_i}\boldsymbol v_i,
 & \boldsymbol o &= \frac{\boldsymbol U}{Z}.
\end{aligned}
\label{eq:method-attention}
\end{equation}
Here $Z$ is the full-cache unnormalized attention mass, and
$\boldsymbol U$ is its value-weighted numerator. SANTA\texttt{++}
approximates these two sums by sampling teams and evaluating all members
of the selected teams (Figure~\ref{fig:santapp algo}). Representatives
guide which teams to read, and they do not replace the selected members'
keys or values in the attention calculation.

\subsection{Approximating group mass}
\label{sec:mean keys underestimate}

A single mean key is a convenient summary for routing, but it gives a
group's mean logit rather than its attention mass. For a group
$\mathcal G$ of $n$ keys, write
$\bar{\boldsymbol k}=n^{-1}\sum_{i\in\mathcal G}\boldsymbol k_i$ and
$\bar s=\boldsymbol q^{\mathsf T}\bar{\boldsymbol k}/\sqrt d$.
Convexity of the exponential gives
\begin{equation}
 \underbrace{n e^{\bar s}}_{\text{centroid mass surrogate}}
 \;\leq\;
 \underbrace{\sum_{i\in\mathcal G}e^{s_i}}_{\text{exact group mass}}.
\label{eq:method-jensen}
\end{equation}
{A mean key can underestimate a parent's attention mass when a few keys
score highly. This error varies across parents and can distort routing
weights. We therefore route through smaller teams.
Section~\ref{sec:representative-diagnostic} compares leaders and centroids;
Appendix~\ref{app:jensen inequality} illustrates the mass approximation.}

\subsection{Preparing parent groups and teams}

SANTA\texttt{++} divides each parent group into teams represented by
actual keys. This lets a query choose among different parts of the
parent instead of assigning all its members one mean-key routing weight.

Parent groups restrict the distance comparisons needed to build teams.
After dense prefill, we form them separately for each layer and KV head
from the $N$ prompt keys after rotary position encoding (RoPE)~\citep{su2024rope}, using
either k-means on the keys or consecutive spans of $P$ tokens. Here
$P$ is a nominal parent size. K-means parent sizes can vary. Contiguous
parents avoid clustering while still covering the full prompt. Both
alternatives use the same team construction, in which each member is
compared only with representatives from its own parent. Our distance diagnostic finds that contiguous groups are more compact
than random groups, although less compact than k-means groups
(Appendix~\ref{app:centroid-distance}).

Within each parent, we choose up to $R$ actual keys as representatives.
The first is nearest the parent's mean key. Each later choice is
farthest from its nearest existing representative, to cover members
poorly represented so far. The mean thus helps choose the first actual
key rather than serving as the routing key itself. Every member joins
its nearest representative's team, and each representative belongs to
its own team. Distances are squared Euclidean distances on the
unnormalized keys. Contiguous parents need not yield contiguous teams,
because assignment uses key distance rather than token order.

The resulting $M$ nonempty teams $\mathcal T_g$ partition all prompt
rows. We store each team's representative $\boldsymbol\ell_g$, size
$n_g$, and member indices. Parent formation and team construction
are performed once after prefill, and the resulting records are
reused throughout decoding. Their preparation cost is therefore
spread over the generated tokens. Every prompt team remains eligible
at each query. Newly generated rows form a separate suffix
$\mathcal E$ evaluated exactly. Appendices~\ref{app:prefill-window} and~\ref{app:key-fingerprint}
give additional diagnostics on contiguous windows and key-versus-probe
clustering, respectively.

\subsection{Sampling and correcting team contributions}

The routing weight should reflect both query relevance and team size.
We compute it using one representative key per team:
\begin{equation}
 \widetilde Z_g=n_g\exp\!\left(
 \frac{\boldsymbol q^{\mathsf T}\boldsymbol\ell_g}{\sqrt d}\right),
 \qquad \phi_g=\log\widetilde Z_g.
\label{eq:method-routing}
\end{equation}
The factor $n_g$ accounts for how many members the representative
describes. We add independent Gumbel noise to $\phi_g$ and select the $K$ teams
with the largest perturbed routing scores using the Gumbel-top-$K$
trick~\citep{kool2019stochastic}, forming a set $\mathcal S$ sampled
globally without replacement.
{The budget $S$ is expressed in nominal prompt-token units:
using nominal team size $P/R$, we select $K=\min\{M,S/(P/R)\}$
teams per query head. With $P=16,R=4$, $S=128$ requests 32 teams
and $S=1024$ requests 256. Actual selected rows total
$\sum_{g\in\mathcal S}n_g$ because team sizes vary; routing and
exact-suffix reads are additional. Appendix~\ref{app:algorithm} gives
the integer-budget rules and sampler.}

After selection, we read the keys and values of every member of each
selected team and evaluate its exact contribution:
$Z_g=\sum_{i\in\mathcal T_g}e^{s_i}$ and
$\boldsymbol U_g=\sum_{i\in\mathcal T_g}e^{s_i}\boldsymbol v_i$.
Thus an approximate representative score decides what to read.

{Without correction, a team's expected contribution depends on its
selection probability. We therefore divide its mass and weighted value
sum by the same factor before normalization. For $K<M$, let $\tau$ be
the $(K+1)$st-largest Gumbel-perturbed routing score. Each selected team
uses $c_g=1-\exp[-\exp(\phi_g-\tau)]$, its inclusion probability
conditional on the other teams' perturbed scores. Appendix~\ref{app:algorithm}
derives this correction for the fixed query, cache, team records, and budget.}
Defining $Z_{\mathcal E}$ and $\boldsymbol U_{\mathcal E}$
by the same sums over the exact suffix, we use
\begin{equation}
\begin{aligned}
 \widehat{\boldsymbol U}
 &=\boldsymbol U_{\mathcal E}
   +\sum_{g\in\mathcal S}\frac{\boldsymbol U_g}{c_g},
 & \widehat Z
 &=Z_{\mathcal E}+\sum_{g\in\mathcal S}\frac{Z_g}{c_g},
 & \widehat{\boldsymbol o}
 &=\frac{\widehat{\boldsymbol U}}{\widehat Z}.
\end{aligned}
\label{eq:method-estimate}
\end{equation}
All members receive the same correction because the team is the
sampling unit. The two unnormalized sums are
unbiased (Equation~\ref{eq:app-unbiased-sums} in
Appendix~\ref{app:algorithm}). Their ratio, which gives the attention
output, is generally biased. Selecting all teams, with $c_g=1$,
recovers dense attention.

For grouped-query attention (GQA), query heads associated with the same
KV head use the same team partition but make independent selections and
retain their own inclusion corrections. Keys and values are read for
the union of those selections, while each team contributes only to the
heads that selected it.

\subsection{{Controlled comparison of routing representatives}}
\label{sec:representative-diagnostic}

Inclusion correction accounts for unequal selection chances, but
routing still determines which teams are likely to be read at a
limited budget. We therefore use Kullback--Leibler (KL) divergence
to compare how closely each routing distribution follows dense
attention over the same prompt tokens. We spread each group's
routing probability equally across its members and compare the
resulting token distribution with dense attention over the prompt.

{We compare leaders, team centroids, and randomly chosen members
on identical teams. We also compare flat centroid-guided baselines:
k-means groups of nominal size $B=4$ or $B=16$, each represented by
its mean key without further subdivision into teams.}
{All variants use the same dense SDPA queries and cache states from
Qwen2.5-7B-Instruct at 8K, with 20 prompts each from CWE, S-NIAH-1,
and QA-1.} {Task abbreviations are defined in Appendix~\ref{app:ruler}.}
No sampling is performed in this diagnostic.

{Table~\ref{tab:attention-kl} shows that both leaders and team centroids
have lower KL than the flat-centroid baselines.} {On identical teams,
centroids have lower KL than leaders on all three tasks, while random
members have higher KL.} {The improvement over flat centroids therefore comes partly from
the team partition, while the representative controls routing quality
within that partition.}

\begin{table}[!htbp]
\centering
\small
\setlength{\tabcolsep}{5pt}
\begin{tabular}{lrrrrr}
\toprule
& \multicolumn{3}{c}{Same teams, $P=16,R=4$} & \multicolumn{2}{c}{Flat centroids}\\
\cmidrule(lr){2-4}\cmidrule(l){5-6}
Task & Leader & {Centroid} & {Random} & $B=4$ & $B=16$ \\
\midrule
CWE      & 1.621 & {1.467} & {1.891} & 2.806 & 3.429 \\
S-NIAH-1 & 1.248 & {1.107} & {1.498} & 2.076 & 2.827 \\
QA-1     & 1.613 & {1.462} & {1.912} & 2.429 & 3.495 \\
\bottomrule
\end{tabular}
\caption{ \textbf{Routing approximation to dense attention.}
Mean $\mathrm{KL}(p\|q)$ in nats on shared dense states. The three team
variants use identical memberships with k-means parents.
Appendix~\ref{app:kl-variants} gives the distributions and averaging.}
\label{tab:attention-kl}
\end{table}

{Both selected leaders and team centroids provide effective routing,
whereas random representatives reduce downstream accuracy.} {In a separate
13-task RULER-8K evaluation with 100 prompts per task, their scores at
$S=128$ are 84.16, 83.02, and 77.28, respectively, at 24.97\%, 24.02\%,
and 24.56\% access.} {These results support the team-based routing design
and show that representative quality matters, while lower routing KL alone
does not determine task accuracy.} {Leaders remain our default because they are already available from
team construction and provide competitive downstream accuracy.
Appendix~\ref{app:kl-variants} gives the complete three-budget comparison.}

\section{{Accuracy and logical KV access}}
\label{sec:accuracy-kv}

SANTA\texttt{++} reduces logical KV access below SANTA's
full-key-scan cost at useful operating points across three benchmarks.
{We compare dense scaled dot-product attention (SDPA), SANTA~\citep{lee2026santa},
which samples from exact attention weights after a full-key scan, and
SANTA\texttt{++} with k-means or contiguous grouping using
Qwen2.5-7B-Instruct~\citep{qwen2024qwen25}.} {The main results use a
32K context budget and cover HELMET retrieval-augmented generation
(RAG)~\citep{yen2025helmet}, LongBench v2~\citep{bai2025longbenchv2},
and RULER~\citep{hsieh2024ruler}.} {Table~\ref{tab:representative-quality}
summarizes contiguous $S=1024$ across these benchmarks.} {The full budget
sweeps and task-level results are reported in Appendices~\ref{app:helmet},
\ref{app:longbench}, and~\ref{app:ruler}.}

{KV access counts key and value vectors read relative to dense
decoding, including routing keys. Shared reads count once per KV head.
Keys account for half of dense key-plus-value reads, so SANTA's full-key
scan alone costs 50\% of dense KV access.}
Appendix~\ref{app:evaluation-setup} gives the accounting details.

\begin{table}[!htbp]
\centering

\small
\setlength{\tabcolsep}{5pt}
\caption{ \textbf{Representative contiguous configuration at $S=1024$.}
Qwen2.5-7B-Instruct with a 32K context budget. Scores use each
benchmark's metric; $\pm$ denotes a 95\% bootstrap interval half-width.
Retained score is relative to dense SDPA. Access includes routing.}
\label{tab:representative-quality}
\begin{tabular}{@{}lrrrr@{}}
\toprule
Benchmark & Dense score (\%) & SANTA++ score (\%) & Retained (\%) & Access (\%)\\
\midrule
LongBench v2 & $32.11\pm3.98$ & $31.46\pm3.65$ & 97.99 & 29.52\\
HELMET RAG & $57.40\pm2.00$ & $56.00\pm2.07$ & 97.56 & 31.22\\
RULER & $83.51\pm0.66$ & $79.40\pm0.73$ & 95.07 & 28.16\\
\bottomrule
\end{tabular}
\end{table}

\paragraph{HELMET RAG.}
Contiguous grouping omits parent k-means while retaining the same
team construction within each parent. At $S=1024$, it retains 97.56\%
of the SDPA score at {31.22\%} access. Its absolute score is 56.00\%,
compared with 56.25\% for k-means: a difference of 0.25 percentage
points with 7.26 percentage points less access, while also avoiding
the parent clustering step.

{Smaller budgets provide a lower-access option.} {K-means at $S=128$
retains 94.43\% of the dense RAG score at 17.49\% access, and contiguous
grouping retains 92.51\% at 15.66\% access.} {SANTA instead retains
99.13\% at 50.43\% access.} {The RAG summaries weight Natural Questions,
TriviaQA, PopQA, and HotpotQA equally.} {Other HELMET tasks are less
robust: at $S=128$, JSON-KV retrieval falls from 96.80\% for dense
SDPA to 45.60\% with k-means and 33.40\% with contiguous grouping
(Table~\ref{tab:helmet-32k-task}).} {We examine these failures below.}

\begin{table}[!htbp]
\centering
\small
\caption{ \textbf{HELMET RAG: full 32K budget sweep.}
Score is mean substring exact match on four RAG tasks. Normalized score
is relative to SDPA. KV is the equal-task mean access, including routing.
$\pm$ gives 95\% bootstrap interval half-widths.}
\label{tab:helmet-rag}
\label{tab:helmet-rag-32k}
\begin{tabular*}{\linewidth}{@{\extracolsep{\fill}}lrrrr@{}}
\toprule
Method & $S$ & KV (\%) & Score (norm) (\%) & Score (abs) (\%) \\
\midrule
SDPA & -- & 100.00 & 100.00 & 57.40 $\pm$ 2.00 \\
\midrule
\multirow{5}{*}{SANTA} & 128 & 50.43 & 99.13 & 56.90 $\pm$ 2.05 \\
 & 256 & 50.75 & 99.04 & 56.85 $\pm$ 2.05 \\
 & 512 & 51.30 & 100.17 & 57.50 $\pm$ 2.02 \\
 & 1024 & {52.16} & 99.48 & 57.10 $\pm$ 2.07 \\
 & 2048 & 53.53 & 99.65 & 57.20 $\pm$ 2.08 \\
\midrule
\multirow{5}{*}{\shortstack[l]{SANTA\texttt{++}\\(k-means)}} & 128 & 17.49 & 94.43 & 54.20 $\pm$ 2.05 \\
 & 256 & 21.82 & 95.56 & 54.85 $\pm$ 2.03 \\
 & 512 & 28.58 & 96.43 & 55.35 $\pm$ 2.05 \\
 & 1024 & 38.49 & 98.00 & 56.25 $\pm$ 2.02 \\
 & 2048 & 52.13 & 99.22 & 56.95 $\pm$ 2.07 \\
\midrule
\multirow{5}{*}{\shortstack[l]{SANTA\texttt{++}\\(contiguous)}} & 128 & 15.66 & 92.51 & 53.10 $\pm$ 2.05 \\
 & 256 & 18.49 & 94.77 & 54.40 $\pm$ 2.05 \\
 & 512 & {23.34} & 95.03 & 54.55 $\pm$ 2.07 \\
 & 1024 & {31.22} & 97.56 & 56.00 $\pm$ 2.07 \\
 & 2048 & 43.34 & 97.30 & 55.85 $\pm$ 2.05 \\
\bottomrule
\end{tabular*}
\end{table}

\paragraph{LongBench v2.}
{With a 32K context cap, SANTA\texttt{++} with k-means retains
98.45\% of the SDPA score at 16.68\% KV access
($S=128$; Table~\ref{tab:longbench_results}).} {Its absolute accuracy
is 31.61\%, compared with 32.11\% for SDPA.} This configuration uses
66.9\% less access than SANTA's lowest-access tested setting,
which uses 50.36\%. Near-dense accuracy is therefore possible well
below the access cost of a full-key scan.

{Contiguous grouping reaches 30.91\% accuracy at $S=128$ and 15.27\%
access, rising to 31.46\% at $S=1024$ and 29.52\% access.} {Unlike RULER,
LongBench shows little consistent benefit from increasing the budget.}
{Results average four runs on each of 503 prompts; evaluation and
confidence-interval details appear in Appendix~\ref{app:longbench}.}

\begin{table}[!htbp]
\centering

\small
\setlength{\tabcolsep}{4.5pt}
\caption{ \textbf{LongBench v2: full budget sweep with a 32K context cap.}
Qwen2.5-7B-Instruct on 503 prompts, with four runs per setting.
$\pm$ gives 95\% prompt-bootstrap interval half-widths. Normalized
scores are relative to SDPA; access includes routing.}
\label{tab:longbench_results}
\begin{tabular}{@{}llrrr@{}}
\toprule
Method & $S$ & Access (\%) & Normalized accuracy (\%) & Absolute accuracy (\%)\\
\midrule
SDPA  & ---  & 100.00  & {100.00} & {$32.11\pm3.98$}\\
\midrule
SANTA  & 128  & 50.36  & {$95.36\pm7.26$} & $30.62{\pm3.38}$\\
  & 256  & 50.59  & {$102.17\pm7.69$} & $32.80{\pm3.65}$\\
  & 512  & 50.97  & {$106.19\pm7.76$} & $34.10{\pm3.75}$\\
  & 1024  & 51.56  & {$99.23\pm6.18$} & $31.86{\pm3.78}$\\
\midrule
SANTA++ k-means  & 128  & 16.68  & {$98.45\pm8.80$} & $31.61{\pm3.60}$\\
  & 256  & 20.41  & {$94.74\pm8.55$} & $30.42{\pm3.63}$\\
  & 512  & 26.46  & {$101.08\pm7.81$} & $32.46{\pm3.43}$\\
  & 1024  & 35.62  & {$98.76\pm7.64$} & $31.71{\pm3.73}$\\
\midrule
SANTA++ contiguous  & 128  & 15.27  & {$96.28\pm8.76$} & $30.91{\pm3.50}$\\
  & 256  & 17.74  & {$96.75\pm8.21$} & $31.06{\pm3.53}$\\
  & 512  & 22.15  & {$97.83\pm8.81$} & $31.41{\pm3.68}$\\
  & 1024  & 29.52  & {$97.99\pm8.01$} & $31.46{\pm3.65}$\\
\bottomrule
\end{tabular}
\end{table}

\paragraph{RULER.}
At 28.16\% access ($S=1024$), contiguous grouping reaches 79.40\%
absolute score, retaining 95.07\% of the SDPA aggregate.
The simpler parent formation therefore remains useful below
30\% access, without requiring parent clustering.
{Increasing the budget improves the aggregate for both grouping
policies (Table~\ref{tab:ruler-budget-main}).} {K-means reaches 82.29\%
at $S=2048$ and 46.95\% access; contiguous grouping reaches 82.56\%
at $S=4096$ and 54.68\% access.} {At these larger budgets, the access
advantage over SANTA's full-key scan becomes small or disappears.}

The aggregate includes all 13 RULER tasks, with 500 completed prompts
per task. Table~\ref{tab:ruler_by_task_32k} shows that accuracy
retention varies across tasks. For contiguous grouping at $S=1024$,
the three S-NIAH tasks each score at least 98.80\%, but MK-NIAH-3
scores 60.20\%, compared with 89.60\% for SDPA. {These harder retrieval tasks benefit from larger budgets.}

\begin{table}[!htbp]
\centering

\small
\setlength{\tabcolsep}{8pt}
\caption{ \textbf{SANTA++ budget sweep on 32K RULER.}
Scores are the 13-task mean with 95\% bootstrap interval half-widths;
access includes routing. SDPA scores $83.51\pm0.66$ at 100\% access.
Bold marks the representative contiguous setting.}
\label{tab:ruler-budget-main}
\begin{tabular}{@{}rrrrr@{}}
\toprule
& \multicolumn{2}{c}{Contiguous parents} & \multicolumn{2}{c}{K-means parents} \\
\cmidrule(lr){2-3}\cmidrule(l){4-5}
Nominal $S$ & Score (\%) & Access (\%) & Score (\%) & Access (\%) \\
\midrule
128 & $64.72\pm0.77$ & 15.03 & $71.33\pm0.74$ & 16.38 \\
256 & $70.54\pm0.76$ & 17.31 & $76.05\pm0.74$ & 19.95 \\
512 & $75.37\pm0.75$ & 21.37 & $78.49\pm0.74$ & 25.70 \\
\textbf{1024} & $\mathbf{79.40\pm0.73}$ & \textbf{28.16} & $80.40\pm0.72$ & 34.44 \\
2048 & $81.42\pm0.72$ & 38.89 & $82.29\pm0.66$ & 46.95 \\
4096 & $82.56\pm0.67$ & 54.68 & $83.08\pm0.64$ & 63.65 \\
\bottomrule
\end{tabular}
\end{table}

\paragraph{{Comparison with selection baselines.}}
\label{sec:selection-baselines}
{We compare with Quest~\citep{tang2024quest} and MagicPIG~\citep{chen2025magicpig}
on all 13 RULER tasks at 32K, using 500 prompts per task and the same
Qwen2.5-7B-Instruct evaluation harness.} {Each baseline retains its own
dense layers and exact-attention policy.}

{\textbf{Quest.} SANTA++ outperforms Quest across the tested accuracy--access sweep.}
{Every evaluated Quest configuration has a higher-scoring,
lower-access SANTA++ point under both grouping policies
(Figure~\ref{fig:selection-tradeoff}; Table~\ref{tab:quest-tradeoff}).} {Around our representative
setting, SANTA++ scores 79.40\% at 28.16\% access, versus 68.70\% at
26.88\% for Quest $B=2048$.} {Quest $B=8192$ reaches a similar score
of 78.89\% at 54.88\% access; SANTA++ uses 48.7\% fewer KV reads.}

\begin{figure}[!htbp]
\centering
\begingroup
\color{black}
\begin{tikzpicture}
\begin{axis}[
  width=\linewidth, height=5.3cm,
  xmin=12, xmax=68, ymin=42, ymax=87,
  xlabel={Logical access (\% of dense)}, ylabel={RULER score (\%)},
  xtick={15,25,35,45,55,65}, ytick={45,55,65,75,85},
  tick label style={font=\small}, label style={font=\small},
  grid=major, grid style={black!12}, axis line style={black!60},
  tick style={black!60},
  legend style={at={(0.5,1.03)},anchor=south,legend columns=4,
    draw=none,fill=none,font=\small,column sep=4pt,
    /tikz/every even column/.append style={column sep=4pt}},
  legend cell align=left,
  clip=false,
]
\addplot+[color=barblue,solid,thick,mark=square*,mark size=2.2pt,
  mark options={solid},error bars/.cd,y dir=both,y explicit]
  table[x=x,y=y,y error=e,row sep=\\]{x y e\\
15.03 64.72 0.77\\
17.31 70.54 0.76\\
21.37 75.37 0.75\\
28.16 79.40 0.73\\
38.89 81.42 0.72\\
54.68 82.56 0.67\\
};
\addlegendentry{SANTA++ contiguous}
\addplot+[color=teal!80!black,dashdotted,thick,mark=diamond*,mark size=2.4pt,
  mark options={solid},error bars/.cd,y dir=both,y explicit]
  table[x=x,y=y,y error=e,row sep=\\]{x y e\\
16.38 71.33 0.74\\
19.95 76.05 0.74\\
25.70 78.49 0.74\\
34.44 80.40 0.72\\
46.95 82.29 0.66\\
63.65 83.08 0.64\\
};
\addlegendentry{SANTA++ k-means}
\addplot+[color=accorange,dashed,thick,mark=triangle*,mark size=2.5pt,
  mark options={solid},error bars/.cd,y dir=both,y explicit]
  table[x=x,y=y,y error=e,row sep=\\]{x y e\\
16.80 59.84 0.80\\
20.47 65.46 0.79\\
26.88 68.70 0.78\\
37.64 73.65 0.77\\
54.88 78.89 0.76\\
};
\addlegendentry{Quest}
\addplot+[color=violet!80!black,only marks,mark=o,mark size=2.5pt,
  mark options={solid,line width=0.8pt},error bars/.cd,y dir=both,y explicit]
  table[x=x,y=y,y error=e,row sep=\\]{x y e\\
46.85 81.83 0.69\\
63.86 83.44 0.64\\
16.78 44.78 0.92\\
19.69 55.72 0.96\\
25.87 71.39 0.89\\
34.24 80.39 0.73\\
46.91 83.11 0.65\\
63.98 83.43 0.65\\
16.33 54.12 0.95\\
19.88 69.31 0.91\\
25.77 78.91 0.76\\
34.38 82.27 0.69\\
};
\addlegendentry{MagicPIG}
\addplot[black!65,densely dotted,thin,forget plot] coordinates {(12,83.51) (68,83.51)};
\node[anchor=south west,font=\footnotesize,fill=white,inner sep=1pt]
  at (axis cs:13,83.8) {Dense SDPA};
\addplot[barblue,only marks,mark=o,mark size=4pt,thick,forget plot]
  coordinates {(28.16,79.40)};
\node[anchor=north west,font=\small,fill=white,inner sep=1.3pt]
  at (axis cs:30.5,77.5) {$S=1024$};
\draw[barblue,thin,-{Latex[length=1.5mm]}]
  (axis cs:30.4,77.2) -- (axis cs:28.6,78.9);
\end{axis}
\end{tikzpicture}
\endgroup
\caption{ \textbf{Accuracy--access trade-offs on 32K RULER.}
Lines connect SANTA++ and Quest budgets; circles show all twelve MagicPIG
settings. Access includes routing for SANTA++/Quest and KV reads alone for
MagicPIG. Error bars show reported 95\% interval half-spans.}
\label{fig:selection-tradeoff}
\end{figure}

{\textbf{MagicPIG.} SANTA++ is stronger at low access.}
{At $S=128$, contiguous SANTA++ scores 64.72\% with 15.03\% access,
versus 54.12\% with 16.33\% KV access for MagicPIG
$(K_{\mathrm h},L)=(10,145)$.} {This advantage holds before accounting
for hash lookups.} {At $S=1024$, SANTA++ reaches 79.40\% versus
78.91\% for MagicPIG $(10,235)$ (Table~\ref{tab:selection-operating}).}
{Including estimated hash-lookup payload raises MagicPIG's access
from 25.77\% to 27.83\%, close to SANTA++'s 28.16\%.}

{Auxiliary storage also differs.} {MagicPIG $(10,235)$ has a modeled
CPU hash index averaging 3.11\,GiB (3.17\,GiB maximum).} {Our batch-one
systems measurements use a mean 0.245\,GiB of GPU representatives and
metadata, plus 7.0\,MiB of workspace (Table~\ref{tab:systems-steady}).}
{SANTA++ thus provides a competitive low-to-medium-access option with
compact representative routing rather than a hash index.}

\begin{table}[!htbp]
\centering

\small
\setlength{\tabcolsep}{4pt}
\caption{ \textbf{Comparison near the representative operating point.}
$A$ counts KV and routing-key reads for SANTA++/Quest, and KV reads for
MagicPIG. $A+H$ adds MagicPIG's estimated hash-lookup payload.
Scores show 95\% bootstrap interval half-widths.}
\label{tab:selection-operating}
\begin{tabular}{@{}llrrr@{}}
\toprule
Method & Setting & $A$ (\%) & MagicPIG $A+H$ (\%) & Score (\%) \\
\midrule
SDPA & --- & 100.00 & --- & $83.51\pm0.66$ \\
SANTA++ contiguous & $S=1024$ & 28.16 & --- & $79.40\pm0.73$ \\
SANTA++ $k$-means & $S=512$ & 25.70 & --- & $78.49\pm0.74$ \\
Quest & $B=2048$ & 26.88 & --- & $68.70\pm0.78$ \\
Quest & $B=8192$ & 54.88 & --- & $78.89\pm0.76$ \\
MagicPIG & $(10,235)$ & 25.77 & 27.83 & $78.91\pm0.76$ \\
\bottomrule
\end{tabular}
\end{table}

{The methods have different task strengths.} {SANTA++ is stronger on
S-NIAH-3 and MQ-NIAH at the representative comparison, whereas MagicPIG
is stronger on MK-NIAH-3 (78.80\% versus 60.20\%) and CWE
(71.60\% versus 58.00\%).} {At higher budgets, MagicPIG achieves better
aggregate accuracy--access trade-offs: $(10,315)$ scores 82.27\% at
37.20\% access including the estimated hash payload, versus 81.42\%
at 38.89\% for contiguous $S=2048$.} {Appendix~\ref{app:selection-baselines}
gives the baseline adaptations, storage model, and full task-level results.}

\paragraph{{Routing failures and design ablations.}}
\label{sec:ablations}
{At low budgets, routing can miss teams needed for an answer.} {On two
50-prompt diagnostic sets with k-means parents at 32K and $S=128$,
forcing answer-containing
teams raises JSON-KV accuracy from 46\% to 92\% and MK-NIAH-3
from 16\% to 66\%.} {An oracle that scans all keys and selects teams
by exact attention mass reaches 96\% and 94\%, respectively.}
{These controls identify routing as a promising target for improving
low-budget retrieval (Appendix~\ref{app:routing-failures}).}

{Removing inclusion correction has mixed task-level effects.} {With
k-means parents and $S=128$, corrected and uncorrected means are
85.21 and 83.97 at 8K, and 68.88 and 71.11 at 32K, over all 13
RULER tasks at each length (Appendix~\ref{app:inclusion-ablation}).}
{The correction estimates full-cache unnormalized sums;
its effect on task accuracy depends on the task.}
{Varying $P$ and $R$ on two 8K tasks shows that finer teams improve
scores while increasing routing reads (Appendix~\ref{app:pr-sensitivity}).}

\FloatBarrier

\section{{GPU implementation and generation latency}}
\label{sec:kernels}

{We evaluate contiguous SANTA++ with $P=16,R=4$ on Qwen2.5-7B-Instruct
using an NVIDIA RTX 5090 Laptop GPU with 24\,GB memory.}
{We compare against the Flash backend of PyTorch SDPA on RULER prompts
of approximately 32K tokens, using the same model weights, precision,
and prefill implementation.}

Under grouped-query attention (GQA)~\citep{ainslie2023gqa}, the implementation
reuses both representative-key loads and selected-member key and value
loads across query heads that share a KV head. Each head keeps its own selected teams and conditional inclusion
corrections, and accumulates contributions only from teams it selected.
Appendix~\ref{app:triton-kernels} describes the five kernel programs.

\paragraph{{Accuracy, access, and attention speed.}}
{The GPU implementation scores representatives, samples teams, forms
the GQA union, evaluates selected members, and combines partial attention
sums in five GPU kernel launches. Table~\ref{tab:kernel-e2e} reports
timings for all five together, including routing and selection, alongside
RULER accuracy and access at the same budgets.}
{Increasing the budget improves RULER accuracy while retaining an
attention speedup: at $S=1024$, contiguous SANTA++ retains 95.1\% of the
dense score with 28.2\% KV access and a $1.38\times$ faster attention
operator.} {The lower $S=128$ budget reaches $1.80\times$ speedup.}

\begin{table}[!htbp]
\centering

\small
\setlength{\tabcolsep}{4pt}
\caption{ \textbf{Accuracy, access, and attention speed at matching budgets.}
Contiguous SANTA++, Qwen2.5-7B-Instruct, FP16, 32K context.
RULER evaluation and complete attention-operator timings use matching
budgets; timings are at batch one. Scores and speedups show 95\% bootstrap
intervals.}
\label{tab:kernel-e2e}
\begin{tabular}{@{}rrrrrr@{}}
\toprule
$S$ & \shortstack{RULER score\\(\%)} & \shortstack{KV access\\(\%)} &
\shortstack{SANTA++\\($\mu$s)} & \shortstack{Flash\\($\mu$s)} &
\shortstack{Attention speedup\\{[95\% interval]}} \\
\midrule
128 & $64.72\pm0.77$ & 15.03 & 80.35 & 144.69 & $1.800\times$ [1.752, 1.844] \\
256 & $70.54\pm0.76$ & 17.31 & 92.52 & 145.05 & $1.568\times$ [1.528, 1.604] \\
512 & $75.37\pm0.75$ & 21.37 & 100.45 & 144.80 & $1.441\times$ [1.418, 1.461] \\
\textbf{1024} & $\mathbf{79.40\pm0.73}$ & \textbf{28.16} & \textbf{104.68} & 144.77 & $\mathbf{1.383\times}$ [1.362, 1.400] \\
\bottomrule
\end{tabular}
\end{table}

\paragraph{{Full-model generation.}}
{We integrate $S=1024$ into all 28 model layers and generate 512 outputs
per sequence, evaluating the growing generated suffix exactly.}
{Table~\ref{tab:systems-generation} reports full-model decoding
speedups separately from full requests, which include prefill and preparation.}
{We replicate prompts for batched decoding and prefill each sequence
separately.}
{FP16 at batch three gives $1.044\times$ decode and $1.012\times$
full-request speedup.}
{INT4 model weights enable batch eight within our 24\,GB GPU memory
limit, giving $1.232\times$ decode and $1.032\times$ full-request
speedup. Both dense Flash SDPA and SANTA++ use the same quantized
checkpoint.}
{Activations and KV caches remain FP16.} {Batching is important: INT4 at batch one is slower than dense,
while batches four and eight benefit from sparse attention.} 

\begin{table}[!htbp]
\centering

\small
\caption{ \textbf{Full-model generation at $S=1024$.}
Approximately 32K input tokens and 512 outputs per sequence.
Speedups include all model layers; 95\% bootstrap intervals describe
prompt variation. INT4 denotes model weights; KV caches remain FP16.
Full requests include prefill and preparation. See
Appendix~\ref{app:systems-timing-limits} for timing limitations.}
\label{tab:systems-generation}
\begin{tabular}{lrrr}
\toprule
Weights & Batch & Decode speedup & Full-request speedup \\
\midrule
FP16 & 1 & $1.016\times$ [1.009, 1.023] & $1.007\times$ [1.003, 1.011] \\
FP16 & 2 & $1.025\times$ [1.021, 1.028] & $1.010\times$ [1.008, 1.012] \\
FP16 & 3 & $1.044\times$ [1.040, 1.047] & $1.012\times$ [1.010, 1.014] \\
\midrule
INT4 & 1 & $0.808\times$ [0.796, 0.817] & $0.896\times$ [0.888, 0.901] \\
INT4 & 4 & $1.131\times$ [1.120, 1.141] & $1.028\times$ [1.025, 1.030] \\
INT4 & 8 & $1.232\times$ [1.217, 1.245] & $1.032\times$ [1.030, 1.034] \\
\bottomrule
\end{tabular}
\end{table}

\paragraph{{Preparation and memory.}}
{We store the full prompt KV cache once, reordered by team, with
about 0.25\,GiB of extra routing data and workspace per sequence in
these 32K runs. Peak allocated memory matches dense attention because
temporary allocations during the shared dense prefill set the peak
(Table~\ref{tab:preparation-memory-main}).}

{Sequential prefill accounts for 65\% of the sparse FP16 batch-three
request and 84\% at INT4 batch eight, limiting the full-request gain.
Shorter generations also provide less opportunity to amortize preparation
(Appendix~\ref{app:systems-short-generation}).}

\begin{table}[!htbp]
\centering

\small
\setlength{\tabcolsep}{5pt}
\caption{ \textbf{Preparation and memory at $S=1024$.}
Preparation includes team construction, packing, and workspace setup.
Peak allocation is the maximum over twelve prompts; extra steady
memory is the mean sparse-minus-dense allocation before decoding.}
\label{tab:preparation-memory-main}
\begin{tabular}{lrrrrr}
\toprule
Weights & Batch & \shortstack{Preparation\\(ms)} & \shortstack{Dense peak\\(GiB)} & \shortstack{Sparse peak\\(GiB)} & \shortstack{Extra steady\\(GiB)} \\
\midrule
FP16 & 3 & 118.0 & 22.134 & 22.134 & 0.755 \\
INT4 & 8 & 283.5 & 21.810 & 21.810 & 2.012 \\
\bottomrule
\end{tabular}
\end{table}

{Profiling all five kernels together at $S=1024$ measures
18.42\,MiB of DRAM reads and writes for SANTA\texttt{++}, versus
62.55\,MiB for Flash on the profiled CWE input.} {Further measurements and evaluation details
are in Appendix~\ref{app:kernel-evaluation}, with timing variability
discussed in Appendix~\ref{app:systems-timing-limits}.}
\FloatBarrier

\section{Related Work}
\label{sec:related-work}

\paragraph{Compressed representations.}
{Quantization~\citep{liu2024kivi,frantar2023optq,hooper2024kvquant}
and KV sharing or latent compression~\citep{ainslie2023gqa,deepseekai2024v2}
reduce model-weight or KV-cache storage. SANTA\texttt{++} instead
reduces the entries read per query. Combining selective reads with KV-cache quantization remains
future work.}

\paragraph{Sparse attention and cache retention.}
Sparse attention methods reduce computation by evaluating or approximating
only part of the full attention computation, using structured patterns
~\citep{child2019generating,zaheer2020bigbird,beltagy2020longformer},
low-rank projections~\citep{wang2020linformer},
hashing~\citep{kitaev2020reformer}, or deterministic top-$k$
selection~\citep{gupta2021topk}.
SANTA\texttt{++} instead treats sparse decode-time access as an
estimation problem: it uses query-dependent sampling and corrects the
sampled contributions toward the full-cache attention sums.
{Cache-retention methods such as H$_2$O, StreamingLLM, and
DuoAttention~\citep{zhang2023h2o,xiao2024streamingllm,xiao2025duoattention}
reduce memory use or attention access by retaining selected entries
or restricting attention to a recent window.}
SANTA\texttt{++} does not require permanently dropping prompt tokens,
so entries skipped for one query can still be selected later.
{This distinction matters when choosing a method: eviction can make a
context fit in memory, whereas SANTA\texttt{++} retains the full cache
and seeks to reduce the cost of reading it.}

\paragraph{Query-aware retrieval and grouping.}
Query-aware methods use approximate scores or group summaries to identify
promising parts of the cache~\citep{ribar2024sparq,tang2024quest,sun2025shadowkv,hao2025omnikv}.
InfLLM~\citep{xiao2024infllm}, for example, uses selected tokens to
represent contiguous memory blocks, while centroid-based methods use
group means for retrieval, merging, or approximation
~\citep{liu2025clusterkv,hooper2025squeezed,hu2026centroidkv,hooper2026multipole}.
{HISA~\citep{hisa2026} first filters pooled blocks, then applies a
token-level indexer within the retained blocks.} {SANTA\texttt{++}
builds on the same general idea of routing through small group
summaries, but samples whole teams and corrects their contributions
before normalization.} {Our controlled ablation separates representative
choice from group construction: team centroids have lower routing KL
than leaders, while downstream scores are close.} 
{A learned query-dependent team selector inspired by Fast
KVzip~\citep{kim2026fastkvzip} is another direction for future work,
at the cost of auxiliary training.}

\paragraph{Sampling and importance correction.}
{Prior work explores sampling-based attention approximations~\citep{kim2022fast}.
SANTA~\citep{lee2026santa} scores every cached key before sampling from
exact attention weights; SANTA\texttt{++} replaces that scan with
representative-guided team sampling. MagicPIG~\citep{chen2025magicpig}
uses locality-sensitive hashing for self-normalized importance sampling.
SANTA\texttt{++} instead samples explicit teams and corrects their
contributions using conditional inclusion probabilities.}

\section{Conclusion}
\label{sec:conclusion}

{SANTA\texttt{++} reduces KV reads without retraining by routing through
representative keys, evaluating sampled teams exactly, and correcting
their contributions.}

{At contiguous $S=1024$, it retains 95--98\% of dense aggregate scores
across LongBench v2, HELMET RAG, and RULER with 28--31\% logical KV access.
The complete attention operator is $1.38\times$ faster than Flash SDPA.
With matched INT4 weights, FP16 KV, and batch eight, full-model decoding
is $1.23\times$ faster; dense prefill limits complete-request gains.}

{On RULER, SANTA\texttt{++} improves on Quest across the tested
accuracy--access sweep and outperforms MagicPIG at low access; MagicPIG
is stronger at higher budgets. Improving routing for hard retrieval and
combining SANTA\texttt{++} with KV-cache quantization remain future
directions.}

\FloatBarrier
\label{page:main-end}
\clearpage
\section*{AI Use Statement}
We used generative AI tools to assist with writing and presentation, software implementation and debugging, methodological and mathematical development, and the analysis and interpretation of experimental results. All AI-assisted outputs were reviewed and validated by the authors.

\subsection*{Reproducibility statement}
{Our kernels are available at \url{https://github.com/OPUSLab/santapp-kernel-demo}.}

\bibliography{main-bib}
\bibliographystyle{iclr2027_conference}
\appendix

\section{Algorithm and estimator details}
\label{app:algorithm}
\label{app:santapp-algorithms}

We separate parent formation, team construction, and query-dependent
sampling. Algorithms~\ref{alg:santapp-kmeans-parents}
and~\ref{alg:santapp-contiguous-parents} provide alternative prompt
partitions. Both use the same team builder
(Algorithm~\ref{alg:santapp-build-teams}) and decode estimator
(Algorithm~\ref{alg:santapp-decode}).

\paragraph{Cache state and budget.}
For one layer and KV head, let $\boldsymbol k_i^{\mathrm{pre}}$ be the
key saved during dense prefill. Queries and keys are post-RoPE;
construction uses raw keys without normalization or feature
standardization. The $M$ nonempty teams $\mathcal T_g$ partition the
$N$ prompt indices, $n_g=|\mathcal T_g|$, and $\boldsymbol\ell_g$ is
an actual prefill key from team $g$. Memberships and representative
keys remain fixed after prefill. The current cache is the disjoint
union of the prompt teams and the exact generated suffix $\mathcal E$,
with only query-visible positions. Query heads sharing a KV head share
team records, but sample independently with their own thresholds and
corrections. Reading their union does not add another head's selections
to a head's estimator.

For positive integer parameters $P,R,S$, require $P$ to be divisible
by $R$ and $S$ to be divisible by $P/R$. The selected-team count is
\begin{equation}
 K=\min\!\left\{M,\frac{S}{P/R}\right\}.
\label{eq:method-budget}
\end{equation}
The number of selected prompt rows is
$\sum_{g\in\mathcal S}n_g$; this excludes representative-routing reads
and the exact suffix. After $t$ generated
tokens have been fed back, the suffix is
$\mathcal E=\{N+1,\ldots,N+t\}$. It is empty for the first sparse
prediction, which the reference obtains by recomputing the final prompt
token rather than using dense-prefill logits. No prompt window is
reserved for exact evaluation, including a short final contiguous parent.

\begin{algorithm}[H]
\caption{SANTA\texttt{++} attention for one query head}
\label{alg:santapp-decode}
\small
\begin{algorithmic}[1]
\Require Query $\boldsymbol q$; current KV cache; team records
  $(\mathcal T_g,\boldsymbol\ell_g)_{g=1}^{M}$; exact indices $\mathcal E$;
  requested integer team count $K\geq 1$.
\Ensure Estimated attention output $\widehat{\boldsymbol o}$.
\State $K\gets\min(K,M)$
\If{$K=M$}
    \State $\mathcal S\gets\{1,\ldots,M\}$ and $c_g\gets 1$ for every $g\in\mathcal S$
\Else
    \State $\phi_g\gets\log n_g+\boldsymbol q^{\mathsf T}\boldsymbol\ell_g/\sqrt d$
      for $g=1,\ldots,M$
    \State Draw independent $G_g\sim\operatorname{Gumbel}(0,1)$
      and set $Y_g\gets\phi_g+G_g$ for every $g$
    \State $\mathcal S\gets$ indices of the $K$ largest perturbed routing scores among
      $Y_1,\ldots,Y_M$
    \State $\tau\gets$ the $(K+1)$st-largest perturbed routing score among $Y_1,\ldots,Y_M$
    \State $c_g\gets 1-\exp\!\bigl[-\exp(\phi_g-\tau)\bigr]$
      for every $g\in\mathcal S$
\EndIf
\State $\mathcal I\gets\mathcal E\,\cup\!\displaystyle\bigcup_{g\in\mathcal S}\mathcal T_g$
\For{each $i\in\mathcal I$}
    \State Read $(\boldsymbol k_i,\boldsymbol v_i)$ and compute
      $s_i\gets\boldsymbol q^{\mathsf T}\boldsymbol k_i/\sqrt d$,
      $a_i\gets e^{s_i}$
\EndFor
\State $\displaystyle\widehat{\boldsymbol U}\gets
  \sum_{i\in\mathcal E}a_i\boldsymbol v_i+
  \sum_{g\in\mathcal S}\frac{1}{c_g}
  \sum_{i\in\mathcal T_g}a_i\boldsymbol v_i$
\State $\displaystyle\widehat Z\gets
  \sum_{i\in\mathcal E}a_i+
  \sum_{g\in\mathcal S}\frac{1}{c_g}\sum_{i\in\mathcal T_g}a_i$
\State \Return $\widehat{\boldsymbol o}=\widehat{\boldsymbol U}/\widehat Z$
\end{algorithmic}
\end{algorithm}

\paragraph{Conditional inclusion correction.}
We use the threshold-based inclusion correction for Gumbel-top-$K$
sampling~\citep{kool2019stochastic}. For $1 \le K < M$, condition on
the state $\mathcal D$ before sampling:
the current query, cache, team records, exact suffix, and budget.
Assume finite logits and values, positive routing weights, and fresh
independent standard Gumbel draws conditional on $\mathcal D$, with
cumulative distribution $\Pr(G_g\leq x)=\exp[-\exp(-x)]$.
The perturbed routing scores $Y_g=\phi_g+G_g$ then have no ties with
probability one. The ordered sampling law chooses among remaining teams
in proportion to $\widetilde Z_g$, so
$p_g=\widetilde Z_g/\sum_h\widetilde Z_h$ is the first-draw probability,
not generally the inclusion probability.

Let $Y_{-g}=(Y_h)_{h\ne g}$ and let $T_{-g}$ be the $K$th-largest
perturbed routing score among those other teams. Team $g$ is selected
exactly when $Y_g>T_{-g}$. Independence and the Gumbel cumulative
distribution give
\begin{align}
 \Pr(g\in\mathcal S\mid\mathcal D,Y_{-g})
 &=\rho_g(T_{-g}),
 \label{eq:app-conditional-inclusion}\\
 \rho_g(t)
 &=1-\exp\!\left[-\exp(\phi_g-t)\right].
 \label{eq:method-inclusion}
\end{align}
For a selected team $g$, omitting its perturbed routing score leaves
exactly $K-1$ scores above the global $(K+1)$st. Thus $T_{-g}$ equals
the observed $(K+1)$st-largest perturbed routing score $\tau$, and the
algorithm computes $c_g=\rho_g(\tau)$ for selected teams. This is why
selection retains one extra perturbed routing score, not one extra team.
The marginal inclusion probability is instead
$\pi_g=\mathbb E[\rho_g(T_{-g})\mid\mathcal D]$.
The correction uses the leave-one-out threshold because the global
threshold depends on $Y_g$.

\paragraph{Unnormalized sums.}
Let $X_g$ denote either $Z_g=\sum_{i\in\mathcal T_g}e^{s_i}$ or one
coordinate of
$\boldsymbol U_g=\sum_{i\in\mathcal T_g}e^{s_i}\boldsymbol v_i$.
These quantities are fixed given $\mathcal D$. Let
$\mathbf 1\{g\in\mathcal S\}$ denote the indicator that team $g$ is
selected. With exact arithmetic,
\begin{equation}
\mathbb E\!\left[
 \frac{\mathbf 1\{g\in\mathcal S\}X_g}{\rho_g(T_{-g})}
 \,\middle|\,\mathcal D,Y_{-g}\right]
=\frac{X_g\,\Pr(g\in\mathcal S\mid\mathcal D,Y_{-g})}
       {\rho_g(T_{-g})}
=X_g.
\label{eq:app-conditional-sum}
\end{equation}
For selected teams the denominator equals $c_g$; unselected teams
contribute zero. Summing and taking iterated expectations, with the
disjoint suffix added exactly, gives
\begin{equation}
\mathbb E[\widehat{\boldsymbol U}\mid\mathcal D]=\boldsymbol U,
\qquad
\mathbb E[\widehat Z\mid\mathcal D]=Z.
\label{eq:app-unbiased-sums}
\end{equation}
No factor of $1/K$ is needed. The two sums are unbiased for the current query and cache.
Their normalized ratio is generally biased. When $K=M$, all
corrections are one and the output equals dense attention on the current
cache in exact arithmetic.

\paragraph{Shared team construction.}
Build teams once per layer and KV head. The parent mean chooses only
the first actual-key representative; farthest-first choices then cover
keys poorly represented so far. Resolve representative-selection ties
by increasing token index and assignment ties by representative-selection
order. Forcing each representative to own its row keeps every team
nonempty, even when duplicate keys create ties.

\begin{algorithm}[H]
\caption{Build actual-key teams from a parent partition}
\label{alg:santapp-build-teams}
\small
\begin{algorithmic}[1]
\Require Prefill keys $(\boldsymbol k_i^{\mathrm{pre}})_{i=1}^{N}$;
  nonempty parents $\mathcal P$ partitioning $\{1,\ldots,N\}$;
  maximum representatives per parent $R\geq 1$.
\Ensure Team records $(\mathcal T_g,\boldsymbol\ell_g)_{g=1}^{M}$
  partitioning all prompt indices.
\State $\mathcal T\gets$ empty list
\For{each parent $C\in\mathcal P$}
    \State $r\gets\min(R,|C|)$ and
      $\displaystyle\boldsymbol\mu\gets\frac{1}{|C|}\sum_{i\in C}\boldsymbol k_i^{\mathrm{pre}}$
    \State $\displaystyle j_1\gets\arg\min_{i\in C}
      \|\boldsymbol k_i^{\mathrm{pre}}-\boldsymbol\mu\|_2^2$;
      $L\gets(j_1)$
    \For{$u=2,\ldots,r$}
        \State $\displaystyle j_u\gets\arg\max_{i\in C\setminus\{j_1,\ldots,j_{u-1}\}}
          \min_{j\in L}\|\boldsymbol k_i^{\mathrm{pre}}-
          \boldsymbol k_j^{\mathrm{pre}}\|_2^2$
        \State Append $j_u$ to $L$
    \EndFor
    \For{each $i\in C$}
        \State $\displaystyle b_i\gets\arg\min_{u\in\{1,\ldots,r\}}
          \|\boldsymbol k_i^{\mathrm{pre}}-
          \boldsymbol k_{j_u}^{\mathrm{pre}}\|_2^2$
    \EndFor
    \State $b_{j_u}\gets u$ for $u=1,\ldots,r$
      \Comment{Representative self-ownership}
    \For{$u=1,\ldots,r$}
        \State Append $\bigl(\{i\in C:b_i=u\},\boldsymbol k_{j_u}^{\mathrm{pre}}\bigr)$
          to $\mathcal T$
    \EndFor
\EndFor
\State \Return $\mathcal T$
\end{algorithmic}
\end{algorithm}

\begin{algorithm}[H]
\caption{Global minibatch k-means parent partition}
\label{alg:santapp-kmeans-parents}
\small
\begin{algorithmic}[1]
\Require Prefill keys $(\boldsymbol k_i^{\mathrm{pre}})_{i=1}^{N}$, $N\geq 2$;
  nominal parent size $P\geq 1$; minibatch k-means settings $\theta$ and seed $\xi$.
\Ensure Nonempty parents $\mathcal P$ partitioning all prompt indices.
\State $M_{\mathrm{parent}}\gets\min\!\bigl(N,\max(2,\lfloor N/P\rfloor)\bigr)$
\State $(\boldsymbol\mu_1,\ldots,\boldsymbol\mu_{M_{\mathrm{parent}}})\gets
  \Call{MiniBatchKMeans}{(\boldsymbol k_i^{\mathrm{pre}})_{i=1}^{N},M_{\mathrm{parent}};\theta,\xi}$
\State $\displaystyle b_i\gets\arg\min_{u\in\{1,\ldots,M_{\mathrm{parent}}\}}
  \|\boldsymbol k_i^{\mathrm{pre}}-\boldsymbol\mu_u\|_2^2$
  for every $i=1,\ldots,N$
\State \Return $\mathcal P=\bigl\{\{i:b_i=u\}:u=1,\ldots,M_{\mathrm{parent}},\;
  \{i:b_i=u\}\ne\varnothing\bigr\}$
\end{algorithmic}
\end{algorithm}

\begin{algorithm}[H]
\caption{Contiguous parent partition}
\label{alg:santapp-contiguous-parents}
\small
\begin{algorithmic}[1]
\Require Prompt length $N\geq 1$ and parent size $P\geq 1$.
\Ensure Nonempty parents $\mathcal P$ partitioning all prompt indices in token order.
\State $M_{\mathrm{parent}}\gets\lceil N/P\rceil$
\State $C_u\gets\{(u-1)P+1,\ldots,\min(uP,N)\}$
  for $u=1,\ldots,M_{\mathrm{parent}}$
\State \Return $\mathcal P=\{C_1,\ldots,C_{M_{\mathrm{parent}}}\}$
\end{algorithmic}
\end{algorithm}

\noindent
\textsc{MiniBatchKMeans} fits squared-Euclidean centers to raw
post-RoPE keys. It uses greedy k-means++ initialization, minibatches
sampled with replacement, and cumulative-count center updates. For the
reported runs, $\theta$ uses a minibatch size of 4096, one
initialization, at most 100 iterations, a 10-minibatch no-improvement
stopping rule, and a reassignment ratio of 0.01. The center-change
tolerance is zero and the initialization sample size follows the
implementation default. The random seed is $\xi=0$. Final assignments
break ties by center index. Empty parents are omitted. {The reference grouping records prompt indices.} {The integrated GPU implementation packs members by team and releases the original chronological prompt allocation after packing; the retained cache and overhead are measured in Appendix~\ref{app:kernel-evaluation}.}

\FloatBarrier

\section{Diagnostics and ablations}
\label{app:diagnostics}

\subsection{{Mean-key mass approximation}}
\label{app:jensen inequality}

\begin{figure}[p]
  \centering
  \begin{tikzpicture}
    \node[anchor=south west, inner sep=0] (img)
      {\includegraphics[width=\linewidth]{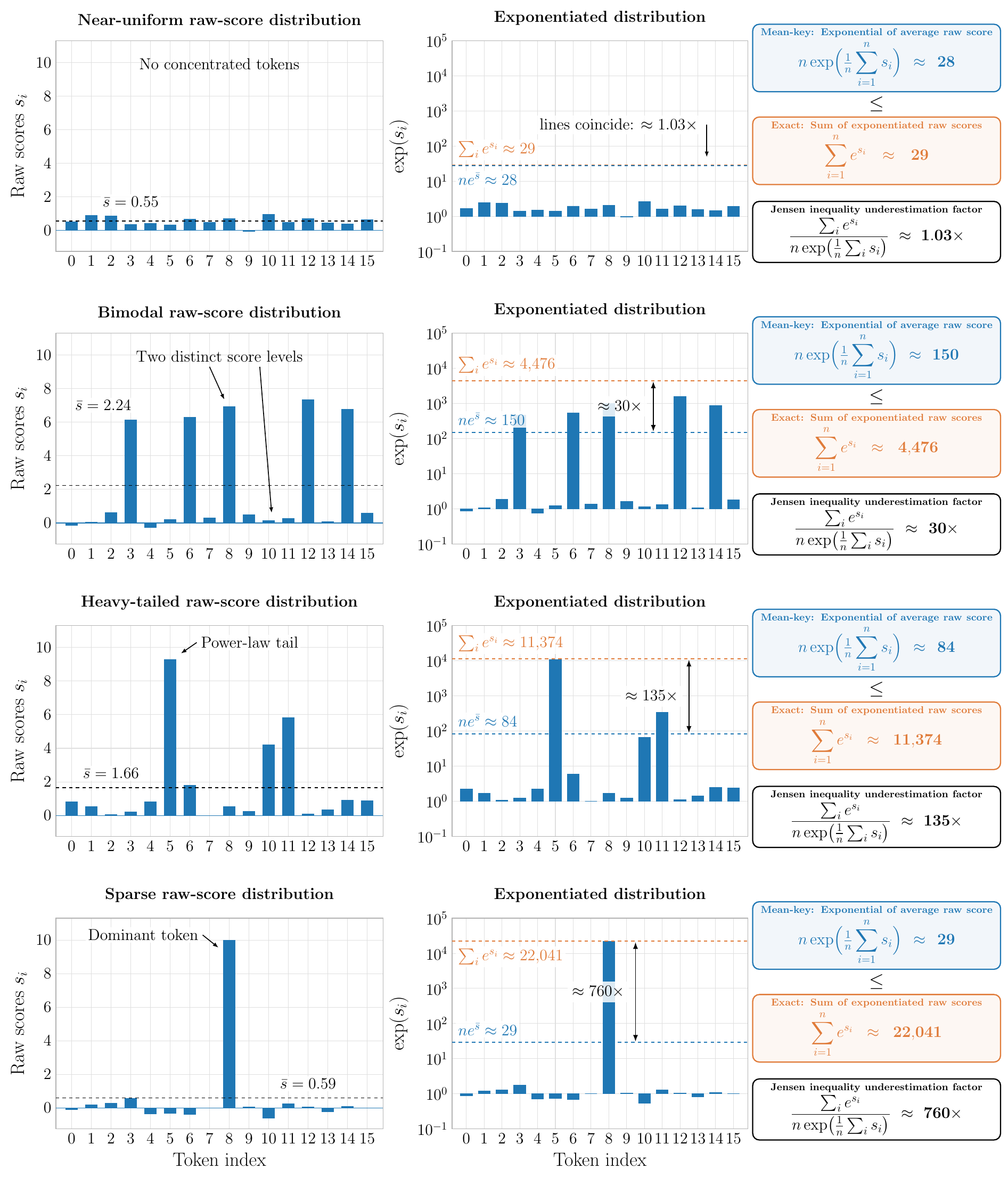}};
    \begin{scope}[x={(img.south east)}, y={(img.north west)}]
      \node[anchor=north west] at (0.00,1.00) {\textbf{(a)}};
      \node[anchor=north west] at (0.00,0.75) {\textbf{(b)}};
      \node[anchor=north west] at (0.00,0.50) {\textbf{(c)}};
      \node[anchor=north west] at (0.00,0.25) {\textbf{(d)}};
    \end{scope}
  \end{tikzpicture}
  \caption{\textbf{Additional Jensen examples.}
  Each row compares 16 raw logits $s_i$, their exponentials, and the
  exact-to-surrogate mass ratio
  $\sum_i e^{s_i}/\bigl[n\exp(n^{-1}\sum_i s_i)\bigr]$.
  The displayed ratios are approximately
  \textbf{(a)}~$1.03$ for near-uniform logits,
  \textbf{(b)}~$30$ for bimodal logits,
  \textbf{(c)}~$135$ for the heavy-tailed example, and
  \textbf{(d)}~$760$ for the sparse example.
  The exponentiated-score axes are logarithmic.}
  \label{fig:jensens inequality appendix}
\end{figure}
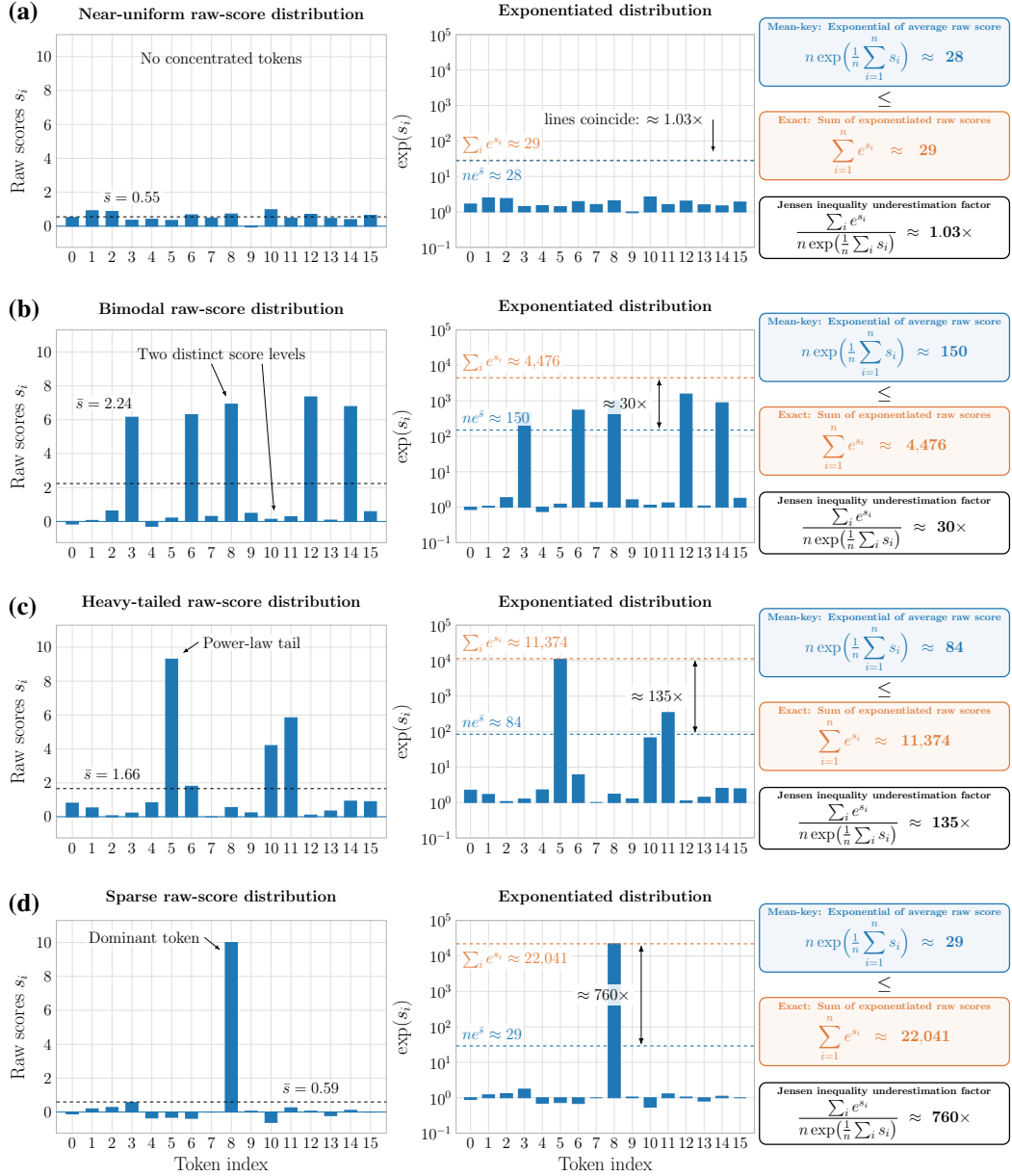

These examples ask how the distribution of logits within a group
changes the error in its mean-key mass surrogate. An arithmetic-mean
key gives the mean logit, whereas attention mass sums the exponentials
of individual logits. For $n$ keys, let
$\bar{\boldsymbol k}=n^{-1}\sum_{i=1}^n\boldsymbol k_i$.
Here only, absorb the scale $1/\sqrt d$ into the fixed query
$\boldsymbol q$. Convexity of the exponential gives
\begin{align}
 \exp\!\left(\frac{1}{n}\sum_{i=1}^n
       \boldsymbol q^{\mathsf T}\boldsymbol k_i\right)
 &=\exp\!\left(\boldsymbol q^{\mathsf T}\bar{\boldsymbol k}\right)
 \leq \frac{1}{n}\sum_{i=1}^n
       \exp\!\left(\boldsymbol q^{\mathsf T}\boldsymbol k_i\right),
 \notag\\
 \Longrightarrow\quad
 \sum_{i=1}^n\exp\!\left(\boldsymbol q^{\mathsf T}\boldsymbol k_i\right)
 &\geq n\exp\!\left(\boldsymbol q^{\mathsf T}\bar{\boldsymbol k}\right).
 \label{eq:Jensons}
\end{align}
The centroid mass surrogate on the right equals the exact unnormalized
mass on the left only when all member logits are equal.
Figure~\ref{fig:jensens inequality appendix} holds the group size at
16 logits while changing their distribution. The near-uniform example
has the smallest gap. In the bimodal, heavy-tailed, and sparse examples,
high-scoring members contribute much more to the exponential sum than
to the mean logit.

The different gaps explain why one mean key can distort the relative
weights of groups with different logit distributions. The inequality
concerns unnormalized mass, and it does not say that every normalized group
probability is underestimated. These are constructed examples of the
approximation.

\FloatBarrier

\begin{figure}[!htbp]
    \centering
    
    \includegraphics[width=\linewidth]{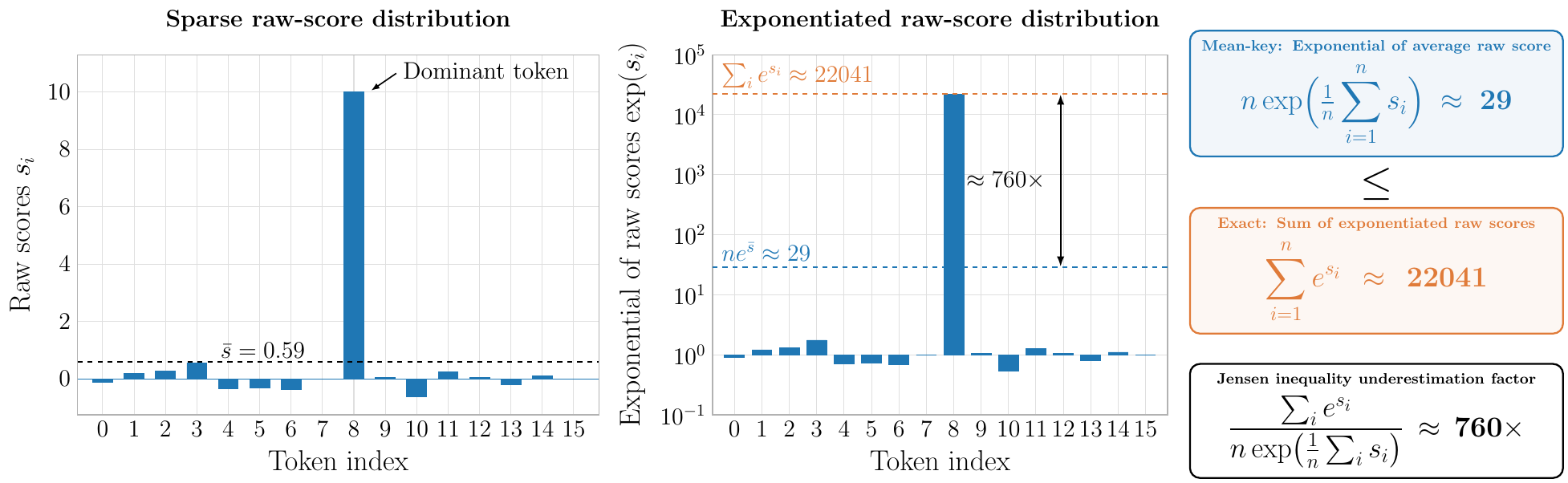}
    \caption{\textbf{A mean-key mass surrogate for 16 constructed logits.}
One high score dominates the exact mass, which is approximately
$760\times$ the surrogate.}
    \label{fig:jensen underestimation}
\end{figure}

{For the sparse-logit example in Figure~\ref{fig:jensen underestimation}, the exact mass is approximately $2.20\times10^4$, compared with approximately $29$ for the mean-key surrogate.} {The inequality explains this mass gap; the fixed-team experiment in Appendix~\ref{app:kl-variants} separately evaluates representative quality.}

\FloatBarrier

\subsection{{Representative choice on shared states and generated tasks}}
\label{app:kl-variants}

{We separate two questions: how closely a routing proposal follows dense
attention on fixed states, and how a representative affects full
generation.} {Both experiments use Qwen2.5-7B-Instruct and nominal 8K
RULER prompts.} {The first compares five proposals offline on 60 shared
dense trajectories.} {The second evaluates 13 tasks with 100 prompts per
task, giving 1,300 prompts per method and budget.}

\paragraph{{Partitions and representatives.}}
{The three team variants use the $P=16,R=4$ construction of
Section~\ref{sec:method}, with global k-means parents.} {They differ only
in the routing vector: the selected leader, the arithmetic mean of its
team, or a uniformly sampled member fixed for the prompt.} {The offline
diagnostic uses identical saved memberships; downstream runs use the
same seeded construction.} {Two additional controls use flat k-means
partitions of nominal size $B=4$ or $B=16$, represented by their mean
keys.} {The requested number of flat clusters is
$\min\{N,\max(2,\lfloor N/B\rfloor)\}$ per layer and KV head.}

\paragraph{{Shared-state routing diagnostic.}}
{We capture dense-generation queries, prompt keys, and group memberships
for 20 prompts each from CWE, S-NIAH-1, and QA-1.} {The model has 28 layers,
28 query heads, four KV heads, and head dimension $d=128$.} {Every variant
uses the same captured states.} {We retain at most the first 32 decode
steps per prompt: 640 steps for CWE, 195 for S-NIAH-1, and 211 for QA-1.}
{No sparse generation or group sampling is performed in this experiment,
so a sampling budget does not enter its KL values.}

{Let $g(i)$ be the group containing prompt token $i$, $n_g$ its size,
$\boldsymbol r_g$ its representative, and
$s_i=\boldsymbol q^\mathsf{T}\boldsymbol k_i/\sqrt d$.} {We define}
\begingroup
\begin{equation}
\begin{aligned}
 p_i&=\frac{e^{s_i}}{\sum_{j\in\mathrm{prompt}}e^{s_j}},&
 \ell_g&=\log n_g+\frac{\boldsymbol q^\mathsf{T}\boldsymbol r_g}{\sqrt d},\\
 Q_g&=\operatorname{softmax}(\ell)_g,&
 q_i&=\frac{Q_{g(i)}}{n_{g(i)}}.
\end{aligned}
\label{eq:representative-proposal}
\end{equation}
\endgroup
{The dense reference is restricted to the prompt prefix and renormalized
there.} {Generated suffix tokens are excluded from both distributions.}
{The diagnostic proposal $q_i$ is distinct from the conditional
team-inclusion probability used by the sampled estimator.}
{Writing $P_g=\sum_{i\in g}p_i$, the forward KL decomposes as}
\begingroup
\begin{align}
 \mathrm{KL}(p\|q)
 &=\sum_g P_g\log\frac{P_g}{Q_g}
   +\sum_g P_g\sum_{i\in g}\frac{p_i}{P_g}
            \log\frac{p_i/P_g}{1/n_g}.
\label{eq:diagnostic-kl}
\end{align}
\endgroup
{The first term measures disagreement in group mass.} {The second measures
within-group variation and is identical across the three fixed-team
arms.} {We use natural logarithms and report nats.} {We first average over
each prompt's retained steps, layers, and query heads, then average the
20 prompt means equally.} {Longer generations do not receive greater
prompt weight.}

{Dense reference dot products and representative means are formed in
float32 from the captured states; representative routing dot products use
float64.} {Probability and KL calculations use float64 log-space
arithmetic without flooring either distribution; saved metric arrays use
float32.} {The random-member diagnostic uses a deterministic seed derived
from the prompt UID, layer, KV head, and base seed 20260805, and preserves
that representative across steps.} {Table~\ref{tab:representative-kl-full}
reports all five arms and both KL components.}

\begin{table}[!htbp]
\centering

\footnotesize
\setlength{\tabcolsep}{3pt}
\renewcommand{\arraystretch}{1.08}
\begin{tabular}{llrrr}
\toprule
Task & Routing representative & Total KL & Group KL & Within-group KL \\
\midrule
CWE & Selected leader & 1.6212 & 0.9635 & 0.6577 \\
CWE & Team centroid & 1.4666 & 0.8088 & 0.6577 \\
CWE & Random member & 1.8908 & 1.2330 & 0.6577 \\
CWE & Flat centroid $B=16$ & 3.4293 & 1.7597 & 1.6696 \\
CWE & Flat centroid $B=4$ & 2.8060 & 1.7383 & 1.0677 \\
\midrule
S-NIAH-1 & Selected leader & 1.2480 & 0.6626 & 0.5854 \\
S-NIAH-1 & Team centroid & 1.1074 & 0.5220 & 0.5854 \\
S-NIAH-1 & Random member & 1.4976 & 0.9122 & 0.5854 \\
S-NIAH-1 & Flat centroid $B=16$ & 2.8271 & 1.1469 & 1.6803 \\
S-NIAH-1 & Flat centroid $B=4$ & 2.0759 & 1.1530 & 0.9229 \\
\midrule
QA-1 & Selected leader & 1.6128 & 0.9372 & 0.6756 \\
QA-1 & Team centroid & 1.4624 & 0.7868 & 0.6756 \\
QA-1 & Random member & 1.9123 & 1.2367 & 0.6756 \\
QA-1 & Flat centroid $B=16$ & 3.4946 & 1.7824 & 1.7122 \\
QA-1 & Flat centroid $B=4$ & 2.4289 & 1.4886 & 0.9403 \\
\bottomrule
\end{tabular}
\caption{  \textbf{Routing KL with all five representative/grouping variants.} Values are in nats, with 20 equally weighted prompts per task and at most the first 32 decode steps per prompt. All proposals use the same captured dense states. The three team arms share identical memberships; flat-centroid arms use their own partitions.}
\label{tab:representative-kl-full}
\end{table}

{Team centroids reduce total KL relative to leaders by
0.1546, 0.1406, and 0.1504 nats on CWE, S-NIAH-1, and QA-1, respectively.}
{The within-team component is unchanged.} {Both team choices have lower
total KL than either flat-centroid control.} {Thus the fixed-state
experiment attributes part of the gain to team construction and shows
that the selected actual key need not be the best representative under
this diagnostic.}

\paragraph{{Downstream generation.}}
{We evaluate 100 prompts per task, selected with seed 42 from
\nolinkurl{SaylorTwift/RULER-8192-Qwen2.5-3B-tokenizer} and re-tokenized
with Qwen2.5-7B-Instruct.} {The model and cache use FP16, batch size one,
the Qwen chat template, greedy generation, seed 0, and the standard
RULER task output limits.} {The prompt-plus-output admission guard is
8,704 tokens to accommodate formatting and generation beyond the
nominal 8K dataset length.}

{All sparse arms use dense prefill, sparse decoding in every layer,
and the exact generated suffix.} {We replay the final prompt token for
the first sparse prediction.} {The team variants and flat $B=4$ select
$S/4$ groups; flat $B=16$ selects $S/16$, for
$S\in\{128,256,512\}$.} {All use Gumbel sampling and the conditional
inclusion correction in Appendix~\ref{app:algorithm}.} {Grouping follows
that appendix's k-means settings.} {Generated histories may diverge
between arms.}

\paragraph{{Scores and uncertainty.}}
{We use the RULER grader, including partial credit on multi-answer tasks.}
{Per-task 95\% percentile-bootstrap intervals use 10,000 resamples and
seed 20260903.} {The 13-task mean uses a task-stratified bootstrap with
10,000 resamples and seed 20260805.} {Dense SDPA is evaluated once and
reused across budgets.} {These are individual-score intervals, not
paired-difference intervals.}

\paragraph{{Logical access.}}
{We use Equation~\ref{eq:qwen-logical-access}, counting selected KV rows
and routing vectors separately.} {Within each task, we pool logical
counts across prompts before forming access ratios; aggregate access is
the equal-weight mean of the 13 task-level percentages.}
{Different representatives can select different row unions, so scores
and actual access are reported together.}

\begin{table}[!htbp]
\centering

\footnotesize
\setlength{\tabcolsep}{3pt}
\renewcommand{\arraystretch}{1.08}
\begin{tabular}{lrcrrr}
\toprule
Variant & $S$ & Score [95\% CI] & KV (\%) & Routing (\%) & Total (\%) \\
\midrule
Dense SDPA & --- & 87.09 [85.78, 88.39] & 100.000 & 0.000 & 100.000 \\
\midrule
Selected leader & 128 & 84.16 [82.71, 85.61] & 12.641 & 12.333 & 24.974 \\
Selected leader & 256 & 84.94 [83.59, 86.28] & 21.459 & 12.333 & 33.792 \\
Selected leader & 512 & 86.23 [84.9, 87.59] & 34.643 & 12.334 & 46.977 \\
\midrule
Team centroid & 128 & 83.02 [81.62, 84.41] & 11.690 & 12.334 & 24.024 \\
Team centroid & 256 & 84.37 [83.05, 85.71] & 20.528 & 12.334 & 32.862 \\
Team centroid & 512 & 85.58 [84.28, 86.92] & 33.998 & 12.334 & 46.332 \\
\midrule
Random member & 128 & 77.28 [75.68, 78.94] & 12.223 & 12.332 & 24.555 \\
Random member & 256 & 81.18 [79.64, 82.69] & 20.917 & 12.333 & 33.250 \\
Random member & 512 & 83.37 [81.92, 84.83] & 34.038 & 12.334 & 46.372 \\
\midrule
Flat centroid $B=16$ & 128 & 71.86 [70.19, 73.5] & 8.912 & 3.111 & 12.023 \\
Flat centroid $B=16$ & 256 & 78.29 [76.79, 79.78] & 16.326 & 3.112 & 19.438 \\
Flat centroid $B=16$ & 512 & 83.04 [81.74, 84.41] & 28.671 & 3.113 & 31.784 \\
\midrule
Flat centroid $B=4$ & 128 & 82.31 [80.96, 83.66] & 9.837 & 12.459 & 22.296 \\
Flat centroid $B=4$ & 256 & 84.23 [82.97, 85.5] & 17.735 & 12.460 & 30.195 \\
Flat centroid $B=4$ & 512 & 85.66 [84.34, 86.99] & 30.412 & 12.460 & 42.872 \\
\bottomrule
\end{tabular}
\caption{  \textbf{Representative/grouping ablation on all 13 RULER-8K tasks.}
Each score macro-averages 100 prompts per task; intervals use a
task-stratified prompt bootstrap. Access is computed from pooled logical
counts within each task and then averaged equally across the 13 tasks;
total access includes routing. Routing is total minus KV access using
the reported rounded values.}
\label{tab:ruler_variants_8k}
\end{table}

{The selected leader has the higher aggregate score point estimate at
each evaluated budget, with 0.645--0.950 percentage points more total
access than the same-team centroid.} {This small downstream difference
coexists with the centroid's lower shared-state KL: the diagnostic
uses three tasks and fixed dense states, while the generated-task
evaluation covers 13 tasks and allows trajectories to diverge.}
{Tables~\ref{tab:representative-dense}--\ref{tab:representative-access-512}
give every task's scores, intervals, and access.} {These experiments
evaluate k-means parents at 8K; they do not compare representatives at
the contiguous $S=1024$ systems operating point.}

\begin{table}[!htbp]
\centering

\footnotesize
\setlength{\tabcolsep}{3pt}
\renewcommand{\arraystretch}{1.08}
\begin{tabular}{lc}
\toprule
Task & Dense SDPA score [95\% CI] \\
\midrule
S-NIAH-1 & 100 [100, 100] \\
S-NIAH-2 & 100 [100, 100] \\
S-NIAH-3 & 100 [100, 100] \\
MK-NIAH-1 & 100 [100, 100] \\
MK-NIAH-2 & 98 [95, 100] \\
MK-NIAH-3 & 97 [93, 100] \\
MV-NIAH & 98 [96.5, 99.01] \\
MQ-NIAH & 99.75 [99.25, 100] \\
VT & 35.8 [26.6, 44.8] \\
CWE & 93.3 [91.4, 94.9] \\
FWE & 80.33 [77, 83.67] \\
QA-1 & 71 [62, 80] \\
QA-2 & 59 [50, 68] \\
\bottomrule
\end{tabular}
\caption{  \textbf{Dense control for the representative ablation.} The same 100 prompts per task are used for every sparse arm and budget. Dense access is 100\% for each task.}
\label{tab:representative-dense}
\end{table}

\begin{table}[!htbp]
\centering

\footnotesize
\setlength{\tabcolsep}{4pt}
\renewcommand{\arraystretch}{1.15}
\begin{tabular}{lccccc}
\toprule
Task & Leader & Team centroid & Random member & Flat $B=16$ & Flat $B=4$ \\
\midrule
S-NIAH-1 & \shortstack{100\\{\scriptsize[100, 100]}} & \shortstack{98\\{\scriptsize[95, 100]}} & \shortstack{99\\{\scriptsize[97, 100]}} & \shortstack{97\\{\scriptsize[93, 100]}} & \shortstack{100\\{\scriptsize[100, 100]}} \\
S-NIAH-2 & \shortstack{100\\{\scriptsize[100, 100]}} & \shortstack{99\\{\scriptsize[97, 100]}} & \shortstack{99\\{\scriptsize[97, 100]}} & \shortstack{97\\{\scriptsize[93, 100]}} & \shortstack{100\\{\scriptsize[100, 100]}} \\
S-NIAH-3 & \shortstack{100\\{\scriptsize[100, 100]}} & \shortstack{98\\{\scriptsize[95, 100]}} & \shortstack{98\\{\scriptsize[95, 100]}} & \shortstack{87\\{\scriptsize[80, 93]}} & \shortstack{98\\{\scriptsize[95, 100]}} \\
MK-NIAH-1 & \shortstack{99\\{\scriptsize[97, 100]}} & \shortstack{100\\{\scriptsize[100, 100]}} & \shortstack{95\\{\scriptsize[90, 99]}} & \shortstack{96\\{\scriptsize[92, 99]}} & \shortstack{99\\{\scriptsize[97, 100]}} \\
MK-NIAH-2 & \shortstack{94\\{\scriptsize[89, 98]}} & \shortstack{96\\{\scriptsize[92, 99]}} & \shortstack{83\\{\scriptsize[76, 89]}} & \shortstack{81\\{\scriptsize[73, 88]}} & \shortstack{98\\{\scriptsize[95, 100]}} \\
MK-NIAH-3 & \shortstack{91\\{\scriptsize[85, 96]}} & \shortstack{88\\{\scriptsize[81, 94]}} & \shortstack{35\\{\scriptsize[26, 45]}} & \shortstack{21\\{\scriptsize[14, 29]}} & \shortstack{93\\{\scriptsize[88, 97]}} \\
MV-NIAH & \shortstack{93.5\\{\scriptsize[90.5, 95.76]}} & \shortstack{90.75\\{\scriptsize[88, 93.25]}} & \shortstack{88.25\\{\scriptsize[84.75, 91.5]}} & \shortstack{84.5\\{\scriptsize[81.75, 87.25]}} & \shortstack{88\\{\scriptsize[85.5, 90.5]}} \\
MQ-NIAH & \shortstack{98.25\\{\scriptsize[96.75, 99.5]}} & \shortstack{99.25\\{\scriptsize[98.25, 100]}} & \shortstack{98.5\\{\scriptsize[97, 99.75]}} & \shortstack{96.75\\{\scriptsize[95, 98.25]}} & \shortstack{98.5\\{\scriptsize[97.25, 99.5]}} \\
VT & \shortstack{33.8\\{\scriptsize[24, 42.8]}} & \shortstack{17.4\\{\scriptsize[10.39, 24.2]}} & \shortstack{21.2\\{\scriptsize[14.2, 28.4]}} & \shortstack{15.2\\{\scriptsize[8.8, 22]}} & \shortstack{16.2\\{\scriptsize[9.6, 24]}} \\
CWE & \shortstack{88.9\\{\scriptsize[86.5, 91.1]}} & \shortstack{88.2\\{\scriptsize[85.9, 90.5]}} & \shortstack{87.7\\{\scriptsize[85.5, 89.9]}} & \shortstack{66\\{\scriptsize[62.3, 69.4]}} & \shortstack{82.3\\{\scriptsize[79.2, 85.4]}} \\
FWE & \shortstack{78.67\\{\scriptsize[74.67, 82.34]}} & \shortstack{78.67\\{\scriptsize[74.33, 82.67]}} & \shortstack{76\\{\scriptsize[71.67, 80.33]}} & \shortstack{70.67\\{\scriptsize[66, 75.01]}} & \shortstack{77\\{\scriptsize[73.66, 80.01]}} \\
QA-1 & \shortstack{65\\{\scriptsize[56, 74]}} & \shortstack{68\\{\scriptsize[59, 77]}} & \shortstack{70\\{\scriptsize[60, 79]}} & \shortstack{69\\{\scriptsize[60, 78]}} & \shortstack{65\\{\scriptsize[55, 74]}} \\
QA-2 & \shortstack{52\\{\scriptsize[42, 61]}} & \shortstack{58\\{\scriptsize[48, 67]}} & \shortstack{54\\{\scriptsize[44, 64]}} & \shortstack{53\\{\scriptsize[43, 62]}} & \shortstack{55\\{\scriptsize[45, 64]}} \\
\bottomrule
\end{tabular}
\caption{  \textbf{Per-task representative-ablation scores at $S=128$.} Each cell gives the mean RULER score and its 95\% percentile bootstrap interval on 100 prompts. The nominal context is 8K; the dense control is in Table~\ref{tab:representative-dense}.}
\label{tab:representative-score-128}
\end{table}

\begin{table}[!htbp]
\centering

\footnotesize
\setlength{\tabcolsep}{3pt}
\renewcommand{\arraystretch}{1.08}
\begin{tabular}{lccccc}
\toprule
Task & Leader & Team centroid & Random member & Flat $B=16$ & Flat $B=4$ \\
\midrule
S-NIAH-1 & 10.104 / 22.535 & 9.693 / 22.126 & 9.900 / 22.331 & 7.636 / 10.753 & 8.425 / 20.902 \\
S-NIAH-2 & 13.072 / 25.369 & 11.782 / 24.080 & 12.564 / 24.860 & 8.847 / 11.959 & 9.581 / 22.057 \\
S-NIAH-3 & 13.222 / 25.500 & 11.901 / 24.184 & 12.744 / 25.016 & 8.997 / 12.105 & 10.083 / 22.530 \\
MK-NIAH-1 & 12.652 / 24.952 & 11.403 / 23.702 & 12.169 / 24.466 & 8.434 / 11.549 & 9.503 / 21.979 \\
MK-NIAH-2 & 13.193 / 25.566 & 12.457 / 24.829 & 12.760 / 25.132 & 9.575 / 12.692 & 10.399 / 22.874 \\
MK-NIAH-3 & 11.427 / 23.791 & 10.815 / 23.179 & 11.363 / 23.720 & 8.997 / 12.102 & 9.558 / 21.991 \\
MV-NIAH & 13.573 / 25.822 & 12.276 / 24.528 & 12.992 / 25.242 & 9.104 / 12.207 & 10.362 / 22.787 \\
MQ-NIAH & 13.384 / 25.615 & 12.082 / 24.313 & 12.802 / 25.034 & 8.983 / 12.082 & 10.099 / 22.507 \\
VT & 9.695 / 22.105 & 9.229 / 21.639 & 9.506 / 21.916 & 7.210 / 10.326 & 7.912 / 20.382 \\
CWE & 12.525 / 24.847 & 12.003 / 24.325 & 12.274 / 24.592 & 9.111 / 12.215 & 10.216 / 22.651 \\
FWE & 11.187 / 23.606 & 10.560 / 22.980 & 10.820 / 23.236 & 8.255 / 11.373 & 9.168 / 21.653 \\
QA-1 & 16.041 / 28.367 & 14.796 / 27.125 & 15.556 / 27.885 & 11.191 / 14.308 & 12.122 / 24.600 \\
QA-2 & 14.253 / 26.585 & 12.969 / 25.299 & 13.445 / 25.778 & 9.511 / 12.629 & 10.455 / 22.939 \\
\bottomrule
\end{tabular}
\caption{  \textbf{Per-task logical access at $S=128$.} Each cell gives KV access / total access in percent; total includes routing keys. Values are pooled-count ratios over each arm's generation. Dense is 100 / 100 for every task.}
\label{tab:representative-access-128}
\end{table}

\begin{table}[!htbp]
\centering

\footnotesize
\setlength{\tabcolsep}{4pt}
\renewcommand{\arraystretch}{1.15}
\begin{tabular}{lccccc}
\toprule
Task & Leader & Team centroid & Random member & Flat $B=16$ & Flat $B=4$ \\
\midrule
S-NIAH-1 & \shortstack{100\\{\scriptsize[100, 100]}} & \shortstack{100\\{\scriptsize[100, 100]}} & \shortstack{100\\{\scriptsize[100, 100]}} & \shortstack{99\\{\scriptsize[97, 100]}} & \shortstack{100\\{\scriptsize[100, 100]}} \\
S-NIAH-2 & \shortstack{99\\{\scriptsize[97, 100]}} & \shortstack{100\\{\scriptsize[100, 100]}} & \shortstack{98\\{\scriptsize[95, 100]}} & \shortstack{98\\{\scriptsize[95, 100]}} & \shortstack{100\\{\scriptsize[100, 100]}} \\
S-NIAH-3 & \shortstack{100\\{\scriptsize[100, 100]}} & \shortstack{98\\{\scriptsize[95, 100]}} & \shortstack{99\\{\scriptsize[97, 100]}} & \shortstack{97\\{\scriptsize[93.98, 100]}} & \shortstack{98\\{\scriptsize[95, 100]}} \\
MK-NIAH-1 & \shortstack{100\\{\scriptsize[100, 100]}} & \shortstack{99\\{\scriptsize[97, 100]}} & \shortstack{99\\{\scriptsize[97, 100]}} & \shortstack{98\\{\scriptsize[95, 100]}} & \shortstack{100\\{\scriptsize[100, 100]}} \\
MK-NIAH-2 & \shortstack{98\\{\scriptsize[95, 100]}} & \shortstack{97\\{\scriptsize[93, 100]}} & \shortstack{93\\{\scriptsize[87, 98]}} & \shortstack{92\\{\scriptsize[87, 97]}} & \shortstack{97\\{\scriptsize[94, 100]}} \\
MK-NIAH-3 & \shortstack{95\\{\scriptsize[91, 99]}} & \shortstack{95\\{\scriptsize[91, 99]}} & \shortstack{63\\{\scriptsize[54, 72]}} & \shortstack{68\\{\scriptsize[59, 77]}} & \shortstack{98\\{\scriptsize[95, 100]}} \\
MV-NIAH & \shortstack{94.25\\{\scriptsize[92, 96.5]}} & \shortstack{93.25\\{\scriptsize[91, 95.25]}} & \shortstack{91.25\\{\scriptsize[88.5, 93.75]}} & \shortstack{88\\{\scriptsize[85.25, 90.75]}} & \shortstack{91\\{\scriptsize[88.75, 93.25]}} \\
MQ-NIAH & \shortstack{99.75\\{\scriptsize[99.25, 100]}} & \shortstack{99\\{\scriptsize[98, 99.75]}} & \shortstack{98.5\\{\scriptsize[97.25, 99.5]}} & \shortstack{97.75\\{\scriptsize[96.25, 99]}} & \shortstack{98\\{\scriptsize[96.25, 99.5]}} \\
VT & \shortstack{26\\{\scriptsize[18, 35]}} & \shortstack{18.8\\{\scriptsize[11, 26.8]}} & \shortstack{21.4\\{\scriptsize[13.6, 30]}} & \shortstack{7.8\\{\scriptsize[3, 13.01]}} & \shortstack{16.4\\{\scriptsize[9.6, 24]}} \\
CWE & \shortstack{89.2\\{\scriptsize[86.6, 91.6]}} & \shortstack{89.8\\{\scriptsize[87.4, 92]}} & \shortstack{88.5\\{\scriptsize[86.2, 90.6]}} & \shortstack{77.6\\{\scriptsize[74.5, 80.7]}} & \shortstack{85.9\\{\scriptsize[83.5, 88.1]}} \\
FWE & \shortstack{79\\{\scriptsize[75, 82.67]}} & \shortstack{82\\{\scriptsize[78.33, 85.33]}} & \shortstack{82.67\\{\scriptsize[79, 86.01]}} & \shortstack{74.67\\{\scriptsize[70.67, 78.33]}} & \shortstack{79.67\\{\scriptsize[76, 83.33]}} \\
QA-1 & \shortstack{67\\{\scriptsize[57, 76]}} & \shortstack{69\\{\scriptsize[60, 77]}} & \shortstack{69\\{\scriptsize[60, 78]}} & \shortstack{67\\{\scriptsize[58, 76]}} & \shortstack{72\\{\scriptsize[63, 80]}} \\
QA-2 & \shortstack{57\\{\scriptsize[48, 67]}} & \shortstack{56\\{\scriptsize[46, 65]}} & \shortstack{52\\{\scriptsize[43, 61]}} & \shortstack{53\\{\scriptsize[44, 62.02]}} & \shortstack{59\\{\scriptsize[49, 68]}} \\
\bottomrule
\end{tabular}
\caption{  \textbf{Per-task representative-ablation scores at $S=256$.} Each cell gives the mean RULER score and its 95\% percentile bootstrap interval on 100 prompts. The nominal context is 8K; the dense control is in Table~\ref{tab:representative-dense}.}
\label{tab:representative-score-256}
\end{table}

\begin{table}[!htbp]
\centering

\footnotesize
\setlength{\tabcolsep}{3pt}
\renewcommand{\arraystretch}{1.08}
\begin{tabular}{lccccc}
\toprule
Task & Leader & Team centroid & Random member & Flat $B=16$ & Flat $B=4$ \\
\midrule
S-NIAH-1 & 17.896 / 30.329 & 17.516 / 29.948 & 17.657 / 30.090 & 14.565 / 17.683 & 15.814 / 28.292 \\
S-NIAH-2 & 22.457 / 34.754 & 21.211 / 33.509 & 21.848 / 34.140 & 16.422 / 19.537 & 17.845 / 30.318 \\
S-NIAH-3 & 22.499 / 34.775 & 21.122 / 33.407 & 21.705 / 33.982 & 16.505 / 19.614 & 18.067 / 30.523 \\
MK-NIAH-1 & 21.748 / 34.047 & 20.406 / 32.706 & 21.050 / 33.348 & 15.894 / 19.008 & 17.372 / 29.849 \\
MK-NIAH-2 & 22.224 / 34.597 & 21.618 / 33.991 & 21.708 / 34.081 & 17.590 / 20.708 & 18.661 / 31.136 \\
MK-NIAH-3 & 19.307 / 31.671 & 18.729 / 31.093 & 19.190 / 31.554 & 15.888 / 18.995 & 16.823 / 29.257 \\
MV-NIAH & 22.598 / 34.852 & 21.320 / 33.574 & 21.871 / 34.124 & 16.563 / 19.666 & 18.306 / 30.730 \\
MQ-NIAH & 22.295 / 34.528 & 21.042 / 33.270 & 21.619 / 33.848 & 16.255 / 19.355 & 17.944 / 30.350 \\
VT & 16.734 / 29.143 & 16.216 / 28.626 & 16.459 / 28.869 & 13.229 / 16.345 & 14.397 / 26.867 \\
CWE & 21.052 / 33.371 & 20.546 / 32.867 & 20.699 / 33.018 & 16.224 / 19.332 & 18.166 / 30.605 \\
FWE & 19.660 / 32.079 & 19.019 / 31.438 & 19.247 / 31.665 & 15.583 / 18.702 & 16.904 / 29.389 \\
QA-1 & 26.735 / 39.060 & 25.533 / 37.862 & 25.754 / 38.082 & 20.079 / 23.197 & 21.416 / 33.895 \\
QA-2 & 23.763 / 36.095 & 22.581 / 34.909 & 23.116 / 35.449 & 17.437 / 20.557 & 18.840 / 31.323 \\
\bottomrule
\end{tabular}
\caption{  \textbf{Per-task logical access at $S=256$.} Each cell gives KV access / total access in percent; total includes routing keys. Values are pooled-count ratios over each arm's generation. Dense is 100 / 100 for every task.}
\label{tab:representative-access-256}
\end{table}

\begin{table}[!htbp]
\centering

\footnotesize
\setlength{\tabcolsep}{4pt}
\renewcommand{\arraystretch}{1.15}
\begin{tabular}{lccccc}
\toprule
Task & Leader & Team centroid & Random member & Flat $B=16$ & Flat $B=4$ \\
\midrule
S-NIAH-1 & \shortstack{100\\{\scriptsize[100, 100]}} & \shortstack{100\\{\scriptsize[100, 100]}} & \shortstack{100\\{\scriptsize[100, 100]}} & \shortstack{100\\{\scriptsize[100, 100]}} & \shortstack{100\\{\scriptsize[100, 100]}} \\
S-NIAH-2 & \shortstack{100\\{\scriptsize[100, 100]}} & \shortstack{100\\{\scriptsize[100, 100]}} & \shortstack{100\\{\scriptsize[100, 100]}} & \shortstack{100\\{\scriptsize[100, 100]}} & \shortstack{100\\{\scriptsize[100, 100]}} \\
S-NIAH-3 & \shortstack{100\\{\scriptsize[100, 100]}} & \shortstack{100\\{\scriptsize[100, 100]}} & \shortstack{99\\{\scriptsize[97, 100]}} & \shortstack{100\\{\scriptsize[100, 100]}} & \shortstack{100\\{\scriptsize[100, 100]}} \\
MK-NIAH-1 & \shortstack{100\\{\scriptsize[100, 100]}} & \shortstack{100\\{\scriptsize[100, 100]}} & \shortstack{100\\{\scriptsize[100, 100]}} & \shortstack{100\\{\scriptsize[100, 100]}} & \shortstack{98\\{\scriptsize[95, 100]}} \\
MK-NIAH-2 & \shortstack{95\\{\scriptsize[90, 99]}} & \shortstack{94\\{\scriptsize[89, 98]}} & \shortstack{93\\{\scriptsize[87, 98]}} & \shortstack{94\\{\scriptsize[89, 98]}} & \shortstack{96\\{\scriptsize[92, 99]}} \\
MK-NIAH-3 & \shortstack{98\\{\scriptsize[95, 100]}} & \shortstack{98\\{\scriptsize[95, 100]}} & \shortstack{83\\{\scriptsize[75.98, 90]}} & \shortstack{95\\{\scriptsize[91, 99]}} & \shortstack{98\\{\scriptsize[95, 100]}} \\
MV-NIAH & \shortstack{95.75\\{\scriptsize[93.75, 97.5]}} & \shortstack{94.5\\{\scriptsize[92.25, 96.5]}} & \shortstack{93.75\\{\scriptsize[91.5, 95.75]}} & \shortstack{88.75\\{\scriptsize[86, 91.25]}} & \shortstack{94.5\\{\scriptsize[92.25, 96.5]}} \\
MQ-NIAH & \shortstack{100\\{\scriptsize[100, 100]}} & \shortstack{99.25\\{\scriptsize[98.25, 100]}} & \shortstack{98.75\\{\scriptsize[97.5, 99.75]}} & \shortstack{98.75\\{\scriptsize[97.75, 99.75]}} & \shortstack{99.25\\{\scriptsize[98.25, 100]}} \\
VT & \shortstack{34.8\\{\scriptsize[26, 44]}} & \shortstack{27\\{\scriptsize[18.98, 36]}} & \shortstack{25.6\\{\scriptsize[17.6, 34.6]}} & \shortstack{18.8\\{\scriptsize[11.8, 26]}} & \shortstack{24.8\\{\scriptsize[16.6, 33]}} \\
CWE & \shortstack{92.1\\{\scriptsize[90.2, 93.8]}} & \shortstack{91.4\\{\scriptsize[89.3, 93.2]}} & \shortstack{90.4\\{\scriptsize[87.9, 92.4]}} & \shortstack{80.9\\{\scriptsize[77.2, 84.4]}} & \shortstack{90.4\\{\scriptsize[88, 92.5]}} \\
FWE & \shortstack{80.33\\{\scriptsize[77, 83.34]}} & \shortstack{82.33\\{\scriptsize[79, 85.67]}} & \shortstack{78.33\\{\scriptsize[74.67, 81.33]}} & \shortstack{79.33\\{\scriptsize[75.33, 83]}} & \shortstack{83.67\\{\scriptsize[80, 87]}} \\
QA-1 & \shortstack{70\\{\scriptsize[61, 79]}} & \shortstack{69\\{\scriptsize[60, 78]}} & \shortstack{68\\{\scriptsize[59, 77]}} & \shortstack{69\\{\scriptsize[60, 78]}} & \shortstack{70\\{\scriptsize[60.98, 79]}} \\
QA-2 & \shortstack{55\\{\scriptsize[45, 64]}} & \shortstack{57\\{\scriptsize[48, 66.02]}} & \shortstack{54\\{\scriptsize[44, 64]}} & \shortstack{55\\{\scriptsize[45, 65]}} & \shortstack{59\\{\scriptsize[49, 68]}} \\
\bottomrule
\end{tabular}
\caption{  \textbf{Per-task representative-ablation scores at $S=512$.} Each cell gives the mean RULER score and its 95\% percentile bootstrap interval on 100 prompts. The nominal context is 8K; the dense control is in Table~\ref{tab:representative-dense}.}
\label{tab:representative-score-512}
\end{table}

\begin{table}[!htbp]
\centering

\footnotesize
\setlength{\tabcolsep}{3pt}
\renewcommand{\arraystretch}{1.08}
\begin{tabular}{lccccc}
\toprule
Task & Leader & Team centroid & Random member & Flat $B=16$ & Flat $B=4$ \\
\midrule
S-NIAH-1 & 30.158 / 42.596 & 29.623 / 42.059 & 29.700 / 42.135 & 25.975 / 29.094 & 27.719 / 40.200 \\
S-NIAH-2 & 36.190 / 48.488 & 35.422 / 47.717 & 35.471 / 47.769 & 29.267 / 32.383 & 30.857 / 43.333 \\
S-NIAH-3 & 35.910 / 48.192 & 35.176 / 47.460 & 35.201 / 47.484 & 29.144 / 32.256 & 30.993 / 43.450 \\
MK-NIAH-1 & 35.331 / 47.630 & 34.405 / 46.704 & 34.535 / 46.835 & 28.247 / 31.363 & 30.195 / 42.670 \\
MK-NIAH-2 & 35.404 / 47.779 & 35.089 / 47.462 & 34.895 / 47.268 & 30.146 / 33.264 & 31.425 / 43.901 \\
MK-NIAH-3 & 31.626 / 43.990 & 31.160 / 43.524 & 31.503 / 43.866 & 27.325 / 30.432 & 28.645 / 41.079 \\
MV-NIAH & 36.092 / 48.346 & 35.323 / 47.577 & 35.359 / 47.610 & 29.065 / 32.169 & 31.154 / 43.580 \\
MQ-NIAH & 35.731 / 47.964 & 34.864 / 47.094 & 35.005 / 47.235 & 28.723 / 31.823 & 30.716 / 43.123 \\
VT & 27.971 / 40.381 & 27.407 / 39.816 & 27.632 / 40.042 & 23.651 / 26.767 & 25.334 / 37.805 \\
CWE & 33.808 / 46.129 & 33.576 / 45.898 & 33.404 / 45.725 & 28.124 / 31.233 & 30.715 / 43.153 \\
FWE & 32.883 / 45.302 & 32.453 / 44.872 & 32.250 / 44.669 & 28.116 / 31.236 & 29.649 / 42.133 \\
QA-1 & 41.533 / 53.860 & 40.705 / 53.029 & 40.737 / 53.066 & 34.135 / 37.253 & 35.852 / 48.330 \\
QA-2 & 37.717 / 50.048 & 36.766 / 49.097 & 36.795 / 49.126 & 30.805 / 33.924 & 32.097 / 44.583 \\
\bottomrule
\end{tabular}
\caption{  \textbf{Per-task logical access at $S=512$.} Each cell gives KV access / total access in percent; total includes routing keys. Values are pooled-count ratios over each arm's generation. Dense is 100 / 100 for every task.}
\label{tab:representative-access-512}
\end{table}

\FloatBarrier

\subsection{Centroid-distance diagnostic}
\label{app:centroid-distance}

\begin{figure}[!htb]
  \centering
  \includegraphics[width=\linewidth]{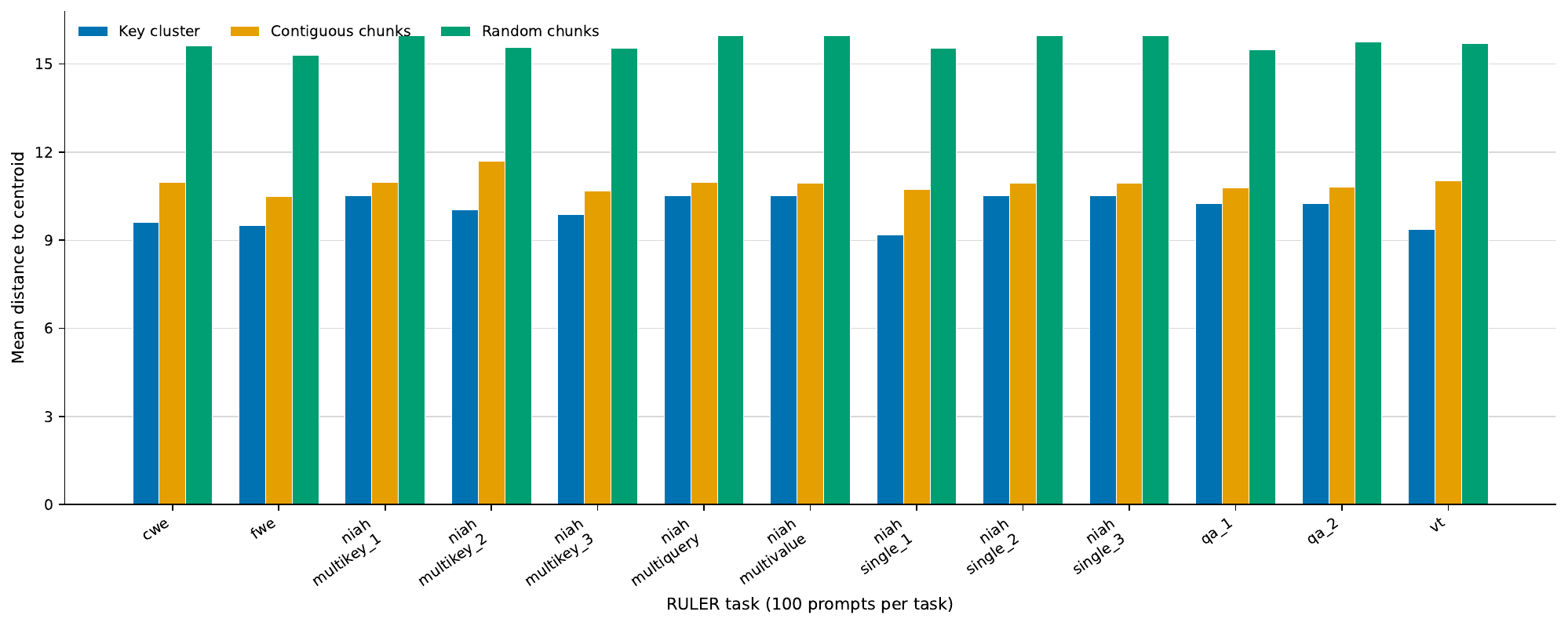}
  \caption{\textbf{Key-to-centroid distance across RULER tasks.}
  Mean distance for key clusters, contiguous chunks, and random chunks
  on 100 prompts per task, using Qwen2.5-7B-Instruct at 8,192-token
  context and nominal group size $P=16$. Contiguous chunks lie between
  key clusters and random chunks on every plotted task.}
  \label{fig:centroid}
\end{figure}

Contiguous parents avoid clustering, but do they still group nearby
keys? Figure~\ref{fig:centroid} compares key clusters, contiguous
chunks, and random chunks using the mean Euclidean distance from each
raw key vector to the arithmetic centroid of its group. Here, distances to group centroids measure group compactness. The experiment uses Qwen2.5-7B-Instruct at
8,192-token context, with 100 prompts per RULER task and group size
$P=16$.

Key clustering applies k-means independently for each layer and KV
head. Contiguous grouping partitions the token sequence into consecutive
groups of 16, while the random baseline uses one fixed random shuffle
of token positions before partitioning them into groups of 16. For each
task, we report the arithmetic mean of token-to-centroid distances over
all tokens, KV heads, layers, and prompts.

Across all 13 tasks, contiguous chunks have larger mean distances than
key clusters but smaller distances than random chunks. Position thus
retains some local structure in the keys without requiring key-space
clustering, while key clustering produces the more compact groups on
these inputs. This gives a geometric reason to consider contiguous
parents rather than random grouping.

\FloatBarrier
\subsection{Oracle prefill-window diagnostic}
\label{app:prefill-window}

We measure how much attention mass a small number of contiguous windows
can retain compared with selecting the same number of tokens individually.
This oracle analysis uses dense attention weights from the final prefill
query token. It uses 10 prompts per task for CWE and MV-NIAH, with
Qwen2.5-7B-Instruct at context lengths of 8,192 and 32,768 tokens. We
partition the prompt into disjoint windows of width $w=16$ and select
$K_{\mathrm w}=32$ windows.

For each prompt, layer, and attention head, let $a_i$ be the dense
attention weight on token $i$. Let $\mathcal I_{K_{\mathrm w}}$ index the
$K_{\mathrm w}$ windows with the largest summed attention weight, and
let $\mathcal W_j$ contain the token positions in window $j$. As a
reference with the same $wK_{\mathrm w}=512$-token budget, let
$\mathcal D_{wK_{\mathrm w}}$ contain the 512 token positions with the
largest individual attention weights for the same prompt, layer, and
head. The window effectiveness ratio (WER) is
\begin{equation}
 \mathrm{WER}
 =\frac{\displaystyle\sum_{j\in\mathcal I_{K_{\mathrm w}}}
                  \sum_{i\in\mathcal W_j}a_i}
        {\displaystyle\sum_{i\in\mathcal D_{wK_{\mathrm w}}}a_i}.
 \label{eq:WER}
\end{equation}
Because the reference set contains the 512 highest-attention tokens,
$0\leq\mathrm{WER}\leq1$. A WER close to one means that the selected
windows retain nearly as much attention mass as the unconstrained
512-token reference set. Table~\ref{tab:wer_ratio} shows that a majority
of heads reach the $0.50$ threshold in every task--context pair, whereas
only $15.1\%$--$23.6\%$ reach $0.90$. Thus the 32 windows capture at
least half of the reference mass for most heads, while near-complete
recovery is less common.

\begin{table}[!htb]
\centering
\begin{tabular}{lccc}
\toprule
Task & Context length & WER threshold & Heads (\%) \\
\midrule
CWE & 8192  & $\geq 0.50$ & 82.1 [79.8, 84.9] \\
CWE & 32768 & $\geq 0.50$ & 80.2 [76.9, 84.8] \\
\midrule
CWE & 8192  & $\geq 0.75$ & 54.0 [48.4, 60.7] \\
CWE & 32768 & $\geq 0.75$ & 49.5 [41.7, 60.7] \\
\midrule
CWE & 8192  & $\geq 0.90$ & 22.1 [18.4, 27.0] \\
CWE & 32768 & $\geq 0.90$ & 23.6 [16.2, 36.4] \\
\midrule
MV-NIAH & 8192  & $\geq 0.50$ & 79.0 [75.2, 83.4] \\
MV-NIAH & 32768 & $\geq 0.50$ & 73.8 [72.0, 75.7] \\
\midrule
MV-NIAH & 8192  & $\geq 0.75$ & 51.3 [43.7, 60.7] \\
MV-NIAH & 32768 & $\geq 0.75$ & 42.6 [39.4, 46.0] \\
\midrule
MV-NIAH & 8192  & $\geq 0.90$ & 22.0 [15.6, 30.8] \\
MV-NIAH & 32768 & $\geq 0.90$ & 15.1 [13.4, 17.0] \\
\bottomrule
\end{tabular}
\caption{\textbf{Oracle prefill-window effectiveness.}
Percentage of heads across all layers meeting each WER threshold from
Equation~\ref{eq:WER}. For each prompt, layer, and head, WER compares
the attention mass in the top $K_{\mathrm w}=32$ disjoint windows of
width $w=16$ with the mass in the $wK_{\mathrm w}=512$
highest-attention tokens. Results use 10 prompts per task at 8K and
32K context. Brackets give 95\% bootstrap confidence intervals from
resampling prompts.}
\label{tab:wer_ratio}
\end{table}

\FloatBarrier
\subsection{Key versus probe-fingerprint clustering}
\label{app:key-fingerprint}

Clustering keys directly avoids forming their attention-score
fingerprints under probe queries. This separate evaluation uses
Qwen2.5-7B-Instruct at 8K context on 50 prompts per task for CWE,
MK-NIAH-3, QA-1, and QA-2, with $S\in\{128,256,512\}$ and
$P=16,R=4$.

For key clustering, we apply k-means directly to the unnormalized key
vectors independently for each prompt, layer, and KV head. For
probe-fingerprint clustering, the final 64 prefill-token queries serve
as probes. Each key is represented by its dot-product scores with these
probe queries from the query heads sharing its KV head, and each
fingerprint dimension is z-score standardized across prompt tokens
before k-means. The two variants otherwise use the same team
construction, attention estimator, and sampling procedure. Error bars
are 95\% percentile-bootstrap confidence intervals from 10,000
resamples.

Key clustering has higher point estimates on MK-NIAH-3 at all three
budgets, whereas probe fingerprints have higher CWE point estimates
at $S=128$ and $256$. On QA-2, the ordering reverses with budget, and
both clustering variants remain below the dense point estimate. The
QA intervals overlap substantially.

There is no uniform accuracy advantage from constructing probe
fingerprints in this comparison. Direct key clustering is a simpler
preparation choice, while the lower-budget CWE results identify a
setting in which probe fingerprints have the higher point estimates.

\begin{figure}[!htb]
  \centering
  \includegraphics[width=\linewidth]{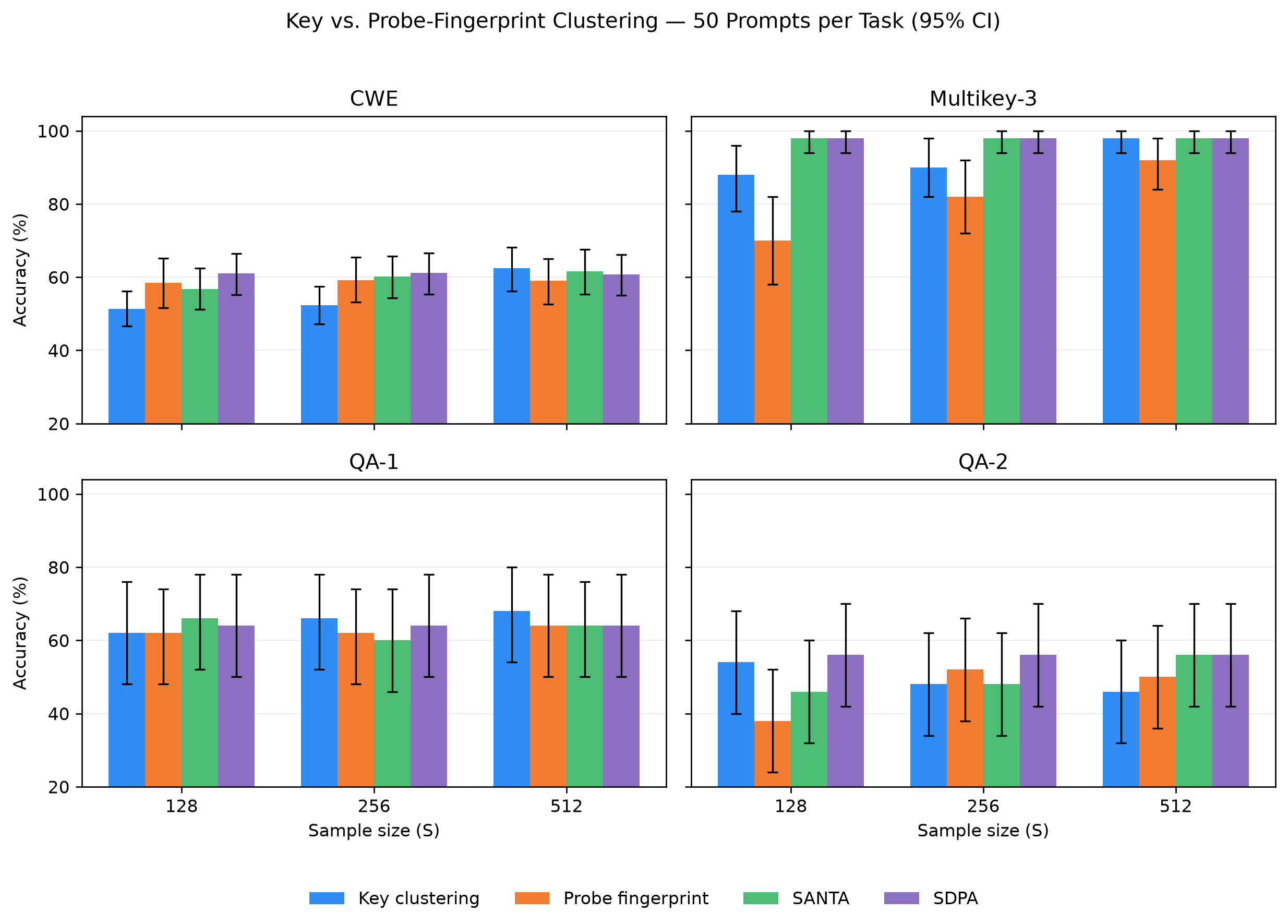}
  \caption{\textbf{Key versus probe-fingerprint clustering.}
  Accuracy on four RULER tasks with 50 prompts per task and nominal
  budgets $S=128,256,512$. Bars compare key clustering, probe-fingerprint
  clustering, SANTA, and dense SDPA; error bars denote the displayed
  95\% confidence intervals.}
  \label{fig:key-fingerprint}
\end{figure}
\FloatBarrier

\section{{Routing, inclusion correction, and grouping ablations}}
\label{app:new-ablations}

\subsection{{Diagnosing low-budget retrieval failures}}
\label{app:routing-failures}

{We study HELMET JSON-KV and native RULER MK-NIAH-3 at 32K,
with 50 matched prompts per control.} {Qwen2.5-7B-Instruct uses FP16,
batch size one, dense prefill, and sparse decoding in all layers.}
{The $P=16,R=4$ k-means-parent construction follows
Appendix~\ref{app:algorithm}, with $\lfloor N/16\rfloor$ parents.}
{Every sparse query selects $K=32$ teams ($S=128$); only generated
tokens are attended exactly.} {Token generation is greedy.}

\paragraph{{Controls.}}
{We compare dense SDPA, the unmodified SANTA\texttt{++} implementation,
an instrumented PyTorch implementation, and two interventions using
that implementation.} {The instrumented baseline uses the ordinary
leader score plus log team size, Gumbel top-32 sampling, and the
conditional inclusion correction.}

{The \emph{answer-guided} intervention locates answer-token sequences
in the prompt and forces their teams into the selected set.} {It matches
the answer alone, its space-prefixed form, and its quoted form, taking
the union of matching positions.} {When more than 32 teams qualify, it
keeps the 32 with highest ordinary routing logits; otherwise it fills
the remaining slots from the original sampled priorities.} {Forced
teams receive correction factor one.} {Retained sampled teams keep their
original factors.} {This intervention changes both selection and weighting.}

{The \emph{exact team-mass oracle} computes exact query--key scores over
the whole prompt, sums the exponentiated scores within each team, and
selects the 32 teams with highest mass.} {It reads every member of those
teams and sets their correction factors to one.} {This oracle uses a full-key scan to test the potential benefit
of more accurate team selection.}

\paragraph{{JSON-KV population and evaluation.}}
{We use HELMET's \nolinkurl{json_kv/test_k440_dep6.jsonl}, with 440
key/value pairs per example, taking the first 50 prompts after the
loader's seed-42 shuffle.} {The HELMET revision is
\nolinkurl{af609c4d51b97fc35012099380aa889da961c42d}.}
{The official prompt contains two demonstrations and ends with
\texttt{Corresponding value:}; no model chat template is applied.}
{The maximum context is 32,768 and the output cap is 100 tokens, with
greedy generation and no newline stopping.} {All outputs reach the cap.} {Prompt length averages 30,413.54 tokens, ranging from 30,280
to 30,545.} {Per-example routing seed is 42 plus the zero-based example
index.} {The score is HELMET substring exact match after its text
normalization and response parsing.} {The unmodified JSON-KV arm uses
Triton decode; the instrumented baseline and interventions use PyTorch.}

\paragraph{{MK-NIAH-3 population and evaluation.}}
{We select 50 prompts from
\nolinkurl{SaylorTwift/RULER-32768-Qwen2.5-3B-tokenizer}, using the native
validation loader and the RULER grader at revision
\nolinkurl{38da79d79519ef87aa46ae804f838e1eab7f86d7}.}
{We sample prompts without replacement using base selection seed 42;
all five controls use the same prompts.} {The complete prepared prompt,
including its answer prefix, is one user turn in the model chat
template.} {Greedy generation stops at EOS or 128 new tokens.} {Mean prompt
length is 32,481.48 tokens, with range 32,329--32,616, and prompt plus
output budget must fit within 32,768.} {Routing uses deterministic prompt-specific seeds derived from
base seed 0.} {Each selected example has one required
answer, so the official case-insensitive substring-recall score is
binary here.} {A missing prediction in the answer-guided condition counts as incorrect.}
{The unmodified and instrumented baselines produce identical
predictions on these 50 RULER examples.} {These runs use Triton
preparation and PyTorch decoding.}

\begin{table}[!htbp]
\centering

\footnotesize
\setlength{\tabcolsep}{4pt}
\renewcommand{\arraystretch}{1.08}
\begin{tabular}{lcrcl}
\toprule
Control & \shortstack{JSON-KV\\correct} & \shortstack{JSON-KV\\score (\%)} & \shortstack{MK-NIAH-3\\correct} & \shortstack{MK-NIAH-3 score (\%)\\{[95\% CI]}} \\
\midrule
Dense SDPA & 49 / 50 & 98.00 & 44 / 50 & 88.00 [78.00, 96.00] \\
Unmodified SANTA++ & 19 / 50 & 38.00 & 8 / 50 & 16.00 [6.00, 26.00] \\
Instrumented baseline & 23 / 50 & 46.00 & 8 / 50 & 16.00 [6.00, 26.00] \\
Answer-guided & 46 / 50 & 92.00 & 33 / 50 & 66.00 [52.00, 78.00] \\
Exact team-mass oracle & 48 / 50 & 96.00 & 47 / 50 & 94.00 [86.00, 100.00] \\
\bottomrule
\end{tabular}
\caption{  \textbf{Low-budget routing interventions on matched 50-prompt cohorts.} The instrumented baseline is the reference for both interventions. JSON-KV uses substring exact match; the selected MK-NIAH-3 population has one required answer per prompt. All sparse arms request $K=32$ teams. Oracle scores are diagnostic results, not sparse-runtime measurements.}
\label{tab:routing-interventions}
\end{table}

{MK-NIAH-3 intervals are prompt-level 95\% percentile bootstrap
intervals from 10,000 resamples with seed 20260903.} {JSON-KV rows report
the exact counts and scores for the 50-prompt cohort.} {These experiments
use one model/routing realization per control and prompt.}

\paragraph{{Answer-token coverage.}}
{For each layer/query-head/decode-step call, we measure whether the
selected teams include any or all located answer tokens, and the
fraction included.} {We average calls within each prompt and then weight
prompts equally.} {The located sets contain an average of 31.76 tokens
for JSON-KV and 32.56 for MK-NIAH-3 because tokenization forms and
multiple occurrences are unioned.} {Coverage is therefore a call-level
diagnostic, not prompt accuracy.}

\begin{table}[!htbp]
\centering

\scriptsize
\setlength{\tabcolsep}{3pt}
\renewcommand{\arraystretch}{1.08}
\begin{tabular}{llrrrrr}
\toprule
Cohort & Arm & \shortstack{Any hit\\(\%)} & \shortstack{Complete hit\\(\%)} & \shortstack{Coverage\\(\%)} & \shortstack{First-step\\any hit (\%)} & \shortstack{First-step\\complete hit (\%)} \\
\midrule
JSON-KV & Instrumented & 15.2172 & 0.0620 & 3.1941 & 22.6122 & 0.1122 \\
JSON-KV & Answer-guided & 100.0000 & 99.3929 & 99.9703 & 100.0000 & 99.3929 \\
JSON-KV & Team-mass oracle & 17.3602 & 0.4799 & 5.5960 & 25.7015 & 0.5918 \\
\midrule
MK-NIAH-3 & Instrumented & 14.0461 & 0.0177 & 2.5516 & 11.1199 & 0.0026 \\
MK-NIAH-3 & Answer-guided & 100.0000 & 99.1607 & 99.9608 & 100.0000 & 99.1607 \\
MK-NIAH-3 & Team-mass oracle & 18.1570 & 0.3366 & 5.4788 & 11.7959 & 0.0102 \\
\bottomrule
\end{tabular}
\caption{  \textbf{Coverage of located answer tokens.} Entries equally weight each prompt's mean over layers, query heads, and decode steps. First-step columns restrict to step zero. Any/complete hit count calls containing at least one/all located answer tokens; coverage averages the fraction included.}
\label{tab:routing-coverage}
\end{table}

{Forcing answer-containing teams recovers most of the JSON-KV accuracy
gap.} {On MK-NIAH-3, nearly complete answer-token coverage raises
accuracy only to 66\%, whereas exact team-mass selection reaches 94\%.}
{This indicates that literal answer availability alone is insufficient
under the intervention.} {Query-dependent contextual selection may also
matter, alongside changes in weighting.} {The oracle results show substantial room to improve routing
at the same 32-team budget.} {Their purpose is to test selection
quality; computing exact team masses requires reading all prompt keys.}

\FloatBarrier
\subsection{{Conditional inclusion-correction ablation}}
\label{app:inclusion-ablation}

{We compare the corrected estimator with an otherwise unchanged
estimator that sets every sampled team's factor to $c_g=1$.}
{For selected prompt rows $A$ and exact generated suffix $E$, the
corrected output is}
\begingroup
\begin{equation}
 \widehat{\boldsymbol o}_{\mathrm{corr}}=
 \frac{\sum_{i\in A}e^{s_i}\boldsymbol v_i/c_{g(i)}+
       \sum_{j\in E}e^{s_j}\boldsymbol v_j}
      {\sum_{i\in A}e^{s_i}/c_{g(i)}+
       \sum_{j\in E}e^{s_j}}.
\label{eq:correction-ablation}
\end{equation}
\endgroup
{The ablation removes $-\log c_g$ from sampled-token log weights before
the same normalized reduction.} {It leaves the routing rule and team
construction unchanged.} {Full greedy generations are compared on
matched prompts; their later queries and sampled sets can diverge
after different predicted tokens.}

\paragraph{{Setup.}}
{We use $S=128$ and 50 matched prompts per task with
Qwen2.5-7B-Instruct, FP16 model/cache storage, and batch size one.}
{Both the 8K and 32K evaluations cover all 13 RULER tasks,
using the same 50 prompts per task for both estimators.}
{Both use k-means parents, $P=16,R=4$, dense prefill, all-layer sparse
decoding, and an exact generated suffix.} {The Qwen chat template,
greedy generation, and standard RULER output limits are unchanged.}
{We report task means and 95\% percentile-bootstrap intervals from
10,000 prompt resamples per estimator.}

\paragraph{{Access and aggregate effects.}}
{Access follows Equation~\ref{eq:qwen-logical-access}.} {The
13-task means are 85.21 with correction and 83.97 without it at 8K,
and 68.88 versus 71.11 at 32K.} {Effects vary by task: correction
adds 12.40 points on 8K VT but reduces the 32K CWE point estimate
by 17.00 points.} {The correction makes the unnormalized attention
sums unbiased; its effect on downstream task accuracy depends on
the task and context.}

\begin{table}[!htbp]
\centering

\footnotesize
\setlength{\tabcolsep}{4pt}
\renewcommand{\arraystretch}{1.08}
\begin{tabular}{llcrrr}
\toprule
Task & Estimator & RULER score [95\% CI] & KV (\%) & Routing (\%) & Total (\%) \\
\midrule
S-NIAH-1 & Corrected & 100.00 [100.00, 100.00] & 10.04 & 12.43 & 22.48 \\
S-NIAH-1 & $c_g=1$ & 100.00 [100.00, 100.00] & 10.17 & 12.43 & 22.61 \\
\addlinespace[2pt]
S-NIAH-2 & Corrected & 100.00 [100.00, 100.00] & 13.11 & 12.30 & 25.40 \\
S-NIAH-2 & $c_g=1$ & 100.00 [100.00, 100.00] & 13.38 & 12.30 & 25.68 \\
\addlinespace[2pt]
S-NIAH-3 & Corrected & 96.00 [90.00, 100.00] & 13.11 & 12.29 & 25.40 \\
S-NIAH-3 & $c_g=1$ & 100.00 [100.00, 100.00] & 13.52 & 12.28 & 25.80 \\
\addlinespace[2pt]
MK-NIAH-1 & Corrected & 100.00 [100.00, 100.00] & 12.69 & 12.30 & 24.99 \\
MK-NIAH-1 & $c_g=1$ & 98.00 [94.00, 100.00] & 12.97 & 12.30 & 25.27 \\
\addlinespace[2pt]
MK-NIAH-2 & Corrected & 98.00 [94.00, 100.00] & 13.20 & 12.38 & 25.57 \\
MK-NIAH-2 & $c_g=1$ & 94.00 [86.00, 100.00] & 13.49 & 12.37 & 25.86 \\
\addlinespace[2pt]
MK-NIAH-3 & Corrected & 92.00 [84.00, 98.00] & 11.43 & 12.36 & 23.80 \\
MK-NIAH-3 & $c_g=1$ & 92.00 [84.00, 98.00] & 11.70 & 12.36 & 24.06 \\
\addlinespace[2pt]
MV-NIAH & Corrected & 91.50 [88.00, 94.50] & 13.56 & 12.25 & 25.82 \\
MV-NIAH & $c_g=1$ & 93.50 [89.50, 97.00] & 13.92 & 12.25 & 26.17 \\
\addlinespace[2pt]
MQ-NIAH & Corrected & 99.50 [98.50, 100.00] & 13.35 & 12.23 & 25.58 \\
MQ-NIAH & $c_g=1$ & 100.00 [100.00, 100.00] & 13.78 & 12.24 & 26.02 \\
\addlinespace[2pt]
VT & Corrected & 40.00 [26.00, 54.00] & 9.68 & 12.41 & 22.09 \\
VT & $c_g=1$ & 27.60 [16.00, 40.00] & 9.83 & 12.41 & 22.24 \\
\addlinespace[2pt]
CWE & Corrected & 89.40 [86.00, 92.60] & 12.48 & 12.32 & 24.81 \\
CWE & $c_g=1$ & 87.80 [85.40, 90.20] & 12.89 & 12.32 & 25.21 \\
\addlinespace[2pt]
FWE & Corrected & 81.34 [76.67, 86.00] & 11.28 & 12.42 & 23.70 \\
FWE & $c_g=1$ & 74.67 [70.00, 80.00] & 11.59 & 12.42 & 24.01 \\
\addlinespace[2pt]
QA-1 & Corrected & 68.00 [54.00, 80.00] & 16.07 & 12.32 & 28.39 \\
QA-1 & $c_g=1$ & 74.00 [62.00, 86.00] & 16.46 & 12.32 & 28.78 \\
\addlinespace[2pt]
QA-2 & Corrected & 52.00 [38.00, 66.00] & 14.10 & 12.33 & 26.43 \\
QA-2 & $c_g=1$ & 50.00 [36.00, 64.00] & 14.47 & 12.33 & 26.80 \\
\bottomrule
\end{tabular}
\caption{  \textbf{Inclusion-correction ablation at 8K, $S=128$.} Each estimator uses 50 matched prompts per task. Score intervals are marginal 95\% percentile bootstrap intervals from 10,000 resamples. KV, routing, and total are logical vector-read percentages relative to dense GQA, pooled across each generation.}
\label{tab:correction-8k}
\end{table}

\begin{table}[!htbp]
\centering

\footnotesize
\setlength{\tabcolsep}{4pt}
\renewcommand{\arraystretch}{1.08}
\begin{tabular}{llcrrr}
\toprule
Task & Estimator & RULER score [95\% CI] & KV (\%) & Routing (\%) & Total (\%) \\
\midrule
S-NIAH-1 & Corrected & 94.00 [86.00, 100.00] & 3.69 & 11.96 & 15.66 \\
S-NIAH-1 & $c_g=1$ & 100.00 [100.00, 100.00] & 3.75 & 11.96 & 15.71 \\
\addlinespace[2pt]
S-NIAH-2 & Corrected & 98.00 [94.00, 100.00] & 5.22 & 11.57 & 16.78 \\
S-NIAH-2 & $c_g=1$ & 98.00 [94.00, 100.00] & 5.33 & 11.57 & 16.90 \\
\addlinespace[2pt]
S-NIAH-3 & Corrected & 98.00 [94.00, 100.00] & 5.17 & 11.58 & 16.75 \\
S-NIAH-3 & $c_g=1$ & 88.00 [78.00, 96.00] & 5.30 & 11.58 & 16.88 \\
\addlinespace[2pt]
MK-NIAH-1 & Corrected & 92.00 [84.00, 98.00] & 5.01 & 11.58 & 16.60 \\
MK-NIAH-1 & $c_g=1$ & 84.00 [74.00, 94.00] & 5.15 & 11.58 & 16.74 \\
\addlinespace[2pt]
MK-NIAH-2 & Corrected & 66.00 [52.00, 78.00] & 4.83 & 11.71 & 16.54 \\
MK-NIAH-2 & $c_g=1$ & 70.00 [58.00, 82.00] & 4.98 & 11.70 & 16.68 \\
\addlinespace[2pt]
MK-NIAH-3 & Corrected & 14.00 [6.00, 24.00] & 4.06 & 11.87 & 15.94 \\
MK-NIAH-3 & $c_g=1$ & 28.00 [16.00, 40.00] & 4.20 & 11.87 & 16.07 \\
\addlinespace[2pt]
MV-NIAH & Corrected & 82.50 [76.50, 88.00] & 5.25 & 11.57 & 16.82 \\
MV-NIAH & $c_g=1$ & 82.00 [76.00, 87.50] & 5.41 & 11.57 & 16.98 \\
\addlinespace[2pt]
MQ-NIAH & Corrected & 94.50 [91.50, 97.00] & 5.04 & 11.56 & 16.60 \\
MQ-NIAH & $c_g=1$ & 92.50 [88.50, 96.50] & 5.24 & 11.56 & 16.80 \\
\addlinespace[2pt]
VT & Corrected & 23.60 [13.60, 34.40] & 3.35 & 11.93 & 15.28 \\
VT & $c_g=1$ & 30.00 [18.80, 42.00] & 3.39 & 11.93 & 15.32 \\
\addlinespace[2pt]
CWE & Corrected & 54.20 [48.60, 59.60] & 4.63 & 11.54 & 16.17 \\
CWE & $c_g=1$ & 71.20 [66.40, 75.80] & 4.78 & 11.54 & 16.32 \\
\addlinespace[2pt]
FWE & Corrected & 70.67 [62.00, 78.67] & 4.19 & 11.82 & 16.02 \\
FWE & $c_g=1$ & 74.67 [68.67, 80.67] & 4.34 & 11.82 & 16.16 \\
\addlinespace[2pt]
QA-1 & Corrected & 68.00 [54.00, 80.00] & 5.55 & 11.54 & 17.09 \\
QA-1 & $c_g=1$ & 64.00 [50.00, 76.00] & 5.73 & 11.54 & 17.27 \\
\addlinespace[2pt]
QA-2 & Corrected & 40.00 [26.00, 54.00] & 5.20 & 11.66 & 16.86 \\
QA-2 & $c_g=1$ & 42.00 [28.00, 56.00] & 5.34 & 11.66 & 17.01 \\
\bottomrule
\end{tabular}
\caption{ \textbf{Inclusion-correction ablation at 32K, $S=128$.}
Each estimator uses 50 matched prompts per task across all 13 RULER tasks.
Score intervals are marginal 95\% percentile bootstrap intervals from
10,000 resamples. KV, routing, and total are logical vector-read
percentages relative to dense GQA, pooled across each generation.}
\label{tab:correction-32k}
\end{table}

\FloatBarrier
\subsection{{Parent size and representatives per parent}}
\label{app:pr-sensitivity}

{We vary k-means parent size $P$ and representatives per parent $R$ on two
RULER-8K tasks with Qwen2.5-7B-Instruct: 100 prompts from MQ-NIAH and
100 from MV-NIAH.} {Every sparse setting uses nominal $S=128$.} {We vary
$P\in\{8,16,32\}$ at $R=4$, and $R\in\{2,4,8\}$ at $P=16$.}
 {Nominal team size is
$m=P/R$ and the requested number of teams is $K=SR/P$.}

{The RULER score is the fraction of required answer substrings
recovered, averaged over prompts.} {We report
task-stratified, prompt-clustered percentile bootstrap intervals from
10,000 resamples, keeping the two tasks as separate strata.} {Values
are mean $\pm$ 95\% CI half-width.} {Access counts the GQA-aware
selected/exact KV reads and routing-key reads separately.}

\begin{table}[!htbp]
\centering

\footnotesize
\setlength{\tabcolsep}{3pt}
\renewcommand{\arraystretch}{1.08}
\begin{tabular}{lrrrrrrrr}
\toprule
Setting & $P$ & $R$ & $m$ & $K$ & \shortstack{KV-only\\(\%)} & \shortstack{Routing\\(\%)} & \shortstack{Total\\(\%)} & \shortstack{RULER score\\(\%)} \\
\midrule
SDPA & --- & --- & --- & --- & 100.00 & 0.00 & 100.00 & 98.88 $\pm$ 0.69 \\
Small parents & 8 & 4 & 2 & 64 & 13.03 & 23.05 & 36.08 & 97.88 $\pm$ 0.94 \\
Default & 16 & 4 & 4 & 32 & 13.44 & 12.24 & 25.68 & 96.00 $\pm$ 1.25 \\
Large parents & 32 & 4 & 8 & 16 & 14.21 & 6.18 & 20.40 & 92.62 $\pm$ 1.75 \\
Fewer representatives & 16 & 2 & 8 & 16 & 12.03 & 6.18 & 18.22 & 90.38 $\pm$ 1.94 \\
More representatives & 16 & 8 & 2 & 64 & 12.64 & 23.68 & 36.32 & 97.38 $\pm$ 1.06 \\
\bottomrule
\end{tabular}
\caption{  \textbf{Parent and representative sensitivity on MQ-NIAH and MV-NIAH at 8K.} All sparse settings use k-means parents and nominal $S=128$; $m=P/R$ and $K=SR/P$. Each setting covers 200 prompts. Scores show mean $\pm$ 95\% CI half-width; the two tasks remain separate bootstrap strata. Access includes the exact generated suffix and routing. The default row is shared by both sweeps.}
\label{tab:pr-sensitivity}
\end{table}

{Finer teams improve the accuracy point estimates while increasing
routing access.} {At fixed $R=4$, reducing $P$ from 32 to 8 increases
the score from 92.62 to 97.88 and total access from 20.40\% to 36.08\%.}
{At fixed $P=16$, increasing $R$ from 2 to 8 raises the score from 90.38
to 97.38 and access from 18.22\% to 36.32\%.} {Fixed nominal $S$ therefore
does not hold actual reads fixed: the number of summaries scanned and
teams selected both change.} {The main benchmark sweeps use $P=16,R=4$.}
\FloatBarrier

\section{Evaluation details and full benchmark results}
\label{app:evaluation}

\subsection{Setup and metrics}
\label{app:evaluation-setup}

We evaluate Qwen2.5-7B-Instruct with dense scaled dot-product attention
(SDPA), SANTA, and SANTA\texttt{++} using k-means or contiguous parent
groups. HELMET and RULER use 8K and 32K contexts; LongBench v2 uses a
32K context cap, retaining shorter prompts at their original length.
Budgets double from $S=128$ through 2048 for HELMET RAG, through
1024 for LongBench v2 and 8K RULER, and through 4096 for 32K RULER.
The complete 32K aggregate sweeps are reported in
Tables~\ref{tab:helmet-rag-32k}, \ref{tab:longbench_results},
and~\ref{tab:ruler_32k}, respectively.

For SANTA\texttt{++}, the nominal budget requests
$K=\min\{M,S/(P/R)\}$ teams, with integral $P/R$ and $S/(P/R)$,
where $M$ is the number of teams. This conversion requires the run's
$P,R$; it does not give the number of unique rows read. As specified in
Appendix~\ref{app:algorithm}, the reference partitions the prompt after
dense prefill and keeps newly generated cache rows in a separate exact
suffix. That suffix starts empty and grows as generated tokens are fed
back; no prompt window is reserved for exact evaluation.

{Absolute scores use each benchmark's own metric. Normalized scores
divide these by the corresponding SDPA score at the same context length
and express the result as a percentage.} HELMET, LongBench, and RULER report 95\%
bootstrap confidence interval half-widths. 

\paragraph{Reference logical-access counter.}
For one layer, KV head, and decoding call $c$, let $N_c$ be the current
cache length, $e_c$ the exact generated-suffix length, and $m_c$ the
number of team representatives scanned for routing. Let $I_{c,h}$ be
the selected prompt-row indices for query head $h$ and set
$u_c=|\bigcup_h I_{c,h}|$, with the union restricted to query heads
sharing that KV head. The reference counts one key and one value per
dense row and computes
\begin{equation}
 \operatorname{KV}_{\mathrm{total}}(\%)
 =100\,\frac{\sum_c\bigl[m_c+2(u_c+e_c)\bigr]}{\sum_c 2N_c}.
 \label{eq:qwen-logical-access}
\end{equation}
The selected prompt rows and exact suffix are disjoint. A representative
read for routing and again as a selected member counts in both passes.
Counts also add across layers and decoding steps. Prefill and team
construction are outside this counter. For SANTA, the corresponding
numerator per call is $N_c$ plus the number of distinct sampled value
rows across sharing query heads, so its full-key scan alone costs 50\%.
{This metric counts decode-time vector reads. Resident-cache storage
and physical DRAM traffic are measured separately in
Appendix~\ref{app:kernel-evaluation}.}

\paragraph{Access aggregation.}
The reference also records a selected-KV-only access measure that omits
the routing term $m_c$. The main RULER tables instead use the total-access
measure in Equation~\ref{eq:qwen-logical-access}, which includes routing.
For RULER aggregates, we sum the raw numerators and denominators across
prompts and tasks before division rather than averaging the displayed
task percentages. For HELMET RAG, aggregate access is the equal-weight
mean of the four RAG task-level total-access percentages, including
representative-key routing reads (Appendix~\ref{app:helmet}).

\subsection{HELMET}
\label{Appendix: helmet}
\label{app:helmet}

{Table~\ref{tab:helmet-rag-32k} reports the 32K RAG sweep, and
Table~\ref{tab:helmet-rag-8k} gives the corresponding 8K sweep.}
Both use $S\in\{128,256,512,1024,2048\}$. The RAG score
is the equal-weight mean of substring exact match on Natural Questions,
TriviaQA, PopQA, and HotpotQA. KV access is the equal-weight mean of
those four tasks' total-access percentages, including representative-key
routing reads. Tables~\ref{tab:helmet-8k-task} and
\ref{tab:helmet-32k-task} give individual RAG task results at all five
budgets; the non-RAG tasks and the all-task summary cover
$S\in\{128,256,512\}$ only.

The additional tasks expose losses outside the RAG mean. At 32K, JSON
KV scores fall from SDPA's 96.80\% to 45.60\% with k-means and
33.40\% with contiguous grouping at $S=128$. At $S=512$, these rise
to 87.00\% and 68.60\%, respectively. Non-RAG tasks use their own scoring metrics. The detailed tables'
``All tasks average'' summarizes all 13 tasks.

\FloatBarrier
\begin{table}[!htbp]
\centering
\small
\begin{tabular}{llrrr}
\toprule
Method & $S$ & KV (\%) & Acc (norm) (\%) & Acc (abs) (\%) \\
\midrule
\multirow{1}{*}{SDPA} & -- & 100.00 & 100.00 & 62.65 $\pm$ 2.02 \\
\midrule
\multirow{5}{*}{SANTA} & 128 & 51.62 & 100.24 & 62.80 $\pm$ 2.02 \\
 & 256 & 52.70 & 99.60 & 62.40 $\pm$ 2.03 \\
 & 512 & 54.39 & 99.60 & 62.40 $\pm$ 2.03 \\
 & 1024 & 56.77 & 100.16 & 62.75 $\pm$ 2.05 \\
 & 2048 & 60.14 & 100.00 & 62.65 $\pm$ 2.07 \\
\midrule
\multirow{5}{*}{\shortstack{SANTA++\\(k-means)}} & 128 & 27.55 & 97.85 & 61.30 $\pm$ 1.98 \\
 & 256 & {37.59} & 98.72 & 61.85 $\pm$ 2.00 \\
 & 512 & {51.82} & 99.28 & 62.20 $\pm$ 2.00 \\
 & 1024 & 69.95 & 98.56 & 61.75 $\pm$ 2.00 \\
 & 2048 & 90.18 & 100.40 & 62.90 $\pm$ 2.05 \\
\midrule
\multirow{5}{*}{\shortstack{SANTA++\\(Contiguous)}} & 128 & 23.69 & 96.09 & 60.20 $\pm$ 2.02 \\
 & 256 & 32.38 & 100.80 & 63.15 $\pm$ 2.03 \\
 & 512 & 45.73 & 99.44 & 62.30 $\pm$ 2.02 \\
 & 1024 & 64.10 & 99.52 & 62.35 $\pm$ 2.05 \\
 & 2048 & {86.27} & 99.76 & 62.50 $\pm$ 2.07 \\
\bottomrule
\end{tabular}
\caption{ \textbf{HELMET RAG: full 8K budget sweep.}
Scores and access weight the four RAG tasks equally. Normalized scores
are relative to SDPA. Access includes routing; $\pm$ gives 95\%
bootstrap interval half-widths.}
\label{tab:helmet-rag-8k}
\end{table}

\begin{table*}[!t]
\centering
\small
\textbf{(a) RAG and reranking}\par\smallskip
\resizebox{\textwidth}{!}{
\begin{tabular}{llrrrrrrrrrr}
\toprule
\multirow{2}{*}{Method} & \multirow{2}{*}{$S$} & \multicolumn{2}{c}{Natural Questions} & \multicolumn{2}{c}{TriviaQA} & \multicolumn{2}{c}{PopQA} & \multicolumn{2}{c}{HotpotQA} & \multicolumn{2}{c}{MS MARCO} \\
 &  & Total access (\%) & Acc (\%) & Total access (\%) & Acc (\%) & Total access (\%) & Acc (\%) & Total access (\%) & Acc (\%) & Total access (\%) & Acc (\%) \\
\midrule
\multirow{1}{*}{SDPA} & -- & 100.00 & 53.20 $\pm$ 4.30 & 100.00 & 82.20 $\pm$ 3.30 & 100.00 & 64.40 $\pm$ 4.10 & 100.00 & 50.80 $\pm$ 4.40 & 100.00 & 58.51 $\pm$ 3.89 \\
\midrule
\multirow{5}{*}{SANTA} & 128 & 51.62 & 54.40 $\pm$ 4.40 & 51.55 & 82.40 $\pm$ 3.40 & 51.79 & 63.00 $\pm$ 4.20 & 51.51 & 51.40 $\pm$ 4.30 & 51.72 & 59.16 $\pm$ 3.81 \\
 & 256 & 52.73 & 53.80 $\pm$ 4.40 & 52.58 & 81.80 $\pm$ 3.40 & 52.97 & 63.80 $\pm$ 4.20 & 52.52 & 50.20 $\pm$ 4.50 & 52.75 & 58.03 $\pm$ 3.75 \\
 & 512 & 54.46 & 53.00 $\pm$ 4.30 & 54.21 & 82.20 $\pm$ 3.30 & 54.79 & 64.20 $\pm$ 4.20 & 54.09 & 50.20 $\pm$ 4.50 & 54.27 & 59.06 $\pm$ 3.84 \\
 & 1024 & 56.90 & 54.40 $\pm$ 4.30 & 56.49 & 82.20 $\pm$ 3.30 & 57.30 & 64.00 $\pm$ 4.10 & 56.39 & 50.40 $\pm$ 4.40 &  &  \\
 & 2048 & 60.37 & 54.00 $\pm$ 4.30 & 59.77 & 82.20 $\pm$ 3.30 & 60.83 & 64.00 $\pm$ 4.20 & 59.58 & 50.40 $\pm$ 4.40 &  &  \\
\midrule
\multirow{5}{*}{\shortstack{SANTA++\\(K-means)}} & 128 & 27.49 & 53.80 $\pm$ 4.40 & 26.87 & 83.20 $\pm$ 3.30 & 28.88 & 61.60 $\pm$ 4.30 & 26.96 & 46.60 $\pm$ 4.40 & 28.23 & 56.99 $\pm$ 3.84 \\
 & 256 & 37.46 & 54.80 $\pm$ 4.30 & 36.64 & 81.20 $\pm$ 3.50 & 39.55 & 64.20 $\pm$ 4.30 & 36.73 & 47.20 $\pm$ 4.30 & 37.32 & 57.91 $\pm$ 3.91 \\
 & 512 & 51.58 & 53.20 $\pm$ 4.50 & 50.57 & 83.20 $\pm$ 3.30 & 54.46 & 63.00 $\pm$ 4.20 & 50.69 & 49.40 $\pm$ 4.40 & 50.36 & 59.02 $\pm$ 3.67 \\
 & 1024 & 69.43 & 52.40 $\pm$ 4.40 & 68.53 & 81.80 $\pm$ 3.30 & 73.13 & 63.00 $\pm$ 4.20 & 68.70 & 49.80 $\pm$ 4.30 &  &  \\
 & 2048 & 89.46 & 54.20 $\pm$ 4.30 & 88.82 & 82.80 $\pm$ 3.30 & 93.44 & 64.00 $\pm$ 4.20 & 88.98 & 50.60 $\pm$ 4.40 &  &  \\
\midrule
\multirow{5}{*}{\shortstack{SANTA++\\(Contiguous)}} & 128 & 23.66 & 53.00 $\pm$ 4.40 & 23.05 & 80.40 $\pm$ 3.40 & 24.80 & 60.80 $\pm$ 4.30 & 23.25 & 46.60 $\pm$ 4.30 & 25.03 & 44.95 $\pm$ 3.99 \\
 & 256 & 32.28 & 55.20 $\pm$ 4.30 & 31.41 & 82.20 $\pm$ 3.40 & 34.15 & 63.80 $\pm$ 4.30 & 31.69 & 51.40 $\pm$ 4.40 & 33.13 & 49.59 $\pm$ 4.04 \\
 & 512 & 45.51 & 53.60 $\pm$ 4.40 & 44.35 & 82.40 $\pm$ 3.30 & 48.34 & 63.00 $\pm$ 4.40 & 44.72 & 50.20 $\pm$ 4.40 & 45.37 & 54.81 $\pm$ 3.53 \\
 & 1024 & 63.72 & 54.40 $\pm$ 4.30 & 62.44 & 81.60 $\pm$ 3.30 & 67.18 & 63.60 $\pm$ 4.30 & 63.07 & 49.80 $\pm$ 4.40 &  &  \\
 & 2048 & 85.50 & 53.20 $\pm$ 4.30 & 84.51 & 81.80 $\pm$ 3.40 & 90.06 & 64.20 $\pm$ 4.20 & 85.03 & 50.80 $\pm$ 4.40 &  &  \\
\bottomrule
\end{tabular}
}
\vspace{0.8em}
\resizebox{\textwidth}{!}{
\begin{tabular}{llrrrrrrrrrr}
\toprule
\multirow{2}{*}{Method} & \multirow{2}{*}{$S$} & \multicolumn{2}{c}{TREC coarse} & \multicolumn{2}{c}{TREC fine} & \multicolumn{2}{c}{NLU} & \multicolumn{2}{c}{Banking77} & \multicolumn{2}{c}{CLINC150} \\
 &  & Total access (\%) & Acc (\%) & Total access (\%) & Acc (\%) & Total access (\%) & Acc (\%) & Total access (\%) & Acc (\%) & Total access (\%) & Acc (\%) \\
\midrule
\multirow{1}{*}{SDPA} & -- & 100.00 & 83.00 $\pm$ 3.20 & 100.00 & 41.80 $\pm$ 4.40 & 100.00 & 79.60 $\pm$ 3.50 & 100.00 & 75.20 $\pm$ 3.80 & 100.00 & 81.60 $\pm$ 3.40 \\
\midrule
\multirow{5}{*}{SANTA} & 128 & 52.03 & 83.20 $\pm$ 3.30 & 52.00 & 40.20 $\pm$ 4.30 & 52.00 & 78.00 $\pm$ 3.60 & 52.01 & 76.00 $\pm$ 3.60 & 51.96 & 79.40 $\pm$ 3.60 \\
 & 256 & 53.39 & 83.00 $\pm$ 3.20 & 53.36 & 43.20 $\pm$ 4.30 & 53.38 & 79.60 $\pm$ 3.50 & 53.34 & 76.20 $\pm$ 3.70 & 53.30 & 80.40 $\pm$ 3.50 \\
 & 512 & 55.47 & 82.80 $\pm$ 3.30 & 55.44 & 42.40 $\pm$ 4.30 & 55.50 & 80.00 $\pm$ 3.50 & 55.37 & 75.60 $\pm$ 3.70 & 55.37 & 82.20 $\pm$ 3.40 \\
\midrule
\multirow{5}{*}{\shortstack{SANTA++\\(K-means)}} & 128 & 28.41 & 79.20 $\pm$ 3.60 & 27.95 & 39.00 $\pm$ 4.20 & 27.79 & 77.40 $\pm$ 3.60 & 27.41 & 74.20 $\pm$ 3.90 & 27.61 & 74.20 $\pm$ 3.90 \\
 & 256 & 38.74 & 83.00 $\pm$ 3.40 & 37.99 & 41.60 $\pm$ 4.30 & 37.85 & 78.20 $\pm$ 3.50 & 37.14 & 76.00 $\pm$ 3.70 & 37.39 & 77.80 $\pm$ 3.70 \\
 & 512 & 53.25 & 83.20 $\pm$ 3.20 & 52.16 & 41.80 $\pm$ 4.40 & 51.96 & 79.60 $\pm$ 3.60 & 50.94 & 76.20 $\pm$ 3.70 & 51.15 & 79.60 $\pm$ 3.50 \\
\midrule
\multirow{5}{*}{\shortstack{SANTA++\\(Contiguous)}} & 128 & 23.95 & 80.40 $\pm$ 3.40 & 23.71 & 35.80 $\pm$ 4.20 & 23.89 & 75.20 $\pm$ 3.90 & 23.47 & 72.80 $\pm$ 3.90 & 23.43 & 74.00 $\pm$ 3.80 \\
 & 256 & 32.85 & 83.40 $\pm$ 3.20 & 32.44 & 39.20 $\pm$ 4.40 & 32.74 & 79.20 $\pm$ 3.50 & 31.97 & 76.60 $\pm$ 3.80 & 31.94 & 77.20 $\pm$ 3.70 \\
 & 512 & 46.45 & 83.20 $\pm$ 3.30 & 45.81 & 42.80 $\pm$ 4.40 & 46.19 & 78.00 $\pm$ 3.70 & 44.98 & 76.00 $\pm$ 3.70 & 44.96 & 79.00 $\pm$ 3.50 \\
\bottomrule
\end{tabular}
}
\vspace{0.8em}
\resizebox{\textwidth}{!}{
\begin{tabular}{llrrrrrrrr}
\toprule
\multirow{2}{*}{Method} & \multirow{2}{*}{$S$} & \multicolumn{2}{c}{InfBench QA} & \multicolumn{2}{c}{InfBench MC} & \multicolumn{2}{c}{JSON KV} & \multicolumn{2}{c}{All tasks average} \\
 &  & Total access (\%) & Acc (\%) & Total access (\%) & Acc (\%) & Total access (\%) & Acc (\%) & Total access (\%) & Acc (\%) \\
\midrule
\multirow{1}{*}{SDPA} & -- & 100.00 & 20.08 $\pm$ 3.54 & 100.00 & 40.61 $\pm$ 6.33 & 100.00 & 99.40 $\pm$ 0.70 & 100.00 & 63.88 \\
\midrule
\multirow{5}{*}{SANTA} & 128 & 51.24 & 19.22 $\pm$ 3.51 & 50.99 & 42.36 $\pm$ 6.33 & 51.29 & 99.40 $\pm$ 0.70 & 51.67 & 63.70 \\
 & 256 & 52.07 & 19.95 $\pm$ 3.54 & 51.58 & 43.23 $\pm$ 6.33 & 52.09 & 99.40 $\pm$ 0.70 & 52.77 & 64.05 \\
 & 512 & 53.39 & 19.83 $\pm$ 3.54 & 52.50 & 41.48 $\pm$ 6.33 & 53.34 & 99.60 $\pm$ 0.50 & 54.48 & 64.04 \\
\midrule
\multirow{5}{*}{\shortstack{SANTA++\\(K-means)}} & 128 & 26.02 & 18.68 $\pm$ 3.48 & 24.33 & 43.23 $\pm$ 6.34 & 24.08 & 89.80 $\pm$ 2.70 & 27.08 & 61.38 \\
 & 256 & 35.80 & 20.15 $\pm$ 3.50 & 33.88 & 42.79 $\pm$ 6.55 & 32.21 & 98.00 $\pm$ 1.30 & 36.82 & 63.30 \\
 & 512 & 50.09 & 20.05 $\pm$ 3.62 & 48.38 & 43.67 $\pm$ 6.12 & 45.09 & 99.40 $\pm$ 0.70 & 50.82 & 63.95 \\
\midrule
\multirow{5}{*}{\shortstack{SANTA++\\(Contiguous)}} & 128 & 22.44 & 18.52 $\pm$ 3.40 & 21.19 & 42.36 $\pm$ 6.33 & 22.08 & 74.20 $\pm$ 3.80 & 23.38 & 58.39 \\
 & 256 & 30.71 & 18.51 $\pm$ 3.35 & 29.08 & 41.48 $\pm$ 6.55 & 29.37 & 92.20 $\pm$ 2.30 & 31.83 & 62.31 \\
 & 512 & 43.91 & 20.26 $\pm$ 3.56 & 42.28 & 41.92 $\pm$ 6.55 & 41.36 & 98.20 $\pm$ 1.10 & 44.94 & 63.34 \\
\bottomrule
\end{tabular}
}
\caption{\textbf{HELMET task results at 8K context length.}
{For individual tasks, $\pm$ denotes the half-width of a 95\%
bootstrap confidence interval. ``All tasks average'' is the equal-weight
mean of the 13 task scores.}}
\label{tab:helmet-8k-task}
\end{table*}

\begin{table*}[!t]
\centering
\small
\resizebox{\textwidth}{!}{
\begin{tabular}{llrrrrrrrrrr}
\toprule
\multirow{2}{*}{Method} & \multirow{2}{*}{$S$} & \multicolumn{2}{c}{Natural Questions} & \multicolumn{2}{c}{TriviaQA} & \multicolumn{2}{c}{PopQA} & \multicolumn{2}{c}{HotpotQA} & \multicolumn{2}{c}{MS MARCO} \\
 &  & Total access (\%) & Acc (\%) & Total access (\%) & Acc (\%) & Total access (\%) & Acc (\%) & Total access (\%) & Acc (\%) & Total access (\%) & Acc (\%) \\
\midrule
\multirow{1}{*}{SDPA} & -- & 100.00 & 46.20 $\pm$ 4.30 & 100.00 & 82.00 $\pm$ 3.50 & 100.00 & 62.80 $\pm$ 4.30 & 100.00 & 38.60 $\pm$ 4.20 & 100.00 & 34.39 $\pm$ 4.36 \\
\midrule
\multirow{5}{*}{SANTA} & 128 & 50.43 & 46.40 $\pm$ 4.30 & 50.41 & 81.00 $\pm$ 3.50 & 50.46 & 62.20 $\pm$ 4.20 & 50.41 & 38.00 $\pm$ 4.30 & 50.46 & 37.63 $\pm$ 4.60 \\
 & 256 & 50.77 & 45.60 $\pm$ 4.40 & 50.71 & 81.00 $\pm$ 3.40 & 50.81 & 62.80 $\pm$ 4.20 & 50.71 & 38.00 $\pm$ 4.40 & 50.77 & 35.79 $\pm$ 4.58 \\
 & 512 & 51.33 & 46.00 $\pm$ 4.40 & 51.23 & 81.40 $\pm$ 3.30 & 51.40 & 62.80 $\pm$ 4.20 & 51.23 & 39.80 $\pm$ 4.20 & 51.29 & 34.49 $\pm$ 4.35 \\
 & 1024 & 52.23 & 46.40 $\pm$ 4.40 & 52.05 & 80.40 $\pm$ 3.40 & 52.32 & 62.80 $\pm$ 4.20 & 52.06 & 38.80 $\pm$ 4.10 &  &  \\
 & 2048 & 53.65 & 46.80 $\pm$ 4.40 & 53.35 & 80.40 $\pm$ 3.40 & 53.76 & 62.80 $\pm$ 4.30 & 53.37 & 38.80 $\pm$ 4.30 &  &  \\
\midrule
\multirow{5}{*}{\shortstack{SANTA++\\(K-means)}} & 128 & 17.48 & 43.80 $\pm$ 4.30 & 17.27 & 80.80 $\pm$ 3.40 & 17.86 & 54.40 $\pm$ 4.40 & 17.34 & 37.80 $\pm$ 4.20 & 17.40 & 27.70 $\pm$ 4.14 \\
 & 256 & 21.80 & 44.60 $\pm$ 4.30 & 21.50 & 79.00 $\pm$ 3.50 & 22.37 & 59.60 $\pm$ 4.20 & 21.60 & 36.20 $\pm$ 4.30 & 21.34 & 28.98 $\pm$ 4.33 \\
 & 512 & 28.56 & 46.40 $\pm$ 4.40 & 28.12 & 79.00 $\pm$ 3.50 & 29.39 & 59.40 $\pm$ 4.30 & 28.25 & 36.60 $\pm$ 4.20 & 27.55 & 30.30 $\pm$ 4.35 \\
 & 1024 & 38.43 & 45.00 $\pm$ 4.30 & 37.95 & 81.00 $\pm$ 3.40 & 39.60 & 61.00 $\pm$ 4.30 & 37.97 & 38.00 $\pm$ 4.40 &  &  \\
 & 2048 & 52.02 & 47.20 $\pm$ 4.40 & 51.41 & 81.40 $\pm$ 3.50 & 53.65 & 62.20 $\pm$ 4.30 & 51.44 & 37.00 $\pm$ 4.20 &  &  \\
\midrule
\multirow{5}{*}{\shortstack{SANTA++\\(Contiguous)}} & 128 & 15.66 & 42.80 $\pm$ 4.40 & 15.51 & 75.60 $\pm$ 3.80 & 15.90 & 58.60 $\pm$ 4.40 & 15.58 & 35.40 $\pm$ 4.20 & 15.87 & 17.56 $\pm$ 3.39 \\
 & 256 & 18.46 & 42.60 $\pm$ 4.30 & 18.23 & 77.20 $\pm$ 3.80 & 18.89 & 61.00 $\pm$ 4.30 & 18.36 & 36.80 $\pm$ 4.20 & 18.50 & 23.52 $\pm$ 3.87 \\
 & 512 & 23.29 & 44.60 $\pm$ 4.30 & 22.92 & 76.20 $\pm$ 3.70 & 24.03 & 60.20 $\pm$ 4.30 & 23.14 & 37.20 $\pm$ 4.20 & 23.03 & 24.69 $\pm$ 4.03 \\
 & 1024 & 31.09 & 46.00 $\pm$ 4.40 & 30.66 & 78.60 $\pm$ 3.60 & 32.27 & 61.00 $\pm$ 4.20 & 30.88 & 38.40 $\pm$ 4.30 &  &  \\
 & 2048 & 43.15 & 46.40 $\pm$ 4.40 & 42.50 & 78.00 $\pm$ 3.60 & 44.91 & 62.00 $\pm$ 4.30 & 42.80 & 37.00 $\pm$ 4.20 &  &  \\
\bottomrule
\end{tabular}
}
\vspace{0.8em}
\resizebox{\textwidth}{!}{
\begin{tabular}{llrrrrrrrrrr}
\toprule
\multirow{2}{*}{Method} & \multirow{2}{*}{$S$} & \multicolumn{2}{c}{TREC coarse} & \multicolumn{2}{c}{TREC fine} & \multicolumn{2}{c}{NLU} & \multicolumn{2}{c}{Banking77} & \multicolumn{2}{c}{CLINC150} \\
 &  & Total access (\%) & Acc (\%) & Total access (\%) & Acc (\%) & Total access (\%) & Acc (\%) & Total access (\%) & Acc (\%) & Total access (\%) & Acc (\%) \\
\midrule
\multirow{1}{*}{SDPA} & -- & 100.00 & 83.20 $\pm$ 3.30 & 100.00 & 50.80 $\pm$ 4.40 & 100.00 & 86.20 $\pm$ 3.00 & 100.00 & 83.40 $\pm$ 3.30 & 100.00 & 89.80 $\pm$ 2.60 \\
\midrule
\multirow{5}{*}{SANTA} & 128 & 50.61 & 83.20 $\pm$ 3.30 & 50.59 & 48.80 $\pm$ 4.20 & 50.58 & 85.00 $\pm$ 3.10 & 50.58 & 83.20 $\pm$ 3.30 & 50.56 & 89.00 $\pm$ 2.90 \\
 & 256 & 51.07 & 83.60 $\pm$ 3.30 & 51.03 & 50.80 $\pm$ 4.30 & 51.02 & 86.20 $\pm$ 3.00 & 51.01 & 83.20 $\pm$ 3.20 & 51.00 & 90.80 $\pm$ 2.50 \\
 & 512 & 51.84 & 83.60 $\pm$ 3.20 & 51.78 & 50.80 $\pm$ 4.30 & 51.77 & 84.60 $\pm$ 3.20 & 51.72 & 82.40 $\pm$ 3.30 & 51.72 & 89.60 $\pm$ 2.70 \\
\midrule
\multirow{5}{*}{\shortstack{SANTA++\\(K-means)}} & 128 & 17.95 & 81.60 $\pm$ 3.50 & 17.79 & 46.60 $\pm$ 4.50 & 17.52 & 80.80 $\pm$ 3.40 & 17.31 & 83.00 $\pm$ 3.30 & 17.46 & 86.40 $\pm$ 3.00 \\
 & 256 & 22.50 & 84.40 $\pm$ 3.10 & 22.19 & 47.80 $\pm$ 4.40 & 21.85 & 85.20 $\pm$ 3.20 & 21.47 & 83.40 $\pm$ 3.30 & 21.69 & 89.60 $\pm$ 2.60 \\
 & 512 & 29.57 & 85.00 $\pm$ 3.20 & 29.03 & 50.80 $\pm$ 4.40 & 28.66 & 84.40 $\pm$ 3.10 & 28.05 & 81.40 $\pm$ 3.40 & 28.31 & 88.20 $\pm$ 2.80 \\
\midrule
\multirow{5}{*}{\shortstack{SANTA++\\(Contiguous)}} & 128 & 15.74 & 83.80 $\pm$ 3.20 & 15.61 & 41.60 $\pm$ 4.40 & 15.64 & 78.40 $\pm$ 3.70 & 15.55 & 78.60 $\pm$ 3.60 & 15.56 & 83.60 $\pm$ 3.30 \\
 & 256 & 18.66 & 85.60 $\pm$ 3.00 & 18.41 & 44.00 $\pm$ 4.50 & 18.49 & 80.40 $\pm$ 3.40 & 18.30 & 81.00 $\pm$ 3.50 & 18.30 & 86.40 $\pm$ 3.00 \\
 & 512 & 23.71 & 85.80 $\pm$ 3.00 & 23.28 & 46.20 $\pm$ 4.30 & 23.43 & 82.60 $\pm$ 3.40 & 23.07 & 81.60 $\pm$ 3.30 & 23.07 & 87.80 $\pm$ 2.80 \\
\bottomrule
\end{tabular}
}
\vspace{0.8em}
\resizebox{\textwidth}{!}{
\begin{tabular}{llrrrrrrrr}
\toprule
\multirow{2}{*}{Method} & \multirow{2}{*}{$S$} & \multicolumn{2}{c}{InfBench QA} & \multicolumn{2}{c}{InfBench MC} & \multicolumn{2}{c}{JSON KV} & \multicolumn{2}{c}{All tasks average} \\
 &  & Total access (\%) & Acc (\%) & Total access (\%) & Acc (\%) & Total access (\%) & Acc (\%) & Total access (\%) & Acc (\%) \\
\midrule
\multirow{1}{*}{SDPA} & -- & 100.00 & 15.00 $\pm$ 2.54 & 100.00 & 51.09 $\pm$ 6.55 & 100.00 & 96.80 $\pm$ 1.50 & 100.00 & 63.10 \\
\midrule
\multirow{5}{*}{SANTA} & 128 & 50.36 & 13.87 $\pm$ 2.36 & 50.28 & 46.29 $\pm$ 6.55 & 50.36 & 96.40 $\pm$ 1.70 & 50.47 & 62.38 \\
 & 256 & 50.62 & 14.80 $\pm$ 2.55 & 50.48 & 48.47 $\pm$ 6.55 & 50.60 & 96.80 $\pm$ 1.50 & 50.82 & 62.91 \\
 & 512 & 51.07 & 14.52 $\pm$ 2.51 & 50.79 & 48.47 $\pm$ 6.55 & 51.00 & 96.80 $\pm$ 1.50 & 51.40 & 62.71 \\
\midrule
\multirow{5}{*}{\shortstack{SANTA++\\(K-means)}} & 128 & 16.97 & 13.40 $\pm$ 2.28 & 16.42 & 50.66 $\pm$ 6.55 & 15.87 & 45.60 $\pm$ 4.50 & 17.28 & 56.35 \\
 & 256 & 21.14 & 14.45 $\pm$ 2.54 & 20.51 & 48.91 $\pm$ 6.55 & 19.01 & 70.20 $\pm$ 4.10 & 21.46 & 59.41 \\
 & 512 & 27.80 & 13.73 $\pm$ 2.42 & 27.14 & 48.47 $\pm$ 6.55 & 24.35 & 87.00 $\pm$ 3.00 & 28.06 & 60.82 \\
\midrule
\multirow{5}{*}{\shortstack{SANTA++\\(Contiguous)}} & 128 & 15.27 & 11.92 $\pm$ 2.05 & 14.94 & 46.72 $\pm$ 6.55 & 15.00 & 33.40 $\pm$ 4.10 & 15.53 & 52.92 \\
 & 256 & 17.85 & 12.61 $\pm$ 2.27 & 17.41 & 48.47 $\pm$ 6.55 & 17.18 & 54.60 $\pm$ 4.40 & 18.23 & 56.48 \\
 & 512 & 22.39 & 13.54 $\pm$ 2.30 & 21.95 & 48.91 $\pm$ 6.55 & 21.20 & 68.60 $\pm$ 4.00 & 22.96 & 58.30 \\
\bottomrule
\end{tabular}
}
\caption{HELMET task results at 32K context length.
{For individual tasks, $\pm$ denotes the half-width of a 95\%
bootstrap confidence interval. ``All tasks average'' is the equal-weight
mean of the 13 task scores.}}

\label{tab:helmet-32k-task}

\end{table*}

\FloatBarrier

\subsection{LongBench v2}
\label{app:longbench}

Table~\ref{tab:longbench_results} reports the complete aggregate budget sweep on 503 distinct LongBench v2 prompts using Qwen2.5-7B-Instruct. We use the repository's middle-truncation scheme with a 32K context cap. {Prompts longer than the cap are middle-truncated; shorter prompts retain their original length.}

\paragraph{Prompting.}
Each prompt begins with Qwen's assistant preamble. {We request brief reasoning and a fixed answer format by appending:}
\begin{quote}
\small
Keep your reasoning concise (under 3 sentences). There is exactly one correct option. On the final line, provide your answer in the exact format: 'Final Answer: [X]' where X is a single letter (A, B, C, or D).
\end{quote}

\paragraph{{Runs, scores, and confidence intervals.}}
{Each setting has four complete runs of the same 503 prompts.} {Invalid
or unparseable answers receive zero credit.} {We average the four binary
outcomes within each prompt, then average those 503 prompt means.}
{For 95\% confidence intervals, we resample prompts 10,000 times,
keeping their four outcomes together.} {Normalized-score intervals use
the same sampled prompt indices for the sparse method and SDPA.}

\paragraph{{Budget sensitivity.}}
{Increasing the budget changes the sampled teams and their inclusion
corrections. These changes affect the normalized attention output and
can alter later generated tokens, so task accuracy need not improve
monotonically. LongBench shows little consistent benefit from larger
budgets, while aggregate RULER-32K scores increase across both grouping
sweeps (Tables~\ref{tab:longbench_results}
and~\ref{tab:ruler-budget-main}).}

\FloatBarrier

\subsection{RULER}
\label{app:ruler}

{Task abbreviations follow RULER~\citep{hsieh2024ruler}:
NIAH denotes needle-in-a-haystack retrieval, with single-needle
(S-NIAH), multiple-key (MK-NIAH), multiple-query (MQ-NIAH), and
multiple-value (MV-NIAH) variants. VT denotes variable tracking;
CWE and FWE denote common words extraction and frequent words
extraction; QA denotes question answering. Numbered suffixes identify
benchmark variants.}

Table~\ref{tab:ruler_32k} reports the full 32K aggregate sweep.
Tables~\ref{tab:ruler_by_task_8k} and
\ref{tab:ruler_by_task_32k} give all 13 tasks: S-NIAH-1/2/3,
MK-NIAH-1/2/3, MQ-NIAH, MV-NIAH, VT, CWE, FWE, QA-1, and QA-2.
The main RULER evaluations use the reference repository's task
definitions, prompting, and grading conventions with
Qwen2.5-7B-Instruct. At both 8K and 32K, all 13 tasks have 500
completed prompts per task. Smaller diagnostic and MLA runs are
separate and are not pooled into these results.

Each task has its own reported GQA-aware KV percentage. The main RULER
results use the total-access measure in
Equation~\ref{eq:qwen-logical-access}, which includes both selected GQA
KV reads and representative-key reads used for routing. Aggregate access
pools the underlying read counts across prompts and tasks before division
rather than averaging the displayed task percentages.

\begin{table}[!htbp]
\centering
\small
\caption{\textbf{RULER: full 32K budget sweep with Qwen2.5-7B-Instruct.}
Score (abs) is the equal-weight mean of all 13 task scores over 500
completed prompts per task. Score (norm) is the absolute score divided
by the corresponding SDPA score at the same context length, expressed
as a percentage. KV is pooled GQA-aware logical access, including representative-key
routing reads, and $S$ is the nominal budget. Absolute-score $\pm$
entries are 95\% task-stratified bootstrap confidence interval
half-widths.}
\label{tab:ruler}
\label{tab:ruler_32k}
\begin{tabular*}{\linewidth}{@{\extracolsep{\fill}}lrrrr@{}}
\toprule
Method & $S$ & KV (\%) & Score (norm) (\%) & Score (abs) (\%) \\
\midrule
SDPA & --- & 100.00 & 100.00 & 83.51 $\pm$ 0.66 \\
\midrule
\multirow{6}{*}{SANTA} & 128 & 50.32 & 99.15 & 82.80 $\pm$ 0.64 \\
 & 256 & 50.54 & 99.75 & 83.31 $\pm$ 0.67 \\
 & 512 & 50.91 & 100.11 & 83.61 $\pm$ 0.65 \\
 & 1024 & 51.52 & 99.92 & 83.45 $\pm$ 0.67 \\
 & 2048 & 52.49 & 99.99 & 83.50 $\pm$ 0.66 \\
 & 4096 & 53.96 & 99.91 & 83.44 $\pm$ 0.67 \\
\midrule
\multirow{6}{*}{\shortstack[l]{SANTA\texttt{++}\\(k-means)}} & 128 & 16.38 & 85.41 & 71.33 $\pm$ 0.74 \\
 & 256 & 19.95 & 91.07 & 76.05 $\pm$ 0.74 \\
 & 512 & 25.70 & 93.99 & 78.49 $\pm$ 0.74 \\
 & 1024 & 34.44 & 96.27 & 80.40 $\pm$ 0.72 \\
 & 2048 & 46.95 & 98.54 & 82.29 $\pm$ 0.66 \\
 & 4096 & 63.65 & 99.48 & 83.08 $\pm$ 0.64 \\
\midrule
\multirow{6}{*}{\shortstack[l]{SANTA\texttt{++}\\(contiguous)}} & 128 & 15.03 & 77.50 & 64.72 $\pm$ 0.77 \\
 & 256 & 17.31 & 84.46 & 70.54 $\pm$ 0.76 \\
 & 512 & 21.37 & 90.25 & 75.37 $\pm$ 0.75 \\
 & 1024 & 28.16 & 95.07 & 79.40 $\pm$ 0.73 \\
 & 2048 & 38.89 & 97.49 & 81.42 $\pm$ 0.72 \\
 & 4096 & 54.68 & 98.86 & 82.56 $\pm$ 0.67 \\
\bottomrule
\end{tabular*}
\end{table}
\FloatBarrier

\begin{table}[!htb]
\tiny
\setlength{\tabcolsep}{3pt}
\begin{tabular}{l c c c c c c c c c c c}
\toprule
 &  & \multicolumn{2}{c}{S-NIAH-1} & \multicolumn{2}{c}{S-NIAH-2} & \multicolumn{2}{c}{S-NIAH-3} & \multicolumn{2}{c}{MK-NIAH-1} & \multicolumn{2}{c}{MK-NIAH-2} \\
\cmidrule(lr){3-4}\cmidrule(lr){5-6}\cmidrule(lr){7-8}\cmidrule(lr){9-10}\cmidrule(lr){11-12}
Method & $S$ & KV (\%) & Acc (\%) & KV (\%) & Acc (\%) & KV (\%) & Acc (\%) & KV (\%) & Acc (\%) & KV (\%) & Acc (\%) \\
\midrule
SDPA & --- & 100.00 & 100.00 $\pm$ 0.00 & 100.00 & 100.00 $\pm$ 0.00 & 100.00 & 100.00 $\pm$ 0.00 & 100.00 & 100.00 $\pm$ 0.00 & 100.00 & 98.80 $\pm$ 0.90 \\
\midrule
\multirow{4}{*}{SANTA} & 128 & 50.97 & 100.00 $\pm$ 0.00 & 51.11 & 100.00 $\pm$ 0.00 & 51.06 & 99.80 $\pm$ 0.30 & 51.13 & 100.00 $\pm$ 0.00 & 51.25 & 98.80 $\pm$ 0.90 \\
 & 256 & 51.56 & 100.00 $\pm$ 0.00 & 51.81 & 100.00 $\pm$ 0.00 & 51.72 & 99.80 $\pm$ 0.30 & 51.83 & 100.00 $\pm$ 0.00 & 52.08 & 99.00 $\pm$ 0.90 \\
 & 512 & 52.52 & 100.00 $\pm$ 0.00 & 52.93 & 100.00 $\pm$ 0.00 & 52.76 & 100.00 $\pm$ 0.00 & 52.94 & 100.00 $\pm$ 0.00 & 53.41 & 98.20 $\pm$ 1.10 \\
 & 1024 & 53.98 & 100.00 $\pm$ 0.00 & 54.62 & 100.00 $\pm$ 0.00 & 54.35 & 100.00 $\pm$ 0.00 & 54.62 & 100.00 $\pm$ 0.00 & 55.43 & 98.20 $\pm$ 1.10 \\
\midrule
\multirow{4}{*}{\shortstack{\textbf{SANTA\texttt{++}}\\(K-means)}} & 128 & 22.53 & 100.00 $\pm$ 0.00 & 25.43 & 99.40 $\pm$ 0.80 & 25.46 & 98.00 $\pm$ 1.20 & 24.95 & 98.40 $\pm$ 1.10 & 25.57 & 96.00 $\pm$ 1.60 \\
 & 256 & 30.37 & 100.00 $\pm$ 0.00 & 34.75 & 99.60 $\pm$ 0.50 & 34.63 & 99.60 $\pm$ 0.50 & 34.04 & 99.80 $\pm$ 0.30 & 34.59 & 98.40 $\pm$ 1.10 \\
 & 512 & 42.54 & 100.00 $\pm$ 0.00 & 48.54 & 100.00 $\pm$ 0.00 & 48.16 & 100.00 $\pm$ 0.00 & 47.57 & 100.00 $\pm$ 0.00 & 47.76 & 97.80 $\pm$ 1.20 \\
 & 1024 & 60.33 & 100.00 $\pm$ 0.00 & 66.64 & 100.00 $\pm$ 0.00 & 66.20 & 100.00 $\pm$ 0.00 & 65.69 & 100.00 $\pm$ 0.00 & 65.51 & 97.60 $\pm$ 1.30 \\
\midrule
\multirow{4}{*}{\shortstack{\textbf{SANTA\texttt{++}}\\(Contiguous)}} & 128 & 20.10 & 100.00 $\pm$ 0.00 & 21.99 & 99.20 $\pm$ 0.70 & 22.46 & 98.40 $\pm$ 1.10 & 21.78 & 99.00 $\pm$ 0.90 & 22.59 & 91.40 $\pm$ 2.50 \\
 & 256 & 27.08 & 100.00 $\pm$ 0.00 & 29.86 & 99.00 $\pm$ 0.90 & 30.13 & 99.60 $\pm$ 0.60 & 29.38 & 98.60 $\pm$ 1.10 & 30.71 & 96.20 $\pm$ 1.60 \\
 & 512 & 38.72 & 100.00 $\pm$ 0.00 & 42.36 & 100.00 $\pm$ 0.00 & 42.49 & 99.80 $\pm$ 0.30 & 41.68 & 100.00 $\pm$ 0.00 & 43.34 & 97.40 $\pm$ 1.40 \\
 & 1024 & 55.97 & 100.00 $\pm$ 0.00 & 60.42 & 100.00 $\pm$ 0.00 & 60.36 & 99.80 $\pm$ 0.40 & 59.62 & 100.00 $\pm$ 0.00 & 60.94 & 98.00 $\pm$ 1.30 \\
\bottomrule
\end{tabular}
\vspace{2pt}
\begin{tabular}{l c c c c c c c c c}
\toprule
 &  & \multicolumn{2}{c}{MK-NIAH-3} & \multicolumn{2}{c}{MQ-NIAH} & \multicolumn{2}{c}{MV-NIAH} & \multicolumn{2}{c}{VT} \\
\cmidrule(lr){3-4}\cmidrule(lr){5-6}\cmidrule(lr){7-8}\cmidrule(lr){9-10}
Method & $S$ & KV (\%) & Acc (\%) & KV (\%) & Acc (\%) & KV (\%) & Acc (\%) & KV (\%) & Acc (\%) \\
\midrule
SDPA & --- & 100.00 & 97.00 $\pm$ 1.40 & 100.00 & 99.90 $\pm$ 0.12 & 100.00 & 97.75 $\pm$ 0.65 & 100.00 & 30.60 $\pm$ 3.78 \\
\midrule
\multirow{4}{*}{SANTA} & 128 & 51.19 & 96.40 $\pm$ 1.50 & 51.19 & 99.90 $\pm$ 0.12 & 51.21 & 97.25 $\pm$ 0.70 & 51.05 & 28.44 $\pm$ 3.84 \\
 & 256 & 51.94 & 96.20 $\pm$ 1.60 & 51.90 & 99.85 $\pm$ 0.17 & 51.96 & 97.20 $\pm$ 0.70 & 51.63 & 29.52 $\pm$ 3.80 \\
 & 512 & 53.10 & 96.40 $\pm$ 1.50 & 53.00 & 99.90 $\pm$ 0.12 & 53.12 & 97.50 $\pm$ 0.70 & 52.50 & 30.56 $\pm$ 3.96 \\
 & 1024 & 54.85 & 96.60 $\pm$ 1.50 & 54.64 & 99.85 $\pm$ 0.17 & 54.86 & 97.50 $\pm$ 0.65 & 53.76 & 31.84 $\pm$ 4.04 \\
\midrule
\multirow{4}{*}{\shortstack{\textbf{SANTA\texttt{++}}\\(K-means)}} & 128 & 23.80 & 87.60 $\pm$ 2.80 & 25.59 & 99.00 $\pm$ 0.48 & 25.78 & 92.30 $\pm$ 1.12 & 22.11 & 24.56 $\pm$ 3.64 \\
 & 256 & 31.71 & 94.80 $\pm$ 1.90 & 34.51 & 99.75 $\pm$ 0.23 & 34.86 & 94.85 $\pm$ 0.93 & 29.16 & 32.00 $\pm$ 4.00 \\
 & 512 & 44.02 & 96.80 $\pm$ 1.40 & 47.92 & 99.85 $\pm$ 0.17 & 48.33 & 96.05 $\pm$ 0.82 & 40.39 & 35.52 $\pm$ 4.20 \\
 & 1024 & 61.63 & 96.80 $\pm$ 1.40 & 65.91 & 99.85 $\pm$ 0.17 & 66.35 & 96.80 $\pm$ 0.73 & 56.94 & 30.60 $\pm$ 3.94 \\
\midrule
\multirow{4}{*}{\shortstack{\textbf{SANTA\texttt{++}}\\(Contiguous)}} & 128 & 21.97 & 42.40 $\pm$ 4.30 & 22.37 & 97.80 $\pm$ 0.68 & 22.47 & 87.50 $\pm$ 1.38 & 19.52 & 17.12 $\pm$ 3.34 \\
 & 256 & 29.09 & 80.40 $\pm$ 3.30 & 29.90 & 98.65 $\pm$ 0.48 & 30.14 & 90.65 $\pm$ 1.20 & 25.46 & 21.80 $\pm$ 3.38 \\
 & 512 & 40.49 & 93.00 $\pm$ 2.10 & 42.02 & 99.70 $\pm$ 0.23 & 42.43 & 93.35 $\pm$ 1.05 & 35.86 & 25.32 $\pm$ 3.76 \\
 & 1024 & 57.42 & 95.60 $\pm$ 1.80 & 59.76 & 99.85 $\pm$ 0.17 & 60.26 & 95.90 $\pm$ 0.83 & 51.86 & 28.60 $\pm$ 3.98 \\
\bottomrule
\end{tabular}
\vspace{2pt}
\begin{tabular}{l c c c c c c c c c}
\toprule
 &  & \multicolumn{2}{c}{CWE} & \multicolumn{2}{c}{FWE} & \multicolumn{2}{c}{QA-1} & \multicolumn{2}{c}{QA-2} \\
\cmidrule(lr){3-4}\cmidrule(lr){5-6}\cmidrule(lr){7-8}\cmidrule(lr){9-10}
Method & $S$ & KV (\%) & Acc (\%) & KV (\%) & Acc (\%) & KV (\%) & Acc (\%) & KV (\%) & Acc (\%) \\
\midrule
SDPA & --- & 100.00 & 93.86 $\pm$ 0.86 & 100.00 & 81.60 $\pm$ 1.53 & 100.00 & 74.60 $\pm$ 3.80 & 100.00 & 55.20 $\pm$ 4.50 \\
\midrule
\multirow{4}{*}{SANTA} & 128 & 51.27 & 91.62 $\pm$ 0.99 & 51.35 & 82.33 $\pm$ 1.57 & 51.47 & 75.00 $\pm$ 4.00 & 51.28 & 54.60 $\pm$ 4.30 \\
 & 256 & 52.11 & 93.32 $\pm$ 0.86 & 52.26 & 81.80 $\pm$ 1.53 & 52.45 & 74.40 $\pm$ 3.80 & 52.11 & 55.20 $\pm$ 4.30 \\
 & 512 & 53.42 & 93.92 $\pm$ 0.89 & 53.70 & 81.93 $\pm$ 1.53 & 54.02 & 74.40 $\pm$ 3.80 & 53.43 & 55.40 $\pm$ 4.50 \\
 & 1024 & 55.39 & 94.04 $\pm$ 0.76 & 55.89 & 81.53 $\pm$ 1.50 & 56.39 & 74.40 $\pm$ 3.80 & 55.42 & 56.20 $\pm$ 4.30 \\
\midrule
\multirow{4}{*}{\shortstack{\textbf{SANTA\texttt{++}}\\(K-means)}} & 128 & 24.85 & 89.80 $\pm$ 0.90 & 23.62 & 80.33 $\pm$ 1.53 & 28.35 & 72.20 $\pm$ 3.80 & 26.61 & 52.60 $\pm$ 4.50 \\
 & 256 & 33.40 & 92.10 $\pm$ 0.95 & 32.10 & 81.33 $\pm$ 1.63 & 38.81 & 74.00 $\pm$ 3.90 & 36.31 & 52.00 $\pm$ 4.30 \\
 & 512 & 46.19 & 92.72 $\pm$ 0.94 & 45.33 & 81.67 $\pm$ 1.47 & 53.51 & 73.60 $\pm$ 3.80 & 50.27 & 53.40 $\pm$ 4.40 \\
 & 1024 & 63.64 & 93.16 $\pm$ 0.98 & 63.78 & 82.13 $\pm$ 1.50 & 72.14 & 75.20 $\pm$ 3.80 & 68.31 & 54.60 $\pm$ 4.40 \\
\midrule
\multirow{4}{*}{\shortstack{\textbf{SANTA\texttt{++}}\\(Contiguous)}} & 128 & 21.95 & 73.92 $\pm$ 1.28 & 20.99 & 74.33 $\pm$ 1.73 & 24.45 & 68.80 $\pm$ 4.20 & 23.15 & 51.40 $\pm$ 4.20 \\
 & 256 & 29.34 & 82.72 $\pm$ 1.17 & 28.40 & 78.13 $\pm$ 1.73 & 33.65 & 72.40 $\pm$ 3.80 & 31.60 & 53.80 $\pm$ 4.40 \\
 & 512 & 41.31 & 89.38 $\pm$ 0.98 & 40.66 & 80.60 $\pm$ 1.63 & 47.67 & 73.60 $\pm$ 3.90 & 44.56 & 53.40 $\pm$ 4.20 \\
 & 1024 & 58.57 & 90.72 $\pm$ 0.95 & 58.84 & 81.13 $\pm$ 1.50 & 66.71 & 74.60 $\pm$ 3.80 & 62.86 & 54.40 $\pm$ 4.40 \\
\bottomrule
\end{tabular}
\caption{\textbf{Per-task accuracy and KV memory access on RULER (8K context) on Qwen2.5-7B-Instruct.} \textit{KV} is the theoretical GQA-aware KV memory access (\%), and $S$ denotes the nominal budget. Accuracy values are reported as mean $\pm$ 95\% bootstrap confidence interval half-width.}
\label{tab:ruler_by_task_8k}
\end{table}

\begin{table}[!htb]
\tiny
\setlength{\tabcolsep}{3pt}
\begin{tabular}{l c c c c c c c c c c c}
\toprule
 &  & \multicolumn{2}{c}{S-NIAH-1} & \multicolumn{2}{c}{S-NIAH-2} & \multicolumn{2}{c}{S-NIAH-3} & \multicolumn{2}{c}{MK-NIAH-1} & \multicolumn{2}{c}{MK-NIAH-2} \\
\cmidrule(lr){3-4}\cmidrule(lr){5-6}\cmidrule(lr){7-8}\cmidrule(lr){9-10}\cmidrule(lr){11-12}
Method & $S$ & KV (\%) & Acc (\%) & KV (\%) & Acc (\%) & KV (\%) & Acc (\%) & KV (\%) & Acc (\%) & KV (\%) & Acc (\%) \\
\midrule
SDPA & --- & 100.00 & 100.00 $\pm$ 0.00 & 100.00 & 100.00 $\pm$ 0.00 & 100.00 & 98.60 $\pm$ 1.00 & 100.00 & 99.00 $\pm$ 0.90 & 100.00 & 93.40 $\pm$ 2.20 \\
\midrule
\multirow{6}{*}{SANTA} & 128 & 50.28 & 100.00 $\pm$ 0.00 & 50.31 & 100.00 $\pm$ 0.00 & 50.30 & 99.20 $\pm$ 0.70 & 50.31 & 99.20 $\pm$ 0.70 & 50.37 & 93.20 $\pm$ 2.20 \\
 & 256 & 50.47 & 100.00 $\pm$ 0.00 & 50.53 & 100.00 $\pm$ 0.00 & 50.51 & 98.80 $\pm$ 0.90 & 50.53 & 99.00 $\pm$ 0.90 & 50.64 & 92.80 $\pm$ 2.20 \\
 & 512 & 50.81 & 100.00 $\pm$ 0.00 & 50.91 & 100.00 $\pm$ 0.00 & 50.87 & 98.80 $\pm$ 0.90 & 50.90 & 99.00 $\pm$ 0.90 & 51.12 & 93.60 $\pm$ 2.10 \\
 & 1024 & 51.37 & 100.00 $\pm$ 0.00 & 51.53 & 100.00 $\pm$ 0.00 & 51.45 & 98.60 $\pm$ 1.00 & 51.51 & 99.00 $\pm$ 0.90 & 51.92 & 93.40 $\pm$ 2.10 \\
 & 2048 & 52.28 & 100.00 $\pm$ 0.00 & 52.54 & 100.00 $\pm$ 0.00 & 52.39 & 98.60 $\pm$ 1.00 & 52.49 & 99.00 $\pm$ 0.90 & 53.17 & 93.40 $\pm$ 2.20 \\
 & 4096 & 53.69 & 100.00 $\pm$ 0.00 & 54.03 & 100.00 $\pm$ 0.00 & 53.82 & 98.60 $\pm$ 1.10 & 53.97 & 99.00 $\pm$ 0.90 & 55.02 & 93.60 $\pm$ 2.10 \\
\midrule
\multirow{6}{*}{\shortstack{\textbf{SANTA\texttt{++}}\\(K-means)}} & 128 & 15.65 & 97.40 $\pm$ 1.30 & 16.82 & 98.80 $\pm$ 0.90 & 16.79 & 96.40 $\pm$ 1.70 & 16.69 & 95.40 $\pm$ 1.70 & 16.55 & 77.00 $\pm$ 3.90 \\
 & 256 & 18.71 & 98.80 $\pm$ 0.90 & 20.64 & 99.40 $\pm$ 0.70 & 20.56 & 98.60 $\pm$ 1.00 & 20.50 & 95.80 $\pm$ 1.80 & 20.31 & 87.00 $\pm$ 3.00 \\
 & 512 & 23.77 & 100.00 $\pm$ 0.00 & 26.65 & 99.60 $\pm$ 0.50 & 26.59 & 99.00 $\pm$ 0.90 & 26.49 & 96.60 $\pm$ 1.60 & 26.24 & 88.20 $\pm$ 2.90 \\
 & 1024 & 31.77 & 100.00 $\pm$ 0.00 & 35.71 & 100.00 $\pm$ 0.00 & 35.62 & 99.20 $\pm$ 0.70 & 35.51 & 98.40 $\pm$ 1.00 & 35.17 & 89.60 $\pm$ 2.80 \\
 & 2048 & 43.85 & 100.00 $\pm$ 0.00 & 48.37 & 100.00 $\pm$ 0.00 & 48.33 & 99.40 $\pm$ 0.70 & 48.27 & 99.00 $\pm$ 0.90 & 47.90 & 92.60 $\pm$ 2.30 \\
 & 4096 & 61.16 & 100.00 $\pm$ 0.00 & 64.94 & 100.00 $\pm$ 0.00 & 64.89 & 99.20 $\pm$ 0.70 & 64.85 & 99.00 $\pm$ 0.90 & 64.80 & 93.40 $\pm$ 2.20 \\
\midrule
\multirow{6}{*}{\shortstack{\textbf{SANTA\texttt{++}}\\(Contiguous)}} & 128 & 14.63 & 99.40 $\pm$ 0.70 & 15.08 & 91.40 $\pm$ 2.40 & 15.16 & 86.80 $\pm$ 3.00 & 15.01 & 90.20 $\pm$ 2.60 & 15.19 & 47.40 $\pm$ 4.80 \\
 & 256 & 16.74 & 99.80 $\pm$ 0.30 & 17.42 & 96.60 $\pm$ 1.60 & 17.53 & 94.00 $\pm$ 2.00 & 17.33 & 94.40 $\pm$ 1.90 & 17.72 & 65.20 $\pm$ 4.20 \\
 & 512 & 20.66 & 99.80 $\pm$ 0.30 & 21.55 & 98.20 $\pm$ 1.10 & 21.69 & 97.60 $\pm$ 1.30 & 21.42 & 97.00 $\pm$ 1.50 & 22.15 & 80.20 $\pm$ 3.30 \\
 & 1024 & 27.40 & 100.00 $\pm$ 0.00 & 28.47 & 99.60 $\pm$ 0.50 & 28.63 & 98.80 $\pm$ 1.10 & 28.32 & 98.60 $\pm$ 0.90 & 29.43 & 88.20 $\pm$ 2.80 \\
 & 2048 & 38.21 & 100.00 $\pm$ 0.00 & 39.36 & 100.00 $\pm$ 0.00 & 39.55 & 99.00 $\pm$ 0.90 & 39.17 & 99.00 $\pm$ 0.90 & 40.66 & 91.60 $\pm$ 2.40 \\
 & 4096 & 54.15 & 100.00 $\pm$ 0.00 & 55.24 & 100.00 $\pm$ 0.00 & 55.45 & 99.60 $\pm$ 0.50 & 55.14 & 99.40 $\pm$ 0.70 & 56.81 & 93.20 $\pm$ 2.20 \\
\bottomrule
\end{tabular}
\vspace{2pt}
\begin{tabular}{l c c c c c c c c c}
\toprule
 &  & \multicolumn{2}{c}{MK-NIAH-3} & \multicolumn{2}{c}{MQ-NIAH} & \multicolumn{2}{c}{MV-NIAH} & \multicolumn{2}{c}{VT} \\
\cmidrule(lr){3-4}\cmidrule(lr){5-6}\cmidrule(lr){7-8}\cmidrule(lr){9-10}
Method & $S$ & KV (\%) & Acc (\%) & KV (\%) & Acc (\%) & KV (\%) & Acc (\%) & KV (\%) & Acc (\%) \\
\midrule
SDPA & --- & 100.00 & 89.60 $\pm$ 2.70 & 100.00 & 99.40 $\pm$ 0.33 & 100.00 & 90.35 $\pm$ 1.17 & 100.00 & 33.32 $\pm$ 3.58 \\
\midrule
\multirow{6}{*}{SANTA} & 128 & 50.33 & 88.40 $\pm$ 2.80 & 50.31 & 99.25 $\pm$ 0.38 & 50.32 & 89.00 $\pm$ 1.25 & 50.28 & 30.64 $\pm$ 3.72 \\
 & 256 & 50.56 & 89.40 $\pm$ 2.50 & 50.52 & 99.45 $\pm$ 0.33 & 50.54 & 89.60 $\pm$ 1.23 & 50.46 & 34.20 $\pm$ 3.66 \\
 & 512 & 50.94 & 89.60 $\pm$ 2.60 & 50.87 & 99.40 $\pm$ 0.33 & 50.90 & 90.25 $\pm$ 1.17 & 50.75 & 33.88 $\pm$ 3.78 \\
 & 1024 & 51.57 & 90.00 $\pm$ 2.50 & 51.43 & 99.30 $\pm$ 0.35 & 51.51 & 90.25 $\pm$ 1.18 & 51.21 & 31.80 $\pm$ 3.60 \\
 & 2048 & 52.57 & 89.60 $\pm$ 2.60 & 52.34 & 99.45 $\pm$ 0.30 & 52.47 & 90.25 $\pm$ 1.13 & 51.95 & 32.04 $\pm$ 3.60 \\
 & 4096 & 54.08 & 89.40 $\pm$ 2.70 & 53.72 & 99.35 $\pm$ 0.35 & 53.94 & 90.25 $\pm$ 1.18 & 53.06 & 33.04 $\pm$ 3.82 \\
\midrule
\multirow{6}{*}{\shortstack{\textbf{SANTA\texttt{++}}\\(K-means)}} & 128 & 15.91 & 16.80 $\pm$ 3.20 & 16.63 & 94.20 $\pm$ 1.10 & 16.83 & 81.60 $\pm$ 1.63 & 15.28 & 24.00 $\pm$ 3.44 \\
 & 256 & 19.00 & 45.00 $\pm$ 4.50 & 20.34 & 97.65 $\pm$ 0.67 & 20.58 & 85.05 $\pm$ 1.50 & 18.10 & 27.52 $\pm$ 3.64 \\
 & 512 & 24.16 & 61.80 $\pm$ 4.20 & 26.33 & 98.25 $\pm$ 0.60 & 26.58 & 86.65 $\pm$ 1.25 & 22.85 & 28.60 $\pm$ 3.44 \\
 & 1024 & 32.24 & 75.00 $\pm$ 3.80 & 35.34 & 98.65 $\pm$ 0.47 & 35.65 & 87.80 $\pm$ 1.25 & 30.37 & 28.36 $\pm$ 3.68 \\
 & 2048 & 44.25 & 86.60 $\pm$ 2.90 & 48.11 & 99.30 $\pm$ 0.35 & 48.44 & 89.15 $\pm$ 1.25 & 41.78 & 26.08 $\pm$ 3.54 \\
 & 4096 & 61.04 & 89.00 $\pm$ 2.60 & 64.75 & 99.20 $\pm$ 0.38 & 65.06 & 89.80 $\pm$ 1.20 & 58.62 & 31.84 $\pm$ 3.56 \\
\midrule
\multirow{6}{*}{\shortstack{\textbf{SANTA\texttt{++}}\\(Contiguous)}} & 128 & 15.05 & 1.80 $\pm$ 1.10 & 15.07 & 89.20 $\pm$ 1.27 & 15.14 & 70.25 $\pm$ 2.03 & 14.38 & 31.76 $\pm$ 3.42 \\
 & 256 & 17.27 & 12.40 $\pm$ 2.90 & 17.37 & 93.15 $\pm$ 1.15 & 17.45 & 77.00 $\pm$ 1.58 & 16.24 & 42.12 $\pm$ 3.88 \\
 & 512 & 21.19 & 35.80 $\pm$ 4.20 & 21.46 & 95.65 $\pm$ 1.08 & 21.54 & 80.75 $\pm$ 1.33 & 19.75 & 43.68 $\pm$ 3.90 \\
 & 1024 & 27.73 & 60.20 $\pm$ 4.20 & 28.33 & 97.70 $\pm$ 0.68 & 28.42 & 85.30 $\pm$ 1.25 & 25.88 & 39.96 $\pm$ 3.72 \\
 & 2048 & 38.13 & 73.20 $\pm$ 3.70 & 39.19 & 98.50 $\pm$ 0.52 & 39.25 & 87.75 $\pm$ 1.25 & 35.72 & 35.60 $\pm$ 3.82 \\
 & 4096 & 53.71 & 85.20 $\pm$ 3.10 & 55.12 & 99.15 $\pm$ 0.38 & 55.22 & 88.85 $\pm$ 1.20 & 50.44 & 28.40 $\pm$ 3.68 \\
\bottomrule
\end{tabular}
\vspace{2pt}
\begin{tabular}{l c c c c c c c c c}
\toprule
 &  & \multicolumn{2}{c}{CWE} & \multicolumn{2}{c}{FWE} & \multicolumn{2}{c}{QA-1} & \multicolumn{2}{c}{QA-2} \\
\cmidrule(lr){3-4}\cmidrule(lr){5-6}\cmidrule(lr){7-8}\cmidrule(lr){9-10}
Method & $S$ & KV (\%) & Acc (\%) & KV (\%) & Acc (\%) & KV (\%) & Acc (\%) & KV (\%) & Acc (\%) \\
\midrule
SDPA & --- & 100.00 & 74.34 $\pm$ 1.92 & 100.00 & 91.87 $\pm$ 1.30 & 100.00 & 69.80 $\pm$ 3.90 & 100.00 & 46.00 $\pm$ 4.50 \\
\midrule
\multirow{6}{*}{SANTA} & 128 & 50.31 & 70.20 $\pm$ 1.91 & 50.41 & 89.73 $\pm$ 1.43 & 50.36 & 69.40 $\pm$ 4.00 & 50.34 & 48.20 $\pm$ 4.50 \\
 & 256 & 50.53 & 71.86 $\pm$ 1.94 & 50.73 & 91.67 $\pm$ 1.30 & 50.62 & 69.20 $\pm$ 4.00 & 50.59 & 47.00 $\pm$ 4.40 \\
 & 512 & 50.89 & 73.72 $\pm$ 1.89 & 51.27 & 91.47 $\pm$ 1.37 & 51.08 & 70.40 $\pm$ 4.00 & 51.01 & 46.80 $\pm$ 4.60 \\
 & 1024 & 51.50 & 73.98 $\pm$ 1.98 & 52.17 & 91.67 $\pm$ 1.33 & 51.84 & 70.00 $\pm$ 3.90 & 51.71 & 46.80 $\pm$ 4.30 \\
 & 2048 & 52.45 & 75.02 $\pm$ 1.79 & 53.62 & 91.40 $\pm$ 1.40 & 53.02 & 70.00 $\pm$ 3.80 & 52.82 & 46.80 $\pm$ 4.40 \\
 & 4096 & 53.88 & 74.18 $\pm$ 1.85 & 55.80 & 91.53 $\pm$ 1.33 & 54.81 & 69.60 $\pm$ 3.90 & 54.49 & 46.20 $\pm$ 4.40 \\
\midrule
\multirow{6}{*}{\shortstack{\textbf{SANTA\texttt{++}}\\(K-means)}} & 128 & 16.13 & 56.80 $\pm$ 1.80 & 16.02 & 77.67 $\pm$ 2.13 & 17.06 & 65.00 $\pm$ 4.00 & 16.90 & 46.20 $\pm$ 4.20 \\
 & 256 & 19.71 & 59.08 $\pm$ 1.98 & 19.48 & 78.80 $\pm$ 2.13 & 21.20 & 69.00 $\pm$ 4.10 & 20.95 & 47.00 $\pm$ 4.50 \\
 & 512 & 25.44 & 62.94 $\pm$ 1.90 & 25.22 & 82.33 $\pm$ 1.67 & 27.64 & 69.60 $\pm$ 4.10 & 27.24 & 46.80 $\pm$ 4.30 \\
 & 1024 & 34.14 & 67.26 $\pm$ 2.02 & 34.25 & 85.07 $\pm$ 1.50 & 37.09 & 68.60 $\pm$ 4.00 & 36.60 & 47.20 $\pm$ 4.40 \\
 & 2048 & 46.58 & 70.34 $\pm$ 1.97 & 47.47 & 89.33 $\pm$ 1.40 & 50.10 & 70.20 $\pm$ 4.00 & 49.62 & 47.80 $\pm$ 4.30 \\
 & 4096 & 63.25 & 72.76 $\pm$ 1.84 & 65.20 & 90.07 $\pm$ 1.33 & 66.79 & 70.20 $\pm$ 3.90 & 66.42 & 45.60 $\pm$ 4.50 \\
\midrule
\multirow{6}{*}{\shortstack{\textbf{SANTA\texttt{++}}\\(Contiguous)}} & 128 & 14.98 & 41.32 $\pm$ 1.38 & 14.92 & 83.87 $\pm$ 1.73 & 15.28 & 63.80 $\pm$ 4.20 & 15.25 & 44.20 $\pm$ 4.50 \\
 & 256 & 17.23 & 47.18 $\pm$ 1.43 & 17.28 & 84.93 $\pm$ 1.60 & 17.82 & 65.00 $\pm$ 4.10 & 17.83 & 45.20 $\pm$ 4.20 \\
 & 512 & 21.26 & 51.86 $\pm$ 1.76 & 21.52 & 85.47 $\pm$ 1.83 & 22.24 & 66.80 $\pm$ 3.90 & 22.29 & 47.00 $\pm$ 4.60 \\
 & 1024 & 28.01 & 58.00 $\pm$ 1.70 & 28.75 & 88.20 $\pm$ 1.63 & 29.48 & 69.00 $\pm$ 3.70 & 29.65 & 48.60 $\pm$ 4.40 \\
 & 2048 & 38.65 & 63.42 $\pm$ 1.98 & 40.37 & 91.53 $\pm$ 1.33 & 40.70 & 70.20 $\pm$ 3.90 & 41.05 & 48.60 $\pm$ 4.30 \\
 & 4096 & 54.22 & 69.96 $\pm$ 2.03 & 57.46 & 91.13 $\pm$ 1.43 & 56.90 & 70.20 $\pm$ 3.90 & 57.42 & 48.20 $\pm$ 4.70 \\
\bottomrule
\end{tabular}
\caption{\textbf{Per-task accuracy and KV memory access on RULER (32K context) on Qwen2.5-7B-Instruct.} \textit{KV} is the theoretical GQA-aware KV memory access (\%), and $S$ denotes the nominal budget. Accuracy values are reported as mean $\pm$ 95\% bootstrap confidence interval half-width.}
\label{tab:ruler_by_task_32k}
\end{table}

\FloatBarrier

\section{{Quest and MagicPIG comparisons}}
\label{app:selection-baselines}

\subsection{{Evaluation protocol}}
\label{app:selection-protocol}
{Quest~\citep{tang2024quest} and MagicPIG~\citep{chen2025magicpig} use
Qwen2.5-7B-Instruct in FP16 on all 13 RULER-32K tasks, with 500 prompts
per task.} {We compare dense SDPA and both SANTA++ grouping policies with
five Quest budgets and twelve MagicPIG configurations.} {The complete aggregate and
task results appear in Tables~\ref{tab:selection-aggregate}
and~\ref{tab:selection-tasks}.}

{The model/tokenizer revision is
\nolinkurl{a09a35458c702b33eeacc393d103063234e8bc28}; data come from
\nolinkurl{SaylorTwift/RULER-32768-Qwen2.5-3B-tokenizer} and are
re-tokenized with the 7B tokenizer.} {We use the NVIDIA/RULER grader at
revision \texttt{38da79d}.} {Formatted prompts contain 27,149--32,797
tokens and are not truncated.} {Each is a single user turn in the Qwen
chat template, with any answer prefix appended.}

{Generation uses batch size one, greedy argmax, and EOS stopping.}
{Output caps are 128 for NIAH, 30 for VT, 120 for CWE, 50 for FWE,
and 32 for QA.} {Example selection uses seed 42.} {Stochastic operations use deterministic prompt-specific seeds
derived from base seed 0.}

{All methods densely prefill the complete prompt.} {Sparse methods then
replay its final token to produce the first sparse prediction; SDPA
uses its prefill logits.} {We retain each baseline's exact-token and
dense-layer policies below.} {These are accuracy adaptations, not
benchmarks of the original optimized serving implementations.}

{Scores use RULER's case-insensitive substring metrics: any accepted
answer for QA, and the fraction of required answers for other tasks.}
{Aggregate scores weight the 13 tasks equally.} {The 95\% intervals use
10,000 prompt-bootstrap resamples, stratified by task for aggregates,
with base seed 20260903.} {Tables show the point estimate and half the
percentile interval span.} {Only one stochastic seed per prompt is used.}

\paragraph{{Dense-limit integration checks.}}
{Pilot integration checks recover dense task scores on all 65 prompts
for both Quest with every page selected and MagicPIG with every key selected.}
{The Quest control explicitly includes the partial final page and also
matches the dense generated tokens on 63 of the 65 prompts.}

\subsection{{Baseline policies}}
\label{app:selection-policies}
\paragraph{{Quest.}}
{We adapt Quest's accuracy reference at revision \texttt{01c1623}.} {The cache is partitioned into positional pages of 16 tokens.} {For page $j$, let $m_j$ and $M_j$ be its channel-wise minimum and maximum post-RoPE key vectors.} {The query-dependent page score is}
\begingroup
\begin{equation}
 s_j(q)=\sum_a\max(q_a m_{j,a},q_a M_{j,a}).
\end{equation}
\endgroup
{Ranking omits the common positive attention scale; attention within selected pages uses the model's usual scaling.} {The adapter evaluates page scores with FP32 matrix products.} {For current cache length $N_c$ and token budget $B$, the number of selected pages is}
\begingroup
\begin{equation}
 K_{\rm page}=\min\!\left\{\left\lceil\frac{N_c}{16}\right\rceil,
 \max\!\left(3,\left\lfloor\frac{\min(N_c,B)}{16}\right\rfloor\right)\right\}.
\end{equation}
\endgroup
{This equals $B/16$ at the reported long-context settings.} {Each query head selects its own top pages and computes renormalized softmax over their valid tokens, without importance correction.} {A partial last page is an ordinary candidate; no sink or newest page is forced.} {Layers 0 and 1 remain dense.} {Generated tokens append to the paged cache and update its extrema, with no separate exact suffix.} {We evaluate $B\in\{512,1024,2048,4096,8192\}$.} {Full-prompt dense prefill and final-token replay follow the common sparse-method evaluation protocol.}

\paragraph{{MagicPIG.}}
{We adapt MagicPIG's RULER accuracy reference at revision \texttt{ac9aa36}.} {The first four and last 64 prompt tokens form an exact static set, and all generated tokens are added to it.} {The 64-token prompt tail stays fixed as generation grows.} {Layers 0 and 16 remain dense.} {Only the middle prompt positions $[4,N-64)$ enter the sampling index.}

{For each sparse layer and KV head, we subtract the mean middle key before hashing.} {Attention applies this same mean shift to all cache keys, which adds a common query-dependent constant to their logits.} {Gaussian FP32 projections have shape $[4,128,K_{\rm h}L]$ and are shared within each group of 16 layers and among query heads using the same KV head.} {There are two projection groups for the 28-layer model.} {Query vectors are L2-normalized before hashing.} {Strictly positive projection values set the hash bits; projections and key hash codes are drawn once per prompt and reused.}

{A middle key is selected when its $K_{\rm h}$-bit code matches the query in at least two of $L$ tables.} {If $z$ is the cosine between the query and centered key, define}
\begingroup
\begin{equation}
 p=1-\frac{\arccos(\operatorname{clip}(z,-1,1))}{\pi},\qquad
 a=p^{K_{\rm h}},\qquad
 u=1-(1-a)^L-La(1-a)^{L-1}.
\end{equation}
\endgroup
{The selected key's attention logit receives the correction $-\log(u+10^{-4})$.} {Exact static keys are not corrected.} {A joint softmax over sampled and exact keys produces the attention output.} {Hashing, cosine evaluation, and inclusion correction use FP32; model/cache execution uses FP16.} {Each query head retains its own inclusion mask.} {The GQA union is used for read accounting, not to enlarge an individual head's selected set.}

{The adapter uses GPU tensor operations to simulate this sampling policy, rather than implementing MagicPIG's CPU/GPU offload pipeline.} {Its logical counter records the KV rows needed by the policy.} {The auxiliary LSH quantities below model the reference index layout separately from the simulator's tensor allocations.}

\paragraph{{Selecting configurations.}}
{A pilot with five prompts per task calibrated access levels.} {Quest
covers low, medium, and high access.} {MagicPIG targets the six k-means SANTA++ access levels, with two neighboring hash-bit settings
per target.} {Pilot prompts remain in the full evaluation.} {All twelve MagicPIG configurations are evaluated on the full
benchmark and reported in Table~\ref{tab:selection-aggregate}.}

\paragraph{{Quest trade-off comparisons.}}
{Table~\ref{tab:quest-tradeoff} pairs each evaluated Quest budget with
an evaluated contiguous SANTA++ setting that has higher aggregate accuracy
and lower logical access.}

\begin{table}[!htbp]
\centering

\small
\setlength{\tabcolsep}{6pt}
\caption{ \textbf{Higher RULER accuracy at lower access than Quest.}
Each row compares an evaluated Quest configuration with a contiguous
SANTA++ configuration using fewer KV reads. Scores are the 13-task
mean with 95\% bootstrap interval half-widths.}
\label{tab:quest-tradeoff}
\begin{tabular}{@{}rrrrrr@{}}
\toprule
\multicolumn{3}{c}{Quest} & \multicolumn{3}{c}{SANTA++ contiguous}\\
\cmidrule(lr){1-3}\cmidrule(l){4-6}
$B$ & Access (\%) & Score (\%) & $S$ & Access (\%) & Score (\%)\\
\midrule
512 & 16.80 & $59.84\pm0.80$ & 128 & 15.03 & $64.72\pm0.77$\\
1024 & 20.47 & $65.46\pm0.79$ & 256 & 17.31 & $70.54\pm0.76$\\
2048 & 26.88 & $68.70\pm0.78$ & 512 & 21.37 & $75.37\pm0.75$\\
4096 & 37.64 & $73.65\pm0.77$ & 1024 & 28.16 & $79.40\pm0.73$\\
8192 & 54.88 & $78.89\pm0.76$ & 4096 & 54.68 & $82.56\pm0.67$\\
\bottomrule
\end{tabular}
\end{table}

\subsection{{Logical access and auxiliary storage}}
\label{app:selection-accounting}
{We use the GQA-union convention of
Equation~\ref{eq:qwen-logical-access}.} {Let $c$ index a decode call,
layer, and KV head, and let $D_m=\sum_c2N_{m,c}$ be the dense-equivalent
count along method $m$'s generated trajectory.} {Each query head keeps
its own selection mask; only the read counter takes their union.}
{Access percentages pool counts across prompts before division.}

{For Quest, $P_c=\lceil N_c/16\rceil$ pages contribute two key-sized summary vectors each.} {If $v_c$ is the valid-row union in selected pages, the per-query count is}
\begingroup
\begin{equation}
 a_{Q,c}=\begin{cases}
 2N_c,&\ell\in\{0,1\},\\
 2P_c+2v_c,&\text{otherwise},
 \end{cases}
 \qquad A_Q=100\,\frac{\sum_c a_{Q,c}}{D_Q}.
\end{equation}
\endgroup
{A partial page contributes only its valid KV rows.} {The approximately 5.80 percentage points of summary-vector reads are already included in $A_Q$.} {Dense layers are charged full KV reads without a second summary scan.}

{For MagicPIG, let $e_c$ count exact sink, fixed prompt-tail, and generated rows, and $v_c$ count sampled-middle rows in the GQA union.} {Its KV-only access is}
\begingroup
\begin{equation}
 a_{M,c}=\begin{cases}
 2N_c,&\ell\in\{0,16\},\\
 2(e_c+v_c),&\text{otherwise},
 \end{cases}
 \qquad A_M=100\,\frac{\sum_c a_{M,c}}{D_M}.
\end{equation}
\endgroup
{Its separate auxiliary-payload estimate counts $J$ query-head lookups across sparse layers, $C_{\rm hit}$ candidate occurrences across colliding tables, and $T_{\rm sel}$ sampled-middle rows summed per query head:}
\begingroup
\begin{equation}
 B_{\rm LSH}=5C_{\rm hit}+12LJ+4T_{\rm sel}\quad\text{bytes},\qquad
 H_M=100\,\frac{B_{\rm LSH}}{256D_M}.
\end{equation}
\endgroup
{The terms charge a four-byte token index and one-byte collision counter per candidate occurrence, two four-byte bucket bounds and a four-byte query code per table lookup, and a four-byte norm per sampled key.} {An FP16 key or value vector of dimension 128 occupies 256 bytes.} {Thus $A_M+H_M$ is the augmented payload ratio.} {Projection reads are excluded under the index model's cache-residency assumption.} {These logical counts and payload estimates are not complete measured DRAM traffic.}

\paragraph{{Storage.}}
{Quest's summary allocation includes room for generation up to $T_{\max}$:}
\begingroup
\begin{equation}
 B_{Q,\mathrm{meta}}=26\cdot4\cdot\left\lceil\frac{N+T_{\max}}{16}\right\rceil
 \cdot2\cdot128\cdot2\quad\text{bytes}.
\end{equation}
\endgroup
{Its maximum (mean) across prompts is 0.102 (0.100) GiB at each budget.} {For MagicPIG, the modeled hash-table and bucket-bound storage is}
\begingroup
\begin{equation}
 B_{M,\mathrm{index}}=26\cdot4\cdot L
 \left[4\max(N-68,0)+8\cdot2^{K_{\rm h}}\right]\quad\text{bytes}.
\end{equation}
\endgroup
{The estimate describes the CPU hash index in MagicPIG's reference
design, in addition to the KV cache.} {It excludes key norms, collision
buffers, and other runtime state.} {Tables report maximum (mean)
storage across prompts in GiB.} {At $(10,235)$, the index is
3.17 (3.11) GiB, compared with 1.63 (1.60) GiB at $(9,125)$.}
{The accuracy comparison uses a GPU simulation of the sampling
policy; index storage and lookup payload are analytical estimates,
not measured simulator memory or runtime.}

\begingroup

\footnotesize
\setlength{\tabcolsep}{3pt}
\renewcommand{\arraystretch}{1.03}
\begingroup
\begin{longtable}{@{}llrrrrr@{}}
\caption{ Complete 32K RULER aggregate sweep. Scores are equal-task means with 95\% bootstrap interval half-spans. $A$ includes routing-key reads for SANTA++/Quest and KV-only reads for MagicPIG; $H$ and $A+H$ are MagicPIG's modeled LSH payload and augmented access. Auxiliary GiB is maximum (mean): allocated Quest page summaries or modeled MagicPIG CPU index storage. Dashes denote quantities not reported or not applicable.}\label{tab:selection-aggregate}\\
\toprule
Method & Setting & $A$ (\%) & Score (\%) & $H$ (\%) & $A+H$ (\%) & Aux. GiB \\
\midrule
\endfirsthead
\multicolumn{7}{c}{\tablename\ \thetable\ (continued)}\\
\toprule
Method & Setting & $A$ (\%) & Score (\%) & $H$ (\%) & $A+H$ (\%) & Aux. GiB \\
\midrule
\endhead
\midrule
\multicolumn{7}{r}{Continued on next page}\\
\endfoot
\bottomrule
\endlastfoot
SDPA & --- & 100.00 & $83.51\pm0.66$ & --- & --- & --- \\
\addlinespace[3pt]
SANTA++ contig. & $S=128$ & 15.03 & $64.72\pm0.77$ & --- & --- & --- \\
SANTA++ contig. & $S=256$ & 17.31 & $70.54\pm0.76$ & --- & --- & --- \\
SANTA++ contig. & $S=512$ & 21.37 & $75.37\pm0.75$ & --- & --- & --- \\
SANTA++ contig. & $S=1024$ & 28.16 & $79.40\pm0.73$ & --- & --- & --- \\
SANTA++ contig. & $S=2048$ & 38.89 & $81.42\pm0.72$ & --- & --- & --- \\
SANTA++ contig. & $S=4096$ & 54.68 & $82.56\pm0.67$ & --- & --- & --- \\
\addlinespace[3pt]
SANTA++ $k$-means & $S=128$ & 16.38 & $71.33\pm0.74$ & --- & --- & --- \\
SANTA++ $k$-means & $S=256$ & 19.95 & $76.05\pm0.74$ & --- & --- & --- \\
SANTA++ $k$-means & $S=512$ & 25.70 & $78.49\pm0.74$ & --- & --- & --- \\
SANTA++ $k$-means & $S=1024$ & 34.44 & $80.40\pm0.72$ & --- & --- & --- \\
SANTA++ $k$-means & $S=2048$ & 46.95 & $82.29\pm0.66$ & --- & --- & --- \\
SANTA++ $k$-means & $S=4096$ & 63.65 & $83.08\pm0.64$ & --- & --- & --- \\
\addlinespace[3pt]
Quest & $B=512$ & 16.80 & $59.84\pm0.80$ & --- & --- & 0.102 (0.100) \\
Quest & $B=1024$ & 20.47 & $65.46\pm0.79$ & --- & --- & 0.102 (0.100) \\
Quest & $B=2048$ & 26.88 & $68.70\pm0.78$ & --- & --- & 0.102 (0.100) \\
Quest & $B=4096$ & 37.64 & $73.65\pm0.77$ & --- & --- & 0.102 (0.100) \\
Quest & $B=8192$ & 54.88 & $78.89\pm0.76$ & --- & --- & 0.102 (0.100) \\
\addlinespace[3pt]
MagicPIG & $(8,120)$ & 46.85 & $81.83\pm0.69$ & 3.93 & 50.78 & 1.55 (1.52) \\
MagicPIG & $(8,175)$ & 63.86 & $83.44\pm0.64$ & 5.88 & 69.74 & 2.25 (2.21) \\
MagicPIG & $(9,80)$ & 16.78 & $44.78\pm0.92$ & 1.28 & 18.05 & 1.05 (1.03) \\
MagicPIG & $(9,95)$ & 19.69 & $55.72\pm0.96$ & 1.53 & 21.23 & 1.24 (1.22) \\
MagicPIG & $(9,125)$ & 25.87 & $71.39\pm0.89$ & 2.05 & 27.92 & 1.63 (1.60) \\
MagicPIG & $(9,165)$ & 34.24 & $80.39\pm0.73$ & 2.77 & 37.00 & 2.16 (2.12) \\
MagicPIG & $(9,230)$ & 46.91 & $83.11\pm0.65$ & 3.95 & 50.86 & 3.01 (2.95) \\
MagicPIG & $(9,340)$ & 63.98 & $83.43\pm0.65$ & 5.98 & 69.96 & 4.45 (4.36) \\
MagicPIG & $(10,145)$ & 16.33 & $54.12\pm0.95$ & 1.23 & 17.56 & 1.95 (1.92) \\
MagicPIG & $(10,180)$ & 19.88 & $69.31\pm0.91$ & 1.55 & 21.43 & 2.43 (2.38) \\
MagicPIG & $(10,235)$ & 25.77 & $78.91\pm0.76$ & 2.06 & 27.83 & 3.17 (3.11) \\
MagicPIG & $(10,315)$ & 34.38 & $82.27\pm0.69$ & 2.82 & 37.20 & 4.24 (4.17) \\
\end{longtable}
\endgroup
\endgroup

\subsection{{Complete task-level baseline results}}
{Table~\ref{tab:selection-tasks} reports all five Quest budgets and twelve
MagicPIG configurations for each task.} {Access and storage use the
definitions above.} {SANTA++ task-level results are in
Appendix~\ref{app:ruler}.}

\begingroup

\footnotesize
\setlength{\tabcolsep}{3pt}
\renewcommand{\arraystretch}{1.03}
\begingroup
\begin{longtable}{@{}lllrrrrr@{}}
\caption{ All Quest and MagicPIG task-level results at 32K (500 prompts per task and configuration). Scores are absolute percentages with 95\% bootstrap interval half-spans. $H$ and $A+H$ are modeled quantities for MagicPIG only. Auxiliary GiB is maximum (mean), with the distinct storage definitions in Appendix~\ref{app:selection-accounting}.}\label{tab:selection-tasks}\\
\toprule
Task & Method & Setting & $A$ (\%) & Score (\%) & $H$ (\%) & $A+H$ (\%) & Aux. GiB \\
\midrule
\endfirsthead
\multicolumn{8}{c}{\tablename\ \thetable\ (continued)}\\
\toprule
Task & Method & Setting & $A$ (\%) & Score (\%) & $H$ (\%) & $A+H$ (\%) & Aux. GiB \\
\midrule
\endhead
\midrule
\multicolumn{8}{r}{Continued on next page}\\
\endfoot
\bottomrule
\endlastfoot
S-NIAH-1 & Quest & $B=512$ & 16.68 & $99.60\pm0.50$ & --- & --- & 0.101 (0.101) \\
S-NIAH-1 & Quest & $B=1024$ & 20.71 & $99.80\pm0.30$ & --- & --- & 0.101 (0.101) \\
S-NIAH-1 & Quest & $B=2048$ & 27.85 & $99.60\pm0.50$ & --- & --- & 0.101 (0.101) \\
S-NIAH-1 & Quest & $B=4096$ & 39.79 & $100.00\pm0.00$ & --- & --- & 0.101 (0.101) \\
S-NIAH-1 & Quest & $B=8192$ & 58.38 & $100.00\pm0.00$ & --- & --- & 0.101 (0.101) \\
S-NIAH-1 & MagicPIG & $(8,120)$ & 47.86 & $100.00\pm0.00$ & 3.96 & 51.83 & 1.53 (1.53) \\
S-NIAH-1 & MagicPIG & $(8,175)$ & 65.00 & $100.00\pm0.00$ & 5.92 & 70.92 & 2.24 (2.23) \\
S-NIAH-1 & MagicPIG & $(9,80)$ & 17.16 & $81.60\pm3.40$ & 1.29 & 18.45 & 1.04 (1.04) \\
S-NIAH-1 & MagicPIG & $(9,95)$ & 20.16 & $89.80\pm2.70$ & 1.55 & 21.71 & 1.23 (1.23) \\
S-NIAH-1 & MagicPIG & $(9,125)$ & 26.46 & $97.20\pm1.50$ & 2.08 & 28.54 & 1.62 (1.62) \\
S-NIAH-1 & MagicPIG & $(9,165)$ & 34.96 & $99.40\pm0.70$ & 2.80 & 37.75 & 2.14 (2.14) \\
S-NIAH-1 & MagicPIG & $(9,230)$ & 47.77 & $100.00\pm0.00$ & 3.99 & 51.76 & 2.98 (2.98) \\
S-NIAH-1 & MagicPIG & $(9,340)$ & 64.96 & $100.00\pm0.00$ & 6.03 & 70.99 & 4.41 (4.41) \\
S-NIAH-1 & MagicPIG & $(10,145)$ & 16.70 & $90.20\pm2.70$ & 1.25 & 17.95 & 1.94 (1.94) \\
S-NIAH-1 & MagicPIG & $(10,180)$ & 20.35 & $96.80\pm1.50$ & 1.57 & 21.93 & 2.41 (2.41) \\
S-NIAH-1 & MagicPIG & $(10,235)$ & 26.33 & $99.80\pm0.30$ & 2.09 & 28.41 & 3.14 (3.14) \\
S-NIAH-1 & MagicPIG & $(10,315)$ & 35.01 & $100.00\pm0.00$ & 2.85 & 37.87 & 4.21 (4.21) \\
\addlinespace[4pt]
S-NIAH-2 & Quest & $B=512$ & 16.82 & $97.00\pm1.50$ & --- & --- & 0.101 (0.101) \\
S-NIAH-2 & Quest & $B=1024$ & 20.48 & $97.40\pm1.30$ & --- & --- & 0.101 (0.101) \\
S-NIAH-2 & Quest & $B=2048$ & 26.97 & $96.80\pm1.50$ & --- & --- & 0.101 (0.101) \\
S-NIAH-2 & Quest & $B=4096$ & 37.73 & $99.00\pm0.90$ & --- & --- & 0.101 (0.101) \\
S-NIAH-2 & Quest & $B=8192$ & 55.19 & $100.00\pm0.00$ & --- & --- & 0.101 (0.101) \\
S-NIAH-2 & MagicPIG & $(8,120)$ & 46.70 & $97.80\pm1.30$ & 3.92 & 50.62 & 1.53 (1.53) \\
S-NIAH-2 & MagicPIG & $(8,175)$ & 63.75 & $100.00\pm0.00$ & 5.87 & 69.61 & 2.23 (2.23) \\
S-NIAH-2 & MagicPIG & $(9,80)$ & 16.65 & $49.00\pm4.40$ & 1.27 & 17.92 & 1.03 (1.03) \\
S-NIAH-2 & MagicPIG & $(9,95)$ & 19.54 & $63.00\pm4.30$ & 1.53 & 21.07 & 1.23 (1.23) \\
S-NIAH-2 & MagicPIG & $(9,125)$ & 25.71 & $85.60\pm3.00$ & 2.05 & 27.76 & 1.62 (1.61) \\
S-NIAH-2 & MagicPIG & $(9,165)$ & 34.07 & $96.60\pm1.50$ & 2.76 & 36.83 & 2.13 (2.13) \\
S-NIAH-2 & MagicPIG & $(9,230)$ & 46.79 & $99.80\pm0.30$ & 3.94 & 50.72 & 2.97 (2.97) \\
S-NIAH-2 & MagicPIG & $(9,340)$ & 63.89 & $100.00\pm0.00$ & 5.96 & 69.86 & 4.39 (4.39) \\
S-NIAH-2 & MagicPIG & $(10,145)$ & 16.19 & $61.00\pm4.30$ & 1.23 & 17.42 & 1.93 (1.93) \\
S-NIAH-2 & MagicPIG & $(10,180)$ & 19.72 & $84.40\pm3.10$ & 1.54 & 21.27 & 2.40 (2.40) \\
S-NIAH-2 & MagicPIG & $(10,235)$ & 25.59 & $96.20\pm1.70$ & 2.05 & 27.65 & 3.13 (3.13) \\
S-NIAH-2 & MagicPIG & $(10,315)$ & 34.21 & $98.60\pm1.10$ & 2.80 & 37.02 & 4.20 (4.19) \\
\addlinespace[4pt]
S-NIAH-3 & Quest & $B=512$ & 16.83 & $52.00\pm4.40$ & --- & --- & 0.101 (0.101) \\
S-NIAH-3 & Quest & $B=1024$ & 20.44 & $68.40\pm4.10$ & --- & --- & 0.101 (0.101) \\
S-NIAH-3 & Quest & $B=2048$ & 26.80 & $80.20\pm3.50$ & --- & --- & 0.101 (0.101) \\
S-NIAH-3 & Quest & $B=4096$ & 37.64 & $88.80\pm2.70$ & --- & --- & 0.101 (0.101) \\
S-NIAH-3 & Quest & $B=8192$ & 55.12 & $95.40\pm1.90$ & --- & --- & 0.101 (0.101) \\
S-NIAH-3 & MagicPIG & $(8,120)$ & 46.50 & $94.80\pm1.90$ & 3.91 & 50.41 & 1.53 (1.53) \\
S-NIAH-3 & MagicPIG & $(8,175)$ & 63.58 & $98.60\pm1.10$ & 5.85 & 69.44 & 2.23 (2.23) \\
S-NIAH-3 & MagicPIG & $(9,80)$ & 16.56 & $12.20\pm2.90$ & 1.27 & 17.83 & 1.03 (1.03) \\
S-NIAH-3 & MagicPIG & $(9,95)$ & 19.44 & $30.60\pm4.10$ & 1.52 & 20.96 & 1.23 (1.23) \\
S-NIAH-3 & MagicPIG & $(9,125)$ & 25.55 & $70.00\pm4.10$ & 2.04 & 27.59 & 1.62 (1.62) \\
S-NIAH-3 & MagicPIG & $(9,165)$ & 33.90 & $91.60\pm2.40$ & 2.75 & 36.65 & 2.13 (2.13) \\
S-NIAH-3 & MagicPIG & $(9,230)$ & 46.58 & $99.00\pm0.90$ & 3.92 & 50.50 & 2.97 (2.97) \\
S-NIAH-3 & MagicPIG & $(9,340)$ & 63.77 & $99.00\pm0.90$ & 5.94 & 69.71 & 4.40 (4.39) \\
S-NIAH-3 & MagicPIG & $(10,145)$ & 16.10 & $32.80\pm4.10$ & 1.22 & 17.32 & 1.93 (1.93) \\
S-NIAH-3 & MagicPIG & $(10,180)$ & 19.61 & $66.80\pm4.00$ & 1.54 & 21.14 & 2.40 (2.40) \\
S-NIAH-3 & MagicPIG & $(10,235)$ & 25.45 & $88.60\pm2.80$ & 2.04 & 27.49 & 3.13 (3.13) \\
S-NIAH-3 & MagicPIG & $(10,315)$ & 34.03 & $96.00\pm1.70$ & 2.79 & 36.83 & 4.20 (4.20) \\
\addlinespace[4pt]
MK-NIAH-1 & Quest & $B=512$ & 16.78 & $93.80\pm2.10$ & --- & --- & 0.101 (0.101) \\
MK-NIAH-1 & Quest & $B=1024$ & 20.39 & $94.60\pm2.00$ & --- & --- & 0.101 (0.101) \\
MK-NIAH-1 & Quest & $B=2048$ & 26.78 & $96.20\pm1.70$ & --- & --- & 0.101 (0.101) \\
MK-NIAH-1 & Quest & $B=4096$ & 37.60 & $98.20\pm1.10$ & --- & --- & 0.101 (0.101) \\
MK-NIAH-1 & Quest & $B=8192$ & 54.98 & $98.60\pm1.10$ & --- & --- & 0.101 (0.101) \\
MK-NIAH-1 & MagicPIG & $(8,120)$ & 46.66 & $96.40\pm1.70$ & 3.92 & 50.58 & 1.53 (1.53) \\
MK-NIAH-1 & MagicPIG & $(8,175)$ & 63.71 & $98.40\pm1.10$ & 5.86 & 69.57 & 2.23 (2.23) \\
MK-NIAH-1 & MagicPIG & $(9,80)$ & 16.61 & $47.00\pm4.40$ & 1.27 & 17.89 & 1.04 (1.04) \\
MK-NIAH-1 & MagicPIG & $(9,95)$ & 19.51 & $64.00\pm4.20$ & 1.53 & 21.03 & 1.23 (1.23) \\
MK-NIAH-1 & MagicPIG & $(9,125)$ & 25.66 & $83.60\pm3.30$ & 2.05 & 27.71 & 1.62 (1.62) \\
MK-NIAH-1 & MagicPIG & $(9,165)$ & 34.03 & $93.60\pm2.10$ & 2.75 & 36.78 & 2.14 (2.14) \\
MK-NIAH-1 & MagicPIG & $(9,230)$ & 46.73 & $97.80\pm1.30$ & 3.93 & 50.67 & 2.98 (2.98) \\
MK-NIAH-1 & MagicPIG & $(9,340)$ & 63.86 & $99.00\pm0.90$ & 5.96 & 69.82 & 4.40 (4.40) \\
MK-NIAH-1 & MagicPIG & $(10,145)$ & 16.15 & $60.00\pm4.20$ & 1.23 & 17.37 & 1.93 (1.93) \\
MK-NIAH-1 & MagicPIG & $(10,180)$ & 19.68 & $81.40\pm3.40$ & 1.54 & 21.23 & 2.40 (2.40) \\
MK-NIAH-1 & MagicPIG & $(10,235)$ & 25.55 & $92.20\pm2.30$ & 2.05 & 27.60 & 3.14 (3.13) \\
MK-NIAH-1 & MagicPIG & $(10,315)$ & 34.16 & $96.80\pm1.50$ & 2.80 & 36.96 & 4.20 (4.20) \\
\addlinespace[4pt]
MK-NIAH-2 & Quest & $B=512$ & 17.30 & $44.00\pm4.40$ & --- & --- & 0.101 (0.100) \\
MK-NIAH-2 & Quest & $B=1024$ & 21.38 & $56.20\pm4.30$ & --- & --- & 0.101 (0.100) \\
MK-NIAH-2 & Quest & $B=2048$ & 28.34 & $64.40\pm4.20$ & --- & --- & 0.101 (0.100) \\
MK-NIAH-2 & Quest & $B=4096$ & 39.84 & $72.60\pm3.90$ & --- & --- & 0.101 (0.100) \\
MK-NIAH-2 & Quest & $B=8192$ & 57.20 & $74.80\pm3.90$ & --- & --- & 0.101 (0.100) \\
MK-NIAH-2 & MagicPIG & $(8,120)$ & 47.26 & $92.00\pm2.40$ & 3.95 & 51.22 & 1.53 (1.52) \\
MK-NIAH-2 & MagicPIG & $(8,175)$ & 64.29 & $93.20\pm2.20$ & 5.91 & 70.20 & 2.23 (2.22) \\
MK-NIAH-2 & MagicPIG & $(9,80)$ & 16.97 & $39.80\pm4.30$ & 1.29 & 18.26 & 1.03 (1.03) \\
MK-NIAH-2 & MagicPIG & $(9,95)$ & 19.92 & $59.20\pm4.40$ & 1.55 & 21.47 & 1.23 (1.22) \\
MK-NIAH-2 & MagicPIG & $(9,125)$ & 26.15 & $80.20\pm3.40$ & 2.07 & 28.22 & 1.61 (1.61) \\
MK-NIAH-2 & MagicPIG & $(9,165)$ & 34.60 & $92.20\pm2.40$ & 2.79 & 37.39 & 2.13 (2.12) \\
MK-NIAH-2 & MagicPIG & $(9,230)$ & 47.37 & $92.00\pm2.30$ & 3.98 & 51.34 & 2.97 (2.96) \\
MK-NIAH-2 & MagicPIG & $(9,340)$ & 64.44 & $93.60\pm2.10$ & 6.02 & 70.46 & 4.39 (4.38) \\
MK-NIAH-2 & MagicPIG & $(10,145)$ & 16.52 & $59.00\pm4.40$ & 1.25 & 17.77 & 1.93 (1.92) \\
MK-NIAH-2 & MagicPIG & $(10,180)$ & 20.12 & $75.40\pm3.80$ & 1.57 & 21.68 & 2.40 (2.39) \\
MK-NIAH-2 & MagicPIG & $(10,235)$ & 26.07 & $89.40\pm2.70$ & 2.08 & 28.15 & 3.13 (3.12) \\
MK-NIAH-2 & MagicPIG & $(10,315)$ & 34.79 & $92.00\pm2.40$ & 2.84 & 37.63 & 4.19 (4.18) \\
\addlinespace[4pt]
MK-NIAH-3 & Quest & $B=512$ & 16.56 & $0.00\pm0.00$ & --- & --- & 0.102 (0.101) \\
MK-NIAH-3 & Quest & $B=1024$ & 19.75 & $0.80\pm0.70$ & --- & --- & 0.102 (0.101) \\
MK-NIAH-3 & Quest & $B=2048$ & 25.12 & $5.40\pm1.90$ & --- & --- & 0.102 (0.101) \\
MK-NIAH-3 & Quest & $B=4096$ & 33.93 & $19.60\pm3.40$ & --- & --- & 0.102 (0.101) \\
MK-NIAH-3 & Quest & $B=8192$ & 49.00 & $55.80\pm4.40$ & --- & --- & 0.102 (0.101) \\
MK-NIAH-3 & MagicPIG & $(8,120)$ & 46.99 & $86.40\pm3.10$ & 3.94 & 50.92 & 1.54 (1.53) \\
MK-NIAH-3 & MagicPIG & $(8,175)$ & 63.97 & $90.00\pm2.70$ & 5.89 & 69.86 & 2.24 (2.23) \\
MK-NIAH-3 & MagicPIG & $(9,80)$ & 16.91 & $5.40\pm1.90$ & 1.28 & 18.18 & 1.04 (1.04) \\
MK-NIAH-3 & MagicPIG & $(9,95)$ & 19.83 & $20.00\pm3.50$ & 1.53 & 21.36 & 1.24 (1.23) \\
MK-NIAH-3 & MagicPIG & $(9,125)$ & 26.02 & $53.80\pm4.40$ & 2.06 & 28.08 & 1.63 (1.62) \\
MK-NIAH-3 & MagicPIG & $(9,165)$ & 34.40 & $82.80\pm3.30$ & 2.77 & 37.17 & 2.15 (2.14) \\
MK-NIAH-3 & MagicPIG & $(9,230)$ & 47.06 & $88.40\pm2.80$ & 3.96 & 51.02 & 3.00 (2.98) \\
MK-NIAH-3 & MagicPIG & $(9,340)$ & 64.08 & $89.40\pm2.70$ & 6.00 & 70.08 & 4.43 (4.41) \\
MK-NIAH-3 & MagicPIG & $(10,145)$ & 16.47 & $15.00\pm3.00$ & 1.24 & 17.71 & 1.95 (1.94) \\
MK-NIAH-3 & MagicPIG & $(10,180)$ & 20.04 & $48.00\pm4.30$ & 1.56 & 21.60 & 2.42 (2.40) \\
MK-NIAH-3 & MagicPIG & $(10,235)$ & 25.93 & $78.80\pm3.50$ & 2.07 & 28.00 & 3.15 (3.14) \\
MK-NIAH-3 & MagicPIG & $(10,315)$ & 34.55 & $86.60\pm3.00$ & 2.83 & 37.38 & 4.23 (4.21) \\
\addlinespace[4pt]
MQ-NIAH & Quest & $B=512$ & 16.63 & $88.50\pm1.55$ & --- & --- & 0.101 (0.101) \\
MQ-NIAH & Quest & $B=1024$ & 20.17 & $91.55\pm1.38$ & --- & --- & 0.101 (0.101) \\
MQ-NIAH & Quest & $B=2048$ & 26.50 & $93.10\pm1.30$ & --- & --- & 0.101 (0.101) \\
MQ-NIAH & Quest & $B=4096$ & 37.33 & $94.00\pm1.48$ & --- & --- & 0.101 (0.101) \\
MQ-NIAH & Quest & $B=8192$ & 54.74 & $99.15\pm0.40$ & --- & --- & 0.101 (0.101) \\
MQ-NIAH & MagicPIG & $(8,120)$ & 46.49 & $96.65\pm0.75$ & 3.91 & 50.40 & 1.53 (1.53) \\
MQ-NIAH & MagicPIG & $(8,175)$ & 63.51 & $99.15\pm0.40$ & 5.85 & 69.36 & 2.23 (2.23) \\
MQ-NIAH & MagicPIG & $(9,80)$ & 16.63 & $33.05\pm2.35$ & 1.27 & 17.90 & 1.04 (1.04) \\
MQ-NIAH & MagicPIG & $(9,95)$ & 19.50 & $53.70\pm2.35$ & 1.52 & 21.02 & 1.23 (1.23) \\
MQ-NIAH & MagicPIG & $(9,125)$ & 25.61 & $81.05\pm1.75$ & 2.04 & 27.65 & 1.62 (1.62) \\
MQ-NIAH & MagicPIG & $(9,165)$ & 33.92 & $94.40\pm0.95$ & 2.75 & 36.67 & 2.14 (2.14) \\
MQ-NIAH & MagicPIG & $(9,230)$ & 46.58 & $98.60\pm0.52$ & 3.92 & 50.50 & 2.98 (2.98) \\
MQ-NIAH & MagicPIG & $(9,340)$ & 63.70 & $99.30\pm0.35$ & 5.94 & 69.64 & 4.41 (4.40) \\
MQ-NIAH & MagicPIG & $(10,145)$ & 16.16 & $52.05\pm2.48$ & 1.22 & 17.39 & 1.94 (1.94) \\
MQ-NIAH & MagicPIG & $(10,180)$ & 19.67 & $78.30\pm1.88$ & 1.54 & 21.21 & 2.40 (2.40) \\
MQ-NIAH & MagicPIG & $(10,235)$ & 25.49 & $90.50\pm1.35$ & 2.04 & 27.53 & 3.14 (3.14) \\
MQ-NIAH & MagicPIG & $(10,315)$ & 34.06 & $98.05\pm0.62$ & 2.79 & 36.85 & 4.21 (4.21) \\
\addlinespace[4pt]
MV-NIAH & Quest & $B=512$ & 16.74 & $70.80\pm1.67$ & --- & --- & 0.101 (0.101) \\
MV-NIAH & Quest & $B=1024$ & 20.31 & $75.20\pm1.55$ & --- & --- & 0.101 (0.101) \\
MV-NIAH & Quest & $B=2048$ & 26.67 & $76.20\pm1.80$ & --- & --- & 0.101 (0.101) \\
MV-NIAH & Quest & $B=4096$ & 37.46 & $83.00\pm1.38$ & --- & --- & 0.101 (0.101) \\
MV-NIAH & Quest & $B=8192$ & 54.85 & $88.70\pm1.25$ & --- & --- & 0.101 (0.101) \\
MV-NIAH & MagicPIG & $(8,120)$ & 46.57 & $86.80\pm1.33$ & 3.91 & 50.48 & 1.53 (1.53) \\
MV-NIAH & MagicPIG & $(8,175)$ & 63.61 & $89.25\pm1.25$ & 5.85 & 69.47 & 2.23 (2.23) \\
MV-NIAH & MagicPIG & $(9,80)$ & 16.60 & $23.85\pm2.18$ & 1.26 & 17.87 & 1.04 (1.04) \\
MV-NIAH & MagicPIG & $(9,95)$ & 19.49 & $44.80\pm2.53$ & 1.52 & 21.01 & 1.23 (1.23) \\
MV-NIAH & MagicPIG & $(9,125)$ & 25.63 & $70.00\pm2.15$ & 2.04 & 27.66 & 1.62 (1.62) \\
MV-NIAH & MagicPIG & $(9,165)$ & 33.97 & $83.15\pm1.55$ & 2.75 & 36.72 & 2.14 (2.14) \\
MV-NIAH & MagicPIG & $(9,230)$ & 46.65 & $88.10\pm1.28$ & 3.92 & 50.58 & 2.98 (2.98) \\
MV-NIAH & MagicPIG & $(9,340)$ & 63.78 & $90.05\pm1.18$ & 5.95 & 69.73 & 4.40 (4.40) \\
MV-NIAH & MagicPIG & $(10,145)$ & 16.14 & $40.05\pm2.50$ & 1.22 & 17.36 & 1.94 (1.93) \\
MV-NIAH & MagicPIG & $(10,180)$ & 19.66 & $66.40\pm2.27$ & 1.54 & 21.20 & 2.40 (2.40) \\
MV-NIAH & MagicPIG & $(10,235)$ & 25.52 & $80.35\pm1.70$ & 2.04 & 27.57 & 3.14 (3.13) \\
MV-NIAH & MagicPIG & $(10,315)$ & 34.11 & $88.00\pm1.32$ & 2.79 & 36.91 & 4.20 (4.20) \\
\addlinespace[4pt]
VT & Quest & $B=512$ & 16.30 & $42.76\pm3.66$ & --- & --- & 0.101 (0.101) \\
VT & Quest & $B=1024$ & 20.08 & $55.08\pm3.76$ & --- & --- & 0.101 (0.101) \\
VT & Quest & $B=2048$ & 27.16 & $50.96\pm3.94$ & --- & --- & 0.101 (0.101) \\
VT & Quest & $B=4096$ & 39.17 & $45.92\pm3.94$ & --- & --- & 0.101 (0.101) \\
VT & Quest & $B=8192$ & 57.80 & $40.36\pm3.94$ & --- & --- & 0.101 (0.101) \\
VT & MagicPIG & $(8,120)$ & 47.56 & $32.68\pm3.80$ & 4.01 & 51.57 & 1.54 (1.54) \\
VT & MagicPIG & $(8,175)$ & 64.35 & $33.40\pm3.82$ & 5.99 & 70.34 & 2.25 (2.25) \\
VT & MagicPIG & $(9,80)$ & 17.36 & $27.04\pm3.24$ & 1.31 & 18.68 & 1.04 (1.04) \\
VT & MagicPIG & $(9,95)$ & 20.39 & $27.12\pm3.38$ & 1.58 & 21.97 & 1.24 (1.24) \\
VT & MagicPIG & $(9,125)$ & 26.66 & $30.44\pm3.62$ & 2.11 & 28.77 & 1.63 (1.63) \\
VT & MagicPIG & $(9,165)$ & 35.00 & $30.76\pm3.76$ & 2.84 & 37.84 & 2.15 (2.15) \\
VT & MagicPIG & $(9,230)$ & 47.49 & $32.40\pm3.74$ & 4.05 & 51.54 & 3.00 (3.00) \\
VT & MagicPIG & $(9,340)$ & 64.25 & $30.84\pm3.76$ & 6.13 & 70.38 & 4.43 (4.43) \\
VT & MagicPIG & $(10,145)$ & 17.01 & $25.52\pm3.38$ & 1.28 & 18.29 & 1.95 (1.95) \\
VT & MagicPIG & $(10,180)$ & 20.66 & $28.92\pm3.60$ & 1.61 & 22.27 & 2.42 (2.42) \\
VT & MagicPIG & $(10,235)$ & 26.59 & $31.16\pm3.74$ & 2.14 & 28.73 & 3.16 (3.16) \\
VT & MagicPIG & $(10,315)$ & 35.10 & $32.12\pm3.84$ & 2.92 & 38.02 & 4.23 (4.23) \\
\addlinespace[4pt]
CWE & Quest & $B=512$ & 17.14 & $36.18\pm1.46$ & --- & --- & 0.102 (0.101) \\
CWE & Quest & $B=1024$ & 21.28 & $35.94\pm1.50$ & --- & --- & 0.102 (0.101) \\
CWE & Quest & $B=2048$ & 28.30 & $41.94\pm1.61$ & --- & --- & 0.102 (0.101) \\
CWE & Quest & $B=4096$ & 39.58 & $57.36\pm1.73$ & --- & --- & 0.102 (0.101) \\
CWE & Quest & $B=8192$ & 57.26 & $69.08\pm1.84$ & --- & --- & 0.102 (0.101) \\
CWE & MagicPIG & $(8,120)$ & 47.18 & $73.22\pm1.95$ & 3.94 & 51.12 & 1.54 (1.53) \\
CWE & MagicPIG & $(8,175)$ & 64.10 & $74.16\pm1.95$ & 5.90 & 70.00 & 2.25 (2.24) \\
CWE & MagicPIG & $(9,80)$ & 16.97 & $67.16\pm2.03$ & 1.28 & 18.25 & 1.04 (1.04) \\
CWE & MagicPIG & $(9,95)$ & 19.93 & $70.14\pm1.88$ & 1.54 & 21.47 & 1.24 (1.23) \\
CWE & MagicPIG & $(9,125)$ & 26.16 & $71.60\pm1.88$ & 2.06 & 28.22 & 1.63 (1.62) \\
CWE & MagicPIG & $(9,165)$ & 34.56 & $73.36\pm1.89$ & 2.78 & 37.34 & 2.15 (2.14) \\
CWE & MagicPIG & $(9,230)$ & 47.22 & $74.50\pm1.91$ & 3.96 & 51.18 & 3.00 (2.98) \\
CWE & MagicPIG & $(9,340)$ & 64.17 & $74.70\pm1.92$ & 6.00 & 70.17 & 4.43 (4.41) \\
CWE & MagicPIG & $(10,145)$ & 16.53 & $69.28\pm1.96$ & 1.24 & 17.77 & 1.95 (1.94) \\
CWE & MagicPIG & $(10,180)$ & 20.14 & $72.08\pm1.97$ & 1.56 & 21.70 & 2.42 (2.41) \\
CWE & MagicPIG & $(10,235)$ & 26.08 & $71.60\pm1.94$ & 2.07 & 28.15 & 3.16 (3.14) \\
CWE & MagicPIG & $(10,315)$ & 34.72 & $74.24\pm1.94$ & 2.83 & 37.55 & 4.23 (4.21) \\
\addlinespace[4pt]
FWE & Quest & $B=512$ & 17.06 & $64.53\pm2.07$ & --- & --- & 0.102 (0.098) \\
FWE & Quest & $B=1024$ & 20.82 & $71.60\pm2.23$ & --- & --- & 0.102 (0.098) \\
FWE & Quest & $B=2048$ & 27.27 & $75.93\pm2.07$ & --- & --- & 0.102 (0.098) \\
FWE & Quest & $B=4096$ & 38.02 & $83.33\pm1.80$ & --- & --- & 0.102 (0.098) \\
FWE & Quest & $B=8192$ & 55.43 & $87.33\pm1.53$ & --- & --- & 0.102 (0.098) \\
FWE & MagicPIG & $(8,120)$ & 47.68 & $91.60\pm1.30$ & 3.92 & 51.60 & 1.55 (1.48) \\
FWE & MagicPIG & $(8,175)$ & 64.95 & $91.53\pm1.27$ & 5.87 & 70.82 & 2.25 (2.16) \\
FWE & MagicPIG & $(9,80)$ & 16.86 & $90.80\pm1.40$ & 1.28 & 18.14 & 1.05 (1.00) \\
FWE & MagicPIG & $(9,95)$ & 19.85 & $90.40\pm1.43$ & 1.53 & 21.38 & 1.24 (1.19) \\
FWE & MagicPIG & $(9,125)$ & 26.17 & $90.60\pm1.33$ & 2.05 & 28.22 & 1.63 (1.57) \\
FWE & MagicPIG & $(9,165)$ & 34.73 & $90.80\pm1.33$ & 2.76 & 37.49 & 2.16 (2.07) \\
FWE & MagicPIG & $(9,230)$ & 47.68 & $91.40\pm1.30$ & 3.94 & 51.62 & 3.01 (2.88) \\
FWE & MagicPIG & $(9,340)$ & 65.01 & $91.47\pm1.37$ & 5.96 & 70.97 & 4.45 (4.26) \\
FWE & MagicPIG & $(10,145)$ & 16.39 & $90.60\pm1.40$ & 1.23 & 17.62 & 1.95 (1.88) \\
FWE & MagicPIG & $(10,180)$ & 20.02 & $90.67\pm1.33$ & 1.55 & 21.57 & 2.43 (2.33) \\
FWE & MagicPIG & $(10,235)$ & 26.03 & $91.47\pm1.33$ & 2.06 & 28.09 & 3.17 (3.04) \\
FWE & MagicPIG & $(10,315)$ & 34.83 & $91.07\pm1.37$ & 2.81 & 37.64 & 4.24 (4.07) \\
\addlinespace[4pt]
QA-1 & Quest & $B=512$ & 16.85 & $51.80\pm4.30$ & --- & --- & 0.102 (0.098) \\
QA-1 & Quest & $B=1024$ & 20.27 & $61.40\pm4.20$ & --- & --- & 0.102 (0.098) \\
QA-1 & Quest & $B=2048$ & 26.30 & $66.00\pm4.20$ & --- & --- & 0.102 (0.098) \\
QA-1 & Quest & $B=4096$ & 37.00 & $68.60\pm4.00$ & --- & --- & 0.102 (0.098) \\
QA-1 & Quest & $B=8192$ & 54.90 & $69.40\pm4.00$ & --- & --- & 0.102 (0.098) \\
QA-1 & MagicPIG & $(8,120)$ & 46.38 & $69.00\pm4.00$ & 3.92 & 50.30 & 1.54 (1.49) \\
QA-1 & MagicPIG & $(8,175)$ & 63.38 & $70.40\pm4.00$ & 5.87 & 69.25 & 2.25 (2.17) \\
QA-1 & MagicPIG & $(9,80)$ & 16.50 & $60.60\pm4.20$ & 1.27 & 17.78 & 1.05 (1.01) \\
QA-1 & MagicPIG & $(9,95)$ & 19.37 & $65.40\pm4.20$ & 1.53 & 20.90 & 1.24 (1.20) \\
QA-1 & MagicPIG & $(9,125)$ & 25.50 & $67.60\pm4.10$ & 2.05 & 27.54 & 1.63 (1.57) \\
QA-1 & MagicPIG & $(9,165)$ & 33.80 & $69.20\pm4.10$ & 2.76 & 36.56 & 2.16 (2.08) \\
QA-1 & MagicPIG & $(9,230)$ & 46.49 & $70.80\pm4.00$ & 3.93 & 50.42 & 3.00 (2.89) \\
QA-1 & MagicPIG & $(9,340)$ & 63.59 & $70.00\pm4.00$ & 5.96 & 69.55 & 4.44 (4.28) \\
QA-1 & MagicPIG & $(10,145)$ & 16.04 & $63.80\pm4.20$ & 1.23 & 17.27 & 1.95 (1.88) \\
QA-1 & MagicPIG & $(10,180)$ & 19.55 & $66.40\pm4.20$ & 1.54 & 21.10 & 2.42 (2.34) \\
QA-1 & MagicPIG & $(10,235)$ & 25.39 & $69.20\pm4.00$ & 2.05 & 27.44 & 3.16 (3.05) \\
QA-1 & MagicPIG & $(10,315)$ & 33.96 & $69.20\pm4.00$ & 2.80 & 36.76 & 4.24 (4.09) \\
\addlinespace[4pt]
QA-2 & Quest & $B=512$ & 16.85 & $37.00\pm4.30$ & --- & --- & 0.102 (0.096) \\
QA-2 & Quest & $B=1024$ & 20.34 & $43.00\pm4.40$ & --- & --- & 0.102 (0.096) \\
QA-2 & Quest & $B=2048$ & 26.44 & $46.40\pm4.40$ & --- & --- & 0.102 (0.096) \\
QA-2 & Quest & $B=4096$ & 37.14 & $47.00\pm4.50$ & --- & --- & 0.102 (0.096) \\
QA-2 & Quest & $B=8192$ & 55.24 & $47.00\pm4.40$ & --- & --- & 0.102 (0.096) \\
QA-2 & MagicPIG & $(8,120)$ & 46.24 & $46.40\pm4.40$ & 3.91 & 50.16 & 1.54 (1.45) \\
QA-2 & MagicPIG & $(8,175)$ & 63.27 & $46.60\pm4.40$ & 5.86 & 69.13 & 2.25 (2.12) \\
QA-2 & MagicPIG & $(9,80)$ & 16.44 & $44.60\pm4.40$ & 1.27 & 17.71 & 1.04 (0.98) \\
QA-2 & MagicPIG & $(9,95)$ & 19.32 & $46.20\pm4.40$ & 1.53 & 20.85 & 1.24 (1.17) \\
QA-2 & MagicPIG & $(9,125)$ & 25.42 & $46.40\pm4.40$ & 2.04 & 27.46 & 1.63 (1.54) \\
QA-2 & MagicPIG & $(9,165)$ & 33.71 & $47.20\pm4.40$ & 2.75 & 36.46 & 2.16 (2.03) \\
QA-2 & MagicPIG & $(9,230)$ & 46.36 & $47.60\pm4.50$ & 3.93 & 50.29 & 3.00 (2.83) \\
QA-2 & MagicPIG & $(9,340)$ & 63.49 & $47.20\pm4.40$ & 5.95 & 69.45 & 4.44 (4.18) \\
QA-2 & MagicPIG & $(10,145)$ & 15.99 & $44.20\pm4.30$ & 1.23 & 17.22 & 1.95 (1.84) \\
QA-2 & MagicPIG & $(10,180)$ & 19.48 & $45.40\pm4.30$ & 1.54 & 21.02 & 2.42 (2.28) \\
QA-2 & MagicPIG & $(10,235)$ & 25.31 & $46.60\pm4.40$ & 2.05 & 27.36 & 3.16 (2.98) \\
QA-2 & MagicPIG & $(10,315)$ & 33.86 & $46.80\pm4.40$ & 2.80 & 36.66 & 4.24 (4.00) \\
\end{longtable}
\endgroup
\endgroup

\section{Triton decode kernels}
\label{app:triton-kernels}

\paragraph{{Team construction and cache layout.}}
{The systems implementation uses contiguous parents of $P=16$ keys and
up to $R=4$ actual-key representatives per parent.} {It first chooses the
key nearest the parent mean, then chooses the remaining representatives
by farthest-first selection.} {Each key joins its nearest representative;
ties are deterministic, and each representative owns itself.} {The last
partial parent follows the same rule with up to four representatives.}
{The nominal budget is $S=4K$, where $K$ is the number of selected teams
per query head.} {Team sizes vary, so $S$ is not the exact number of keys
read, and the union across GQA query heads can contain more rows.}

{After dense prefill, GPU preparation constructs teams and packs their
members into contiguous key and value arrays.} {Layer-wise packing
releases the corresponding chronological prompt cache after it has
been replaced.} {The resulting cache retains all prompt KV entries,
representatives, team metadata, and operator workspaces.} {It does not
retain a second full prompt KV cache or a full CPU mirror after
preparation.} {Temporary source and destination allocations can overlap
during packing.} {Generated keys and values occupy a separate exact suffix.}

\paragraph{1. Grouped representative scorer.}
{The first kernel computes each query's dot product with one
representative per team.} {Query heads that share a KV head are processed
together to reuse representative loads.} {Each query head receives its
own scores; values and other team members are not read at this stage.}

\paragraph{{2. Parallel top-$K$ selection.}}
{The second kernel scales the scores by $1/\sqrt{d}$, adds the logarithm
of team size, and adds independent Gumbel noise for each query-head--team
pair.} {It processes blocks of teams and retains local candidates for the
global top $K+1$.} {The extra candidate supplies the threshold used for
conditional inclusion correction; its team is not added to the sample.}
{The implementation supports $K\in\{31,32,64,128,256\}$ using
budget-specific tiling and scheduling.}

\paragraph{{3. Exact merge and GQA union.}}
{The third kernel merges the local candidates, recovers the global
top-$K$ teams and next score, and computes the selected teams'
conditional inclusion corrections (Equation~\ref{eq:method-inclusion}).}
{It also constructs the union of selected teams for query heads sharing
one KV head.} {The union records team membership and which query heads
selected each team.} {This lets the attention kernel share KV reads
without changing any head's sample or correction.}

\paragraph{{4. Selected-member attention.}}
{The fourth kernel evaluates exact query--key scores for the members
of selected teams.} {A head accumulates contributions only from its own
selected teams.} {Subtracting $\log c_g$ from each selected member's logit
applies inverse-inclusion correction to both its unnormalized attention
mass and value-weighted sum.} {Selected rows are split across parallel
work units, which produce partial sums and their softmax scales.}

\paragraph{{5. Split-softmax reduction.}}
{The fifth kernel rescales the partial sums to a common softmax scale,
adds the corrected masses and value-weighted sums, and divides to form
the output.} {The complete-operator benchmark includes all five launches.}
{In full-model generation, sampled prompt contributions and exact
generated-suffix contributions are combined under a joint normalization.}
{The integrated generation path uses $K=256$.}

\section{{Systems evaluation protocol and additional results}}
\label{app:kernel-evaluation}

\subsection{{Model, hardware, and workloads}}
\label{app:systems-setup}

{All measurements use one NVIDIA GeForce RTX 5090 Laptop GPU under WSL2,
with 24,463\,MiB reported device memory, 82 streaming multiprocessors,
and compute capability 12.0.} {The driver is 621.14 and the CUDA
runtime/toolkit is 13.0.} {We do not set persistent GPU clocks or power
limits.} {The measurements use no CUDA graphs.} {Operator experiments
disable TF32 matrix multiplication.}

{The full-precision model is Qwen2.5-7B-Instruct loaded in FP16.} {It has
28 layers, 28 query heads, four KV heads, head dimension 128, and hidden
size 3,584.} {The INT4 model is Qwen2.5-7B-Instruct-GPTQ-Int4, run with
GPTQModel/Marlin, group size 128, symmetric quantization, and
no activation-order permutation.} {FP32 Marlin reduction is enabled and atomics
are disabled.} {Activations, representatives, caches, and unquantized
model parameters use FP16; rotary-frequency buffers use FP32.} {The two
precision settings are separate within-checkpoint comparisons.}
{Table~\ref{tab:systems-software} lists the software environments.}

\begin{table}[!htbp]
\centering

\small
\caption{  Software versions used for the systems measurements.}
\label{tab:systems-software}
\begin{tabular}{lll}
\toprule
Component & FP16 & INT4 weights \\
\midrule
Python & 3.12.14 & 3.12.14 \\
PyTorch & 2.12.1+cu130 & 2.12.1+cu130 \\
Triton & 3.7.1 & 3.7.1 \\
Transformers & 5.12.1 & 5.14.0 \\
GPTQModel & -- & 7.3.4 \\
NumPy & 2.5.1 & 2.2.6 \\
safetensors & 0.6.2 & 0.8.0 \\
Nsight Compute & 2025.3.1 & 2025.3.1 \\
\bottomrule
\end{tabular}
\end{table}

{The timing set contains two prompts from each of six RULER tasks:
CWE, MK-NIAH-3, MV-NIAH,
S-NIAH-1, QA-1, and VT.}
{We choose prompts before timing by seeded per-task shuffling and
sorted-task round robin, using selection seed 20261114.} {Inputs are
filtered to 30,000--32,256 tokens so that 512 outputs fit within the
32,768-position model limit.} {No padding or truncation is applied.}
{The input lengths in Table~\ref{tab:systems-prompts} include Qwen
chat formatting and the source answer prefix; both model tokenizers
produce the same input sequences.} {These prompts are used for timing rather than an additional accuracy evaluation.}

\begin{table}[!htbp]
\centering

\small
\caption{  Prompts used for timing. Method order applies to full-model
paired measurements. The two-prompt INT4 batch-two result uses indices
0 and 4; the six-prompt short-generation control uses indices 0--5.}
\label{tab:systems-prompts}
\begin{tabular}{rlrl}
\toprule
Index & Task: example ID & Input tokens & First method \\
\midrule
0  & CWE:7                & 31,931 & Dense \\
1  & MK-NIAH-3:7  & 32,008 & Sparse \\
2  & MV-NIAH:4   & 32,026 & Dense \\
3  & S-NIAH-1:2    & 32,012 & Sparse \\
4  & QA-1:1             & 32,003 & Dense \\
5  & VT:1                & 31,889 & Sparse \\
6  & CWE:2               & 31,970 & Dense \\
7  & MK-NIAH-3:4  & 30,133 & Sparse \\
8  & MV-NIAH:1   & 32,026 & Dense \\
9  & S-NIAH-1:5    & 32,006 & Sparse \\
10 & QA-1:7             & 30,347 & Dense \\
11 & VT:0                & 31,892 & Sparse \\
\bottomrule
\end{tabular}
\end{table}

{The dense comparator explicitly forces PyTorch's Flash SDPA backend
with native GQA.} {Dense and sparse full-model runs use the same checkpoint,
precision, allocator settings, in-place prefill MLP, and native decode
MLP.} {For batch size $B>1$, we replicate the same prompt across $B$
sequences, each with its own KV cache and distinct request and sampling
identities. Prefill processes these sequences one at a time, followed by
jointly batched decoding. Each replicated batch counts as one prompt
workload.}

\subsection{{Timing scopes and statistical summaries}}
\label{app:systems-timing}

\paragraph{{Complete attention operator.}}
{For each prompt we capture the final prompt-token query and prompt KV
at layer 14, with batch one and no generated suffix.} {The query and KV
state are fixed while sampling realizations vary.} {We measure
$K\in\{31,32,64,128,256\}$ using five paired rounds, 50 warmup pairs
per round, and 100 timed invocations per method per round.} {Methods alternate within each round, with the starting method reversed
between rounds.} {A 512\,MiB read/write eviction buffer is touched before
each method's CUDA-event interval.} {Its execution is excluded from the
interval, and no eviction occurs between launches within one operator.}
{Team construction, packing, allocation, and input preparation are
excluded from operator timing.} {Sampling realizations vary across repetitions.}

{Within each prompt and budget, we take the median of 100 durations in
each round and then the median of the five round medians.} {Reported
latencies are arithmetic means of these twelve prompt centers.} {The
speedup is the geometric mean of the twelve paired Flash/sparse ratios.}
{Table~\ref{tab:kernel-e2e} gives the principal power-of-two budget curve.}
{The additional $K=31$, $S=124$ measurement is 84.840\,$\mu$s sparse
versus 144.879\,$\mu$s Flash, with speedup $1.7076\times$
[1.6822, 1.7319].} {Budget-specific tiling and scheduling change between
$K=31$ and $32$, and at larger budgets; so the $K=31$ to $32$ latency change also reflects a change in scheduling.}

\paragraph{{Full-model generation.}}
{Every measured request runs in a fresh worker after an excluded warmup
of 16 predictions with the target suffix capacity allocated.} {Greedy
generation continues to the fixed output count even if EOS is produced.}
{Dense and sparse generation can therefore follow different later
histories.} {The full request includes sequential dense prefill, sparse
team construction and packing, pointer/workspace setup, first-token
replay and selection, and feedback decoding.} {Model loading, tokenizer
and input loading, JIT/warmup, and profiling are excluded.}

{The decode interval is synchronized host wall time after the first
output is produced, covering all model layers for the remaining $G-1$
feedback steps.} {For $G=512$, decode latency per step is
$t_{\mathrm{decode}}/511$ and aggregate decode throughput is
$511B/t_{\mathrm{decode}}$.} {Full-request throughput is
$512B/t_{\mathrm{request}}$.} {Request time is for the complete batch.}
{The $G=128$ control uses 127 feedback steps.} {Both methods use
the native PyTorch allocator with expandable segments and reserve
128\,MiB of free-memory headroom.}

\paragraph{{Aggregation and intervals.}}
{Each retained full-model prompt contributes one paired dense/sparse
measurement.} {Absolute times and throughputs are arithmetic means across
prompts; speedups are equal-prompt geometric means of paired time
ratios.} {A ratio of displayed mean times therefore need not equal the
reported speedup.} {Twelve prompts are measured at every main-table
configuration; INT4 batch two has two prompts.} {The twelve-prompt runs
balance first-method order six/six, although both examples from each
task share an order (Table~\ref{tab:systems-prompts}).}

{All reported 95\% intervals use percentile bootstrapping over prompt
clusters, with 4,000 resamples and seed 20261114.} {Intervals resample prompts, not batch replicas or kernel calls.}
{They describe workload variation, not systematic clock uncertainty.} {No measurements are removed because a method was slower
or a timing-drift warning was raised.}

\subsection{{Generation latency, throughput, and preparation}}
\label{app:systems-generation}

{Tables~\ref{tab:systems-absolute}--\ref{tab:systems-preparation} give
absolute values for the operating points in
Table~\ref{tab:systems-generation}, including the two-prompt INT4
batch-two result.} {That additional configuration measures
$0.9363\times$ decode and $0.9730\times$ full-request speedup.}
{All rows use $S=1024$ and $G=512$.}

\begin{table}[!htbp]
\centering

\small
\caption{  Mean full-model latency. One decode step jointly produces
$B$ outputs; request time covers the whole batch. $N$ is the number of
independent prompts.}
\label{tab:systems-absolute}
\setlength{\tabcolsep}{5pt}
\begin{tabular}{lrrrrrr}
\toprule
& & & \multicolumn{2}{c}{Decode (ms/step)} & \multicolumn{2}{c}{Request (s)} \\
\cmidrule(lr){4-5}\cmidrule(lr){6-7}
Weights & $B$ & $N$ & Dense & Sparse & Dense & Sparse \\
\midrule
FP16 & 1 & 12 & 25.355 & 24.954 & 21.220 & 21.072 \\
FP16 & 2 & 12 & 27.105 & 26.448 & 30.345 & 30.057 \\
FP16 & 3 & 12 & 29.331 & 28.106 & 41.921 & 41.427 \\
INT4 & 1 & 12 & 14.894 & 18.427 & 15.632 & 17.446 \\
INT4 & 2 & 2  & 17.086 & 18.250 & 24.826 & 25.515 \\
INT4 & 4 & 12 & 21.533 & 19.034 & 42.820 & 41.670 \\
INT4 & 8 & 12 & 30.436 & 24.708 & 85.109 & 82.493 \\
\bottomrule
\end{tabular}
\end{table}

\begin{table}[!htbp]
\centering

\small
\caption{  Mean generated-token throughput in tokens/s, aggregated
over the complete batch. Coverage is as in
Table~\ref{tab:systems-absolute}.}
\label{tab:systems-throughput}
\begin{tabular}{lrrrrr}
\toprule
& & \multicolumn{2}{c}{Decode} & \multicolumn{2}{c}{Full request} \\
\cmidrule(lr){3-4}\cmidrule(lr){5-6}
Weights & $B$ & Dense & Sparse & Dense & Sparse \\
\midrule
FP16 & 1 & 39.44  & 40.08  & 24.13 & 24.30 \\
FP16 & 2 & 73.79  & 75.62  & 33.75 & 34.08 \\
FP16 & 3 & 102.29 & 106.74 & 36.65 & 37.09 \\
INT4 & 1 & 67.15  & 54.28  & 32.76 & 29.35 \\
INT4 & 2 & 117.06 & 109.60 & 41.25 & 40.13 \\
INT4 & 4 & 185.80 & 210.17 & 47.85 & 49.17 \\
INT4 & 8 & 262.89 & 323.87 & 48.26 & 49.79 \\
\bottomrule
\end{tabular}
\end{table}

{Sparse preparation includes team construction, packing, pointer setup,
and workspace allocation.} {First-token replay/selection is reported
separately and is also included in the request.} {Phase shares divide summed phase times by summed request times.}
{The directly measured request also includes selection and synchronization.} {Sequential prefill occupies most of the request at larger
batches; faster feedback decoding consequently produces a smaller
full-request gain.}

\begin{table}[!htbp]
\centering

\small
\caption{  Mean recurring request phases at $G=512$. Preparation and
replay columns apply to SANTA\texttt{++}. The final column is the sparse
prefill fraction of sparse request time. All preparation is included
in full-request timing.}
\label{tab:systems-preparation}
\setlength{\tabcolsep}{4.5pt}
\begin{tabular}{lrrrrrr}
\toprule
& & \multicolumn{2}{c}{Prefill (s)} & \shortstack{Preparation\\(ms)} & \shortstack{Replay\\(ms)} & \shortstack{Prefill\\share (\%)} \\
\cmidrule(lr){3-4}
Weights & $B$ & Dense & Sparse & & & \\
\midrule
FP16 & 1 & 8.262  & 8.252  & 35.808  & 31.505 & 39.16 \\
FP16 & 2 & 16.492 & 16.447 & 61.958  & 31.684 & 54.72 \\
FP16 & 3 & 26.931 & 26.909 & 118.038 & 35.282 & 64.96 \\
INT4 & 1 & 8.019  & 7.972  & 35.201  & 21.173 & 45.69 \\
INT4 & 2 & 16.093 & 16.104 & 62.492  & 20.932 & 63.12 \\
INT4 & 4 & 31.813 & 31.805 & 114.058 & 22.150 & 76.33 \\
INT4 & 8 & 69.552 & 69.554 & 283.499 & 25.201 & 84.32 \\
\bottomrule
\end{tabular}
\end{table}

\paragraph{{Full-request speedup and Amdahl's law.}}
\begingroup
The INT4 batch-eight timing breakdown illustrates how unchanged
prefill limits full-request gains. From
Tables~\ref{tab:systems-absolute} and~\ref{tab:systems-preparation},
the dense request takes $T_{\mathrm d}=85.109$\,s, of which
69.552\,s (81.7\%) is prefill. The 84.3\% share in
Table~\ref{tab:systems-preparation} instead uses the sparse request
as its denominator. The dense feedback-decoding fraction is
$f=(511\times0.030436)/85.109\simeq0.1827$.
Holding the remaining baseline time fixed, using decoding speedup
$s=1.232$, and adding the measured sparse preparation and replay
cost $h=0.283499+0.025201=0.308700$\,s gives the first-order estimate
\[
\frac{T_{\mathrm d}}{T_{\mathrm s}}
\approx
\frac{1}{(1-f)+f/s+h/T_{\mathrm d}}
\approx 1.032.
\]
This agrees with the reported full-request ratio and explains why
a 23\% decoding improvement produces about a 3\% request improvement
for these prefill-heavy workloads. The calculation summarizes the
measured phase times; it is not an independent latency measurement
or a prediction for other workloads. The timing limitations in
Appendix~\ref{app:systems-timing-limits} still apply.
\endgroup

\subsection{{Allocated memory and retained storage}}
\label{app:systems-memory}

{Table~\ref{tab:systems-peaks} reports maxima over per-request memory
high-water marks, reset before every measured request.} {Allocated
memory measures live allocations; reserved memory also includes
allocator capacity retained from warmup.} {Shared prefill temporaries
dominate the allocated peaks, so equal dense/sparse peaks do not imply
equal steady storage.} {Table~\ref{tab:systems-steady} reports memory
immediately before feedback decoding and separates the full prompt KV
payload from sparse representatives, metadata, and workspaces.}

\begin{table}[!htbp]
\centering

\small
\caption{  Peak GPU memory in GiB. Values are maxima over all measured
requests, with coverage as in Table~\ref{tab:systems-absolute}.}
\label{tab:systems-peaks}
\begin{tabular}{lrrrrr}
\toprule
& & \multicolumn{2}{c}{Peak allocated} & \multicolumn{2}{c}{Peak reserved} \\
\cmidrule(lr){3-4}\cmidrule(lr){5-6}
Weights & $B$ & Dense & Sparse & Dense & Sparse \\
\midrule
FP16 & 1 & 18.658 & 18.658 & 19.611 & 19.682 \\
FP16 & 2 & 20.396 & 20.396 & 21.352 & 21.486 \\
FP16 & 3 & 22.134 & 22.134 & 22.270 & 22.451 \\
INT4 & 1 & 9.642  & 9.642  & 10.637 & 10.682 \\
INT4 & 2 & 11.376 & 11.376 & 12.375 & 12.488 \\
INT4 & 4 & 14.857 & 14.857 & 15.852 & 16.098 \\
INT4 & 8 & 21.810 & 21.810 & 21.965 & 22.451 \\
\bottomrule
\end{tabular}
\end{table}

\begin{table}[!htbp]
\centering

\footnotesize
\caption{  Mean retained memory before feedback decoding. All columns
are GiB except workspace (MiB). The prompt KV payload is the same size
for dense and sparse. The allocation difference includes sparse
representatives, metadata, pointer tables, and workspace.}
\label{tab:systems-steady}
\setlength{\tabcolsep}{4pt}
\begin{tabular}{lrrrrrrr}
\toprule
Weights & $B$ & \shortstack{Prompt\\KV} & \shortstack{Dense\\allocated} & \shortstack{Sparse\\allocated} & \shortstack{Sparse\\$-$dense} & \shortstack{Reps. +\\metadata} & \shortstack{Workspace\\(MiB)} \\
\midrule
FP16 & 1 & 1.692  & 15.937 & 16.188 & 0.252 & 0.245 & 7.017 \\
FP16 & 2 & 3.385  & 17.657 & 18.160 & 0.503 & 0.489 & 14.034 \\
FP16 & 3 & 5.077  & 19.377 & 20.132 & 0.755 & 0.734 & 21.051 \\
INT4 & 1 & 1.692  & 6.916  & 7.167  & 0.252 & 0.245 & 7.017 \\
INT4 & 2 & 3.414  & 8.666  & 9.173  & 0.508 & 0.494 & 14.098 \\
INT4 & 4 & 6.769  & 12.076 & 13.083 & 1.006 & 0.979 & 28.068 \\
INT4 & 8 & 13.538 & 18.957 & 20.969 & 2.012 & 1.957 & 56.135 \\
\bottomrule
\end{tabular}
\end{table}

{At batch one the mean additional steady allocation is approximately
0.252\,GiB for the full model.} {It grows with batch size because each
sequence has its own cache.} {Prompt KV remains fully resident in
packed form, and the exact suffix has capacity for $G$ tokens.} 

\subsection{{Short-generation control}}
\label{app:systems-short-generation}

{We repeat FP16 batch-three timing at $G=128$ on the first six prompts
in Table~\ref{tab:systems-prompts}.} {Table~\ref{tab:systems-short}
compares these six prompts with their $G=512$ measurements.} {The
128-output full-request interval includes one, whereas the longer
generation has a positive observed request gain.} {Longer outputs better amortize prefill.} {Single-prompt batch-one controls at
$G=128$ measured decode/request ratios of $1.0068/1.0217$ for FP16 and
$0.8371/0.9661$ for INT4 weights.} {No high-batch INT4 short-generation
measurement was collected.}

\begin{table}[!htbp]
\centering

\small
\caption{  Matched six-prompt FP16 batch-three comparison at $S=1024$.
Intervals are 95\% prompt-bootstrap intervals. Both output lengths use
the same six input prompts, measured in separate measurement sessions.}
\label{tab:systems-short}
\begin{tabular}{rrrr}
\toprule
Outputs & Decode speedup & Request speedup & Sparse prefill share \\
\midrule
128 & $1.0602\times$ [1.0526, 1.0679] & $1.0041\times$ [0.9974, 1.0101] & 87.92\% \\
512 & $1.0459\times$ [1.0421, 1.0503] & $1.0132\times$ [1.0121, 1.0142] & 65.15\% \\
\bottomrule
\end{tabular}
\end{table}

\subsection{{Selected rows and isolated stages}}
\label{app:systems-stages}

{Table~\ref{tab:systems-rows} reports selected rows per query head and
the shared union per KV head.} {Counts average the final sampling
realization of each round across heads and prompts.} {They exclude
representative and metadata reads and therefore are not a total
logical-access percentage.} {Table~\ref{tab:kernel-stage} separately
times each stage with 100 repetitions per round and an entry eviction
before each stage.} {The stage measurements have different cache
boundaries from the complete operation and should not be added to
reconstruct its latency.}

\begin{table}[!htbp]
\centering

\small
\caption{  Mean selected-member rows and GQA unions in the operator
sweep. $S=4K$ is a nominal budget rather than an exact row count.}
\label{tab:systems-rows}
\begin{tabular}{rrrr}
\toprule
$K$ & $S$ & Selected rows/query head & Union rows/KV head \\
\midrule
31  & 124  & 127.94  & 591.53 \\
32  & 128  & 132.10  & 611.22 \\
64  & 256  & 267.54  & 1229.05 \\
128 & 512  & 539.25  & 2347.60 \\
256 & 1024 & 1088.14 & 4276.62 \\
\bottomrule
\end{tabular}
\end{table}

\begin{table}[!htbp]
\centering

\small
\caption{  Isolated-stage latency in $\mu$s. These separately measured
times do not form an additive decomposition of
Table~\ref{tab:kernel-e2e}.}
\label{tab:kernel-stage}
\begin{tabular}{rrrrrr}
\toprule
$S$ & Scorer & Top-$K$ & Merge/union & Attention & Reducer \\
\midrule
124  & 25.557 & 20.261 & 21.015 & 22.272 & 9.996 \\
128  & 23.028 & 22.331 & 22.225 & 16.895 & 10.012 \\
256  & 23.041 & 40.359 & 14.052 & 18.863 & 12.096 \\
512  & 23.109 & 41.951 & 18.425 & 21.987 & 12.057 \\
1024 & 24.316 & 45.376 & 12.747 & 26.915 & 12.076 \\
\bottomrule
\end{tabular}
\end{table}

\subsection{{Physical DRAM traffic}}
\label{app:systems-dram}

{We use Nsight Compute 2025.3.1 application-range replay to measure
combined DRAM read and write bytes over the uninterrupted attention
operator.} {The accepted ranges contain all five sparse launches or
both Flash launches, with no intervening checks, copies, or eviction.}
{Profiler-managed cache flushing and clock locking are disabled;
dynamic pipeline boost is enabled.} {A 512\,MiB
read/write eviction and synchronization occur before range entry.}
{Each capture has no generated suffix and uses a fixed sampling
realization.}

\begin{table}[!htbp]
\centering

\small
\caption{  Combined DRAM read/write traffic within complete-operator
ranges. The first three rows are medians of three profiles per method;
the final row is a separate single paired profile. Ratios are medians
of paired ratios.}
\label{tab:ncu-kernel-summary}
\label{tab:systems-dram}
\setlength{\tabcolsep}{4pt}
\begin{tabular}{lrrrrr}
\toprule
State & $S$ & Pairs & Sparse (MiB) & Flash (MiB) & Flash/sparse \\
\midrule
CWE:7 & 128  & 3 & 9.444  & 62.526 & $6.6209\times$ \\
CWE:7 & 512  & 3 & 13.716 & 62.752 & $4.5716\times$ \\
CWE:7 & 1024 & 3 & 18.416 & 62.547 & $3.3954\times$ \\
MK-NIAH-3:7 & 1024 & 1 & 16.278 & 62.667 & $3.8499\times$ \\
\bottomrule
\end{tabular}
\end{table}

{The first state has 31,931 prompt tokens and the second has 32,008;
both are captured at layer 14.} {These counters measure combined
physical read/write bytes incurred within the instrumented range.}
{Deferred dirty writebacks from the entry eviction and asynchronous
chip-wide activity can contribute, so the values are not isolated
KV-read bytes or full-model traffic.} {We report the two states separately; operator latency is measured
without the profiler.}

{For comparison, separately replaying individual kernels with
profiler cache flushing gives summed stage-median traffic of
10.851, 16.544, and 21.686\,MiB for SANTA\texttt{++} at
$S=128,512,1024$, versus 63.201, 63.155, and 63.222\,MiB for Flash.}
{These stage totals use different inter-kernel cache conditions and do
not replace the uninterrupted range measurements.}

\subsection{{Implementation numerical validation}}
\label{app:systems-numerics}

{We compare optimized outputs with an independent sampled-attention
reference that uses the same selected teams and corrections, evaluates
query--key products and softmax in FP64, and returns FP32 outputs.}
{All sixty operator prompt/budget checks pass.}

{All checked INT4 batch-four/eight attention outputs satisfy the
mixed tolerance $|y-\widehat y|\leq10^{-4}(1+|\widehat y|)$.}
{Table~\ref{tab:systems-numerics} reports failures under this tolerance
and the stricter $2\times10^{-5}(1+|\widehat y|)$ criterion.}

{For each batch size, all twelve prompts are checked at layers 0, 14,
and 27 and decode steps 0, 1, 33, and 511 for every sequence in the batch.} {We also check
team construction and repeat complete sparse generations for
determinism.} {One prompt per batch size receives the full configuration
check, including dense-prefix and MLP parity.}

\begin{table}[!htbp]
\centering

\small
\caption{  INT4-weight sampled-output validation over twelve prompts
per batch size. A comparison contains 3,584 correlated output components.
Failures are counted under both mixed absolute/relative tolerances.}
\label{tab:systems-numerics}
\setlength{\tabcolsep}{4pt}
\begin{tabular}{rrrrr}
\toprule
$B$ & Comparisons & \shortstack{$2\times10^{-5}$ failures\\comparisons / elements} & \shortstack{$10^{-4}$ failures\\comparisons / elements} & Maximum absolute error \\
\midrule
4 & 576  & 4 / 9  & 0 / 0 & $4.6015\times10^{-5}$ \\
8 & 1152 & 3 / 41 & 0 / 0 & $7.3195\times10^{-5}$ \\
\bottomrule
\end{tabular}
\end{table}

{No checked output is nonfinite.} {The largest absolute errors are
$4.60\times10^{-5}$ at batch four and $7.32\times10^{-5}$ at batch
eight.} {Small FP32 differences can also change the rounded FP16 output.}
{These tests validate the implementation against sampled attention;
benchmark accuracy measures its approximation to dense attention.}

\subsection{{Timing variability and measurement limits}}
\label{app:systems-timing-limits}

{GPU clocks are not locked. All 60 prompt--budget timing runs
record more than 5\% drift between their first and second halves and
more than 10\% variation across round medians; every round is retained.
Methods alternate within each round, and the starting method reverses
between rounds so that neither method is consistently timed earlier.}
{Table~\ref{tab:systems-cache} separately compares the 512\,MiB
eviction buffer with a 4\,MiB light flush on two states. Absolute latency
also varies between sessions with the same buffer size. Nevertheless,
the $S=1024$ attention speedup is $1.38$--$1.41\times$ across both
states and buffer sizes, supporting an operator-level benefit under
these tested conditions.}

\begin{table}[!htbp]
\centering

\small
\caption{  Separate two-state cache-condition diagnostic. These measurements are not pooled with Table~\ref{tab:kernel-e2e}.}
\label{tab:systems-cache}
\setlength{\tabcolsep}{4pt}
\begin{tabular}{lrrrrr}
\toprule
State & $S$ & \shortstack{Eviction\\(MiB)} & \shortstack{Flash\\($\mu$s)} & \shortstack{Sparse\\($\mu$s)} & Speedup \\
\midrule
CWE:7 & 128 & 512 & 103.200 & 55.296 & $1.8663\times$ \\
CWE:7 & 128 & 4   & 92.384  & 53.280 & $1.7339\times$ \\
CWE:7 & 1024 & 512 & 103.200 & 73.472 & $1.4046\times$ \\
CWE:7 & 1024 & 4 & 93.920 & 67.872 & $1.3838\times$ \\
MK-NIAH-3:7 & 128 & 512 & 147.136 & 82.064 & $1.7929\times$ \\
MK-NIAH-3:7 & 128 & 4 & 140.800 & 82.192 & $1.7131\times$ \\
MK-NIAH-3:7 & 1024 & 512 & 146.816 & 104.256 & $1.4082\times$ \\
MK-NIAH-3:7 & 1024 & 4 & 140.736 & 100.464 & $1.4009\times$ \\
\bottomrule
\end{tabular}
\end{table}

{We observed discrepancies between timing sources.} {Summed
per-step CUDA-event times exceed synchronized host-wall decode time;
elapsed UTC time exceeds the monotonic timer by about
2.0--2.35\% over the full experiment commands.} {Their effect on paired request
ratios is unknown.} {We report host-wall measurements without correction,
and the bootstrap intervals exclude this possible systematic error.}
{This matters especially for the small full-request gains.}

{The evaluation covers one device and model family, with one pair per
prompt, replicated batches, and sequential prefill.} {The results are
not a benchmark of heterogeneous or continuous serving.}

\section{Token sparsity after latent compression}
\label{app:mla}

Multi-head latent attention (MLA) in DeepSeek-V2-Lite-Chat already
compresses the representation stored for each token and shares it
across all 16 attention heads. Each cache row contains one
512-dimensional normalized content latent and one 64-dimensional
rotary-position key. We test whether useful token sparsity remains
after this compression and sharing. SANTA\texttt{++} selects which
token rows to evaluate for a query rather than further compressing
each row. With a shared cache, a small per-head selection need not
give a small shared-row union: a row is omitted from selected
attention only when no head selects it.

We evaluate four budgets, $S\in\{128,256,512,1024\}$, on three RULER
tasks at 8K context, using 50 matched prompts per task. The dense
reference uses SDPA through the same compressed-cache adapter and
precision settings as SANTA\texttt{++}. Access is counted in shared
latent rows, including both routing and the union of selected rows
across heads, as defined below.

\paragraph{Accuracy--access trade-off.}
At $S=512$, SANTA\texttt{++} reaches a mean accuracy of 84.56\% versus
86.22\% for SDPA, retaining 98.07\% of the dense mean with 86.18\%
logical shared-row access (Table~\ref{tab:ruler_mla_8192}). This reduces logical access by a further 13.82\% relative to an
already shared latent cache. A lower budget trades more accuracy
for fewer reads: $S=256$ retains 91.56\% of the dense mean at 70.09\%
logical access. {These results suggest that selective token access
may remain useful after latent compression and head sharing.}

\begin{table}[!htb]
    \centering
    \small
    \setlength{\tabcolsep}{4.5pt}
    \begin{tabular}{l c c c c}
        \toprule
        Method & $S$ & KV (\%) & Acc (norm) (\%) & Acc (abs) (\%) \\
        \midrule
        SDPA & --- & 100.00 & 100.00 & 86.22 $\pm$ 4.56 \\
        \midrule
        \multirow{4}{*}{\shortstack{\textbf{SANTA\texttt{++}}\\(Latent k-means)}} & 128 & 56.28 & 80.80 & 69.67 $\pm$ 5.92 \\
         & 256 & 70.09 & 91.56 & 78.94 $\pm$ 5.14 \\
         & 512 & 86.18 & 98.07 & 84.56 $\pm$ 4.72 \\
         & 1024 & 101.89 & 97.81 & 84.33 $\pm$ 5.00 \\
        \bottomrule
    \end{tabular}
    \caption{\textbf{Token sparsity in a shared-latent cache at 8K.}
    DeepSeek-V2-Lite-Chat, with 50 matched prompts per task on
    MV-NIAH, FWE, and QA-1.
    Absolute accuracy is the equal-weight task mean; normalized accuracy
    divides it by the SDPA mean, times 100. KV denotes logical shared-row
    access, including routing, under Equation~\ref{eq:mla-logical-access}. Throughout these tables, $\pm$
    gives half the span of a 95\% percentile-bootstrap interval using
    10,000 resamples. Prompts are resampled within each task, retaining
    task strata for the aggregate. Interval endpoints need not be
    symmetric about the point estimate.}
    \label{tab:ruler_mla_8192}
\end{table}

\paragraph{Shared-row accounting.}
For a layer-level decoding call $c$, let $\mathcal I_{c,h}$ contain
head $h$'s selected prompt members and exact generated suffix. Let
$r_c$ charge one routing row per representative, and let $d_c$ be the
current dense shared-row count. We report
\begin{equation}
 \mathrm{KV}_{\mathrm{MLA}}\ (\%)
 =100\,\frac{\sum_c\left(r_c+
       \left|\bigcup_h\mathcal I_{c,h}\right|\right)}
                  {\sum_c d_c}.
 \label{eq:mla-logical-access}
\end{equation}
One counted row represents a token's shared content latent and
rotary-position key, without separate charges for an expanded key
and value. The union counts an attention row once across heads
without changing any head's estimator. A row used in both routing
and attention is counted in both passes. Counts are pooled across
prompts, layers, and decoding steps, and across tasks for the
aggregate, before division.

Each method's denominator is the dense shared-row count along its own
generation trajectory, not the total from the separate SDPA run.
Generation length and the sparse first-token replay described below
can therefore change the denominator. The reference implementation
stores expanded routing keys and reconstructs full K/V before
selection. These measurements describe logical access opportunities,
not realized memory-traffic savings or a latent-cache speedup.

\paragraph{Head sharing and routing limit the reduction.}
Independent head selections can cover a large fraction of the shared
cache even when each head selects few teams. At $S=512$, the attention
union and exact suffix account for 63.75\% of the dense shared-row
count, and routing adds 22.43\%. At $S=1024$, the attention term grows
to 79.51\%; including routing raises total logical access to 101.89\%,
without improving the aggregate accuracy point estimate. A value above
100\% is possible because routing and attention are separate passes.
Thus, increasing the per-head budget can consume the remaining
shared-row savings before it improves task accuracy.

\paragraph{Task dependence.}
The additional sparsity is useful on some tasks at a smaller accuracy
cost than on others (Tables~\ref{tab:ruler_mla_8192_fwe}--\ref{tab:ruler_mla_8192_qa_1}).
At $S=512$, FWE matches SDPA's 96.67\% score with 80.34\%
logical access. Multivalue retrieval is one percentage point below
its dense score, whereas QA-1 remains four points below.
Increasing the budget to $S=1024$ raises the QA-1 point
estimate but lowers the multivalue-retrieval estimate. The full sweep
therefore shows a task-dependent trade-off rather than a monotonic
accuracy gain from reading more rows.

\begin{table}[!htb]
    \centering
    \small
    \setlength{\tabcolsep}{4.5pt}
    \begin{tabular}{l c c c c}
        \toprule
        Method & $S$ & KV (\%) & Acc (norm) (\%) & Acc (abs) (\%) \\
        \midrule
        SDPA & --- & 100.00 & 100.00 & 96.67 $\pm$ 2.67 \\
        \midrule
        \multirow{4}{*}{\shortstack{\textbf{SANTA\texttt{++}}\\(Latent k-means)}} & 128 & 47.57 & 84.82 & 82.00 $\pm$ 7.00 \\
         & 256 & 62.25 & 94.48 & 91.33 $\pm$ 5.33 \\
         & 512 & 80.34 & 100.00 & 96.67 $\pm$ 3.00 \\
         & 1024 & 98.45 & 99.31 & 96.00 $\pm$ 3.00 \\
        \bottomrule
    \end{tabular}
    \caption{\textbf{RULER FWE at 8K.}
    Accuracy on 50 matched prompts per method. Normalized accuracy
    divides the reported two-decimal task score by SDPA's score on
    this task, times 100. Access and uncertainty follow
    Table~\ref{tab:ruler_mla_8192}.}
    \label{tab:ruler_mla_8192_fwe}
\end{table}

\begin{table}[!htb]
    \centering
    \small
    \setlength{\tabcolsep}{4.5pt}
    \begin{tabular}{l c c c c}
        \toprule
        Method & $S$ & KV (\%) & Acc (norm) (\%) & Acc (abs) (\%) \\
        \midrule
        SDPA & --- & 100.00 & 100.00 & 94.00 $\pm$ 4.00 \\
        \midrule
        \multirow{4}{*}{\shortstack{\textbf{SANTA\texttt{++}}\\(Latent k-means)}} & 128 & 57.88 & 77.66 & 73.00 $\pm$ 8.75 \\
         & 256 & 71.89 & 93.09 & 87.50 $\pm$ 5.50 \\
         & 512 & 87.67 & 98.94 & 93.00 $\pm$ 4.75 \\
         & 1024 & 102.64 & 92.55 & 87.00 $\pm$ 7.75 \\
        \bottomrule
    \end{tabular}
    \caption{\textbf{RULER MV-NIAH at 8K.}
    Accuracy on 50 matched prompts per method. Normalized accuracy
    divides the reported two-decimal task score by SDPA's score on
    this task, times 100. Access and uncertainty follow
    Table~\ref{tab:ruler_mla_8192}.}
    \label{tab:ruler_mla_8192_niah_multivalue}
\end{table}

\begin{table}[!htb]
    \centering
    \small
    \setlength{\tabcolsep}{4.5pt}
    \begin{tabular}{l c c c c}
        \toprule
        Method & $S$ & KV (\%) & Acc (norm) (\%) & Acc (abs) (\%) \\
        \midrule
        SDPA & --- & 100.00 & 100.00 & 68.00 $\pm$ 13.00 \\
        \midrule
        \multirow{4}{*}{\shortstack{\textbf{SANTA\texttt{++}}\\(Latent k-means)}} & 128 & 58.41 & 79.41 & 54.00 $\pm$ 14.00 \\
         & 256 & 72.81 & 85.29 & 58.00 $\pm$ 14.00 \\
         & 512 & 88.50 & 94.12 & 64.00 $\pm$ 13.00 \\
         & 1024 & 103.87 & 102.94 & 70.00 $\pm$ 13.00 \\
        \bottomrule
    \end{tabular}
    \caption{\textbf{RULER QA-1 at 8K.}
    Accuracy on 50 matched prompts per method. Normalized accuracy
    divides the reported two-decimal task score by SDPA's score on
    this task, times 100. Access and uncertainty follow
    Table~\ref{tab:ruler_mla_8192}.}
    \label{tab:ruler_mla_8192_qa_1}
\end{table}

\paragraph{Model, precision, and generation.}
Weights use bitsandbytes LLM.int8 quantization, except for the
latent-to-K/V projection and language-model output head. These
unquantized modules and the cache use BF16. Routing scores and
sampled attention reductions use FP32. The dense reference uses
SDPA through the same compressed-cache adapter, rather than the
unmodified checkpoint's generation path. SDPA may select its Flash
or memory-efficient backend; the math backend is disabled.

All methods use the same prompts, answer references, and chat
formatting. Decoding is greedy and stops at the end-of-sequence
token or at 50, 128, and 32 new tokens for FWE,
MV-NIAH, and QA-1, respectively.
Data generation and prompt selection use seed 42, and k-means uses
seed 0. Sampling uses deterministic prompt-specific seeds derived
from base seed 0 and each prompt's identifier, shared between
budgets. Prompt scores count the fraction of required answers
recovered for FWE and MV-NIAH;
QA-1 checks whether any accepted answer appears in the output.

\paragraph{Teams and head-specific selection.}
Minibatch k-means and representative assignment use the content
latents after the checkpoint's normalization, without additional
unit-norm normalization or the rotary components. This produces one
partition per layer shared by all heads. Representatives are actual
members selected within each parent, and each representative is
assigned to its own team.

Each head scores a representative using its query and that head's
reconstructed representative key, including the rotary-position
component, with the checkpoint's attention scale. These routing
keys are frozen after dense prefill. Selected contributions use
current-cache keys and values. Heads share teams but draw
independently and retain their own conditional inclusion corrections
(Section~\ref{sec:method}). We use $P=16$ and $R=4$, so the four nominal
budgets request 32, 64, 128, and 256 teams per head, respectively,
capped by availability. 

\paragraph{Exact generated suffix.}
The fixed teams cover the entire prompt. After dense prefill, the
sparse adapter removes the last prompt entry from the cache and
replays that token at its original position through sparse attention.
The first generated token therefore depends on sparse attention as
well. There is no exact prompt tail: the exact suffix is empty on
the first sparse call and grows with newly generated tokens, which
are not added to the fixed teams.

\FloatBarrier

\clearpage
\section{{Accuracy with INT4 model weights}}
\label{app:int4-accuracy}
\begingroup

We evaluate the INT4-weight configuration used in the systems study
on a new matched RULER comparison. This experiment measures the
accuracy cost of sparse attention within the quantized checkpoint;
it does not measure the effect of weight quantization relative to
FP16 weights. KV caches remain FP16.

\subsection{{Evaluation protocol}}
\label{app:int4-accuracy-protocol}

\paragraph{Model and generation.}
Both methods use Qwen2.5-7B-Instruct-GPTQ-Int4
(checkpoint revision \texttt{e9c932a}) with the GPTQModel/Marlin
configuration, hardware, and INT4 software environment of
Appendix~\ref{app:systems-setup}. We compare forced dense Flash SDPA
against the Triton SANTA++ implementation with contiguous parents,
$P=16$, $R=4$, and $K=256$ ($S=1024$) in all 28 layers.
Activations, representatives, and KV caches use FP16. Both methods
use the same dense prefill and in-place prefill MLP. SANTA++ retains
the inclusion correction, replays the final prompt token for its
first prediction, and attends exactly to the growing generated suffix.

We use greedy token generation and RULER's task-specific output caps:
128 tokens for NIAH, 30 for VT, 120 for CWE, 50 for FWE, and 32 for QA.
Each sequence is scored only up to its first EOS, excluding that token
and any subsequent output. These accuracy runs differ from the fixed
512-output timing protocol and do not provide a new speedup measurement.

\paragraph{Data and prompt selection.}
We use the 13-task RULER suite~\citep{hsieh2024ruler} and generate
50 examples per task. The official generator and evaluator use
revision \texttt{db43f92}; the generation seed is 20261117.
We apply the Qwen chat template once to each source input and append
its original answer prefix after the assistant cue. Eligibility is
fixed before evaluating either method: the rendered input must contain
at least 30,000 tokens, and input plus the task's output cap must fit
within 32,768 positions. No input is padded or truncated.
Nine generated examples fall below the minimum length: two FWE,
three QA-1, and four QA-2 examples. The remaining 641 inputs contain
30,022--32,712 tokens (mean 32,267.73). All 641 have completed matched
batch-one evaluations; no example is removed based on its score.
These are newly generated inputs, not the FP16 evaluation set in
Appendix~\ref{app:ruler}.

\paragraph{Batch-eight coverage.}
The batch-eight implementation groups equal-length inputs with matching
output caps. Unfilled batch positions contain duplicate prompts with
separate sampling identities; these filler outputs are not scored or
counted as additional examples. To increase distinct-prompt coverage,
we evaluate the full eligible pool at batch one and a smaller subset
at batch eight. Prompt order and the sampling seed are fixed before
scoring, with stable per-prompt sampling identities across batch sizes.
The batch-eight evaluation follows a fixed task-interleaved order.
At the time limit, 93 of 104 planned distinct examples have both
predictions, one attempt is incomplete, and ten have not been started.
We report only completed pairs, giving 6--8 distinct
examples per task. All 93 are also included in the batch-one evaluation.

\paragraph{Scoring and uncertainty.}
We use the official case-insensitive substring metrics and prediction
postprocessing: QA accepts any reference answer, while the other tasks
award the fraction of required answers recovered. Fractional scores
are retained before aggregation. We first average prompts within each
task and then weight all 13 task means equally, despite unequal sample
counts. For 95\% percentile intervals, we resample matched dense/sparse
prompt pairs within each task 4,000 times with seed 20261117 and
recompute the equal-task mean. The experiment uses one sparse-routing
seed; these intervals describe prompt variation, not variation across
routing seeds.

\subsection{{Aggregate and task-level results}}
\label{app:int4-accuracy-results}

\begin{table}[!htbp]
\centering

\small
\setlength{\tabcolsep}{4pt}
\caption{ Matched RULER comparison with INT4 model weights
and FP16 KV caches. Both rows cover all 13 tasks. $N$ counts distinct
completed pairs, not batch positions. Scores are equal-task means;
$\Delta$ is SANTA++ minus dense in percentage points, calculated before
rounding. Intervals use paired, task-stratified prompt bootstrapping.}
\label{tab:int4-accuracy-aggregate}
\begin{tabular}{@{}rrrrrl@{}}
\toprule
Batch & $N$ & \shortstack{Dense\\(\%)} &
\shortstack{SANTA++\\(\%)} & $\Delta$ & 95\% interval for $\Delta$ \\
\midrule
1 & 641 & 84.09 & 80.77 & $-3.33$ & $[-5.14,-1.57]$ \\
8 & 93  & 86.87 & 85.64 & $-1.22$ & $[-6.05,+3.61]$ \\
\bottomrule
\end{tabular}
\end{table}

Table~\ref{tab:int4-accuracy-aggregate} reports a 3.33-percentage-point
aggregate reduction in the larger batch-one evaluation. The smaller
batch-eight evaluation directly exercises the configuration used for
the reported decoding speedup, but its uncertainty does not establish
accuracy equivalence. Task-level scores and sample counts appear in
Table~\ref{tab:int4-accuracy-tasks}; the aggregate does not imply
uniform retention across tasks.

\begin{table}[!htbp]
\centering

\small
\setlength{\tabcolsep}{4.5pt}
\caption{ Task-level INT4-weight RULER results at $S=1024$.
Scores are percentages. Each $N$ counts distinct prompts evaluated by
both methods at that batch size. The final row weights tasks equally;
the batch-eight subset overlaps the batch-one evaluation.}
\label{tab:int4-accuracy-tasks}
\begin{tabular}{@{}lrrrrrr@{}}
\toprule
& \multicolumn{3}{c}{Batch one} & \multicolumn{3}{c}{Batch eight} \\
\cmidrule(lr){2-4}\cmidrule(lr){5-7}
Task & $N$ & Dense & SANTA++ & $N$ & Dense & SANTA++ \\
\midrule
S-NIAH-1  & 50 & 100.00 & 100.00 & 8 & 100.00 & 100.00 \\
S-NIAH-2  & 50 & 100.00 & 100.00 & 7 & 100.00 & 100.00 \\
S-NIAH-3  & 50 & 100.00 & 100.00 & 7 & 100.00 & 85.71 \\
MK-NIAH-1 & 50 & 96.00 & 94.00 & 8 & 100.00 & 100.00 \\
MK-NIAH-2 & 50 & 96.00 & 94.00 & 7 & 100.00 & 100.00 \\
MK-NIAH-3 & 50 & 82.00 & 56.00 & 7 & 100.00 & 85.71 \\
MQ-NIAH   & 50 & 99.00 & 96.00 & 8 & 100.00 & 96.88 \\
MV-NIAH   & 50 & 90.50 & 81.00 & 8 & 87.50 & 81.25 \\
VT        & 50 & 94.00 & 94.80 & 8 & 100.00 & 100.00 \\
CWE       & 50 & 51.20 & 43.20 & 7 & 52.86 & 47.14 \\
FWE       & 48 & 79.17 & 75.00 & 6 & 72.22 & 66.67 \\
QA-1      & 47 & 55.32 & 65.96 & 6 & 83.33 & 100.00 \\
QA-2      & 46 & 50.00 & 50.00 & 6 & 33.33 & 50.00 \\
\midrule
Equal-task mean & 641 & 84.09 & 80.77 & 93 & 86.87 & 85.64 \\
\bottomrule
\end{tabular}
\end{table}

\paragraph{Same-input batch-size check.}
Restricting both batch sizes to the same 93 prompts gives a dense
aggregate of 86.87\% at both batch sizes. SANTA++ scores 86.52\% at
batch one and 85.64\% at batch eight, a difference of
$-0.88$ percentage points. The scored token sequences match across
batch sizes for 91 of 93 dense outputs and 81 of 93 sparse outputs.
Thus, the difference between the two full rows of
Table~\ref{tab:int4-accuracy-aggregate} cannot be attributed to batch
size alone: they use different prompt subsets.

\paragraph{Numerical checks.}
We use the sampled-attention reference and numerical criteria described
in Appendix~\ref{app:systems-numerics}. An additional batch-eight
diagnostic uses a separate 31,931-token prompt and 512 outputs, with
96 output comparisons at selected layers and decoding steps. Two
comparisons (18 output components) exceed the stricter mixed tolerance
$2\times10^{-5}(1+|\widehat y|)$; all satisfy
$10^{-4}(1+|\widehat y|)$, and none are nonfinite. A separate batch-one
diagnostic on a 32,027-token prompt passes the stricter criterion at
its checked states. These implementation checks are distinct from the
downstream accuracy comparison above.

\FloatBarrier
\endgroup

\end{document}

%% file: math_commands.tex
\usepackage{amsmath,amsfonts,bm}

\def\eqref#1{equation~\ref{#1}}

\def\1{\bm{1}}

\DeclareMathAlphabet{\mathsfit}{\encodingdefault}{\sfdefault}{m}{sl}
\SetMathAlphabet{\mathsfit}{bold}{\encodingdefault}{\sfdefault}{bx}{n}



%% file: main-bib.bib
@article{kim2026fastkvzip,
  title={Fast kvzip: Efficient and accurate llm inference with gated kv eviction},
  author={Kim, Jang-Hyun and Han, Dongyoon and Yun, Sangdoo},
  journal={arXiv preprint arXiv:2601.17668},
  year={2026}
}

@inproceedings{
hisa2026,
title={{HISA}: Efficient Hierarchical Indexing for Fine-Grained Sparse Attention},
author={Yufei Xu and Fanxu Meng and Fan Jiang and Yuxuan Wang and Ruijie Zhou and Zhaohui Wang and Jiexi Wu and Zhixin Pan and Xiaojuan Tang and Wenjie Pei and Tongxuan Liu and di yin and Xing Sun and Muhan Zhang},
booktitle={Third Conference on Language Modeling},
year={2026},
url={https://openreview.net/forum?id=GnWqiwzNzK}
}

@inproceedings{kool2019stochastic,
  title={Stochastic beams and where to find them: The gumbel-top-k trick for sampling sequences without replacement},
  author={Kool, Wouter and Van Hoof, Herke and Welling, Max},
  booktitle={International conference on machine learning},
  pages={3499--3508},
  year={2019},
  organization={PMLR}
}

@inproceedings{
xiao2024infllm,
title={Inf{LLM}: Training-Free Long-Context Extrapolation for {LLM}s with an Efficient Context Memory},
author={Chaojun Xiao and Pengle Zhang and Xu Han and Guangxuan Xiao and Yankai Lin and Zhengyan Zhang and Zhiyuan Liu and Maosong Sun},
booktitle={The Thirty-eighth Annual Conference on Neural Information Processing Systems},
year={2024},
url={https://openreview.net/forum?id=bTHFrqhASY}
}

@InProceedings{tang2024quest,
  title     = {{QUEST}: Query-Aware Sparsity for Efficient Long-Context {LLM} Inference},
  author    = {Tang, Jiaming and Zhao, Yilong and Zhu, Kan and Xiao, Guangxuan and Kasikci, Baris and Han, Song},
  booktitle = {Proceedings of the 41st International Conference on Machine Learning},
  pages     = {47901--47911},
  year      = {2024},
  volume    = {235},
  series    = {Proceedings of Machine Learning Research},
  publisher = {PMLR},
  url = 	 {https://proceedings.mlr.press/v235/tang24l.html}
}

@inproceedings{chen2025magicpig,
  title     = {{MagicPIG}: {LSH} Sampling for Efficient {LLM} Generation},
  author    = {Chen, Zhuoming and Sadhukhan, Ranajoy and Ye, Zihao and Zhou, Yang and Zhang, Jianyu and Nolte, Niklas and Tian, Yuandong and Douze, Matthijs and Bottou, Leon and Jia, Zhihao and Chen, Beidi},
  booktitle = {International Conference on Learning Representations},
  pages     = {44169--44190},
  year      = {2025},
  volume    = {2025},
  url={https://openreview.net/forum?id=ALzTQUgW8a}
}

@inproceedings{zhang2023h2o,
  title     = {{H2O}: Heavy-Hitter Oracle for Efficient Generative Inference of Large Language Models},
  author    = {Zhang, Zhenyu and Sheng, Ying and Zhou, Tianyi and Chen, Tianlong and Zheng, Lianmin and Cai, Ruisi and Song, Zhao and Tian, Yuandong and R\'{e}, Christopher and Barrett, Clark and Wang, Zhangyang "Atlas" and Chen, Beidi},
  booktitle = {Advances in Neural Information Processing Systems},
  pages     = {34661--34710}, 
  year      = {2023},
  volume    = {36},
  url       = {https://proceedings.neurips.cc/paper_files/paper/2023/file/6ceefa7b15572587b78ecfcebb2827f8-Paper-Conference.pdf}
}

@inproceedings{xiao2024streamingllm,
  title     = {Efficient Streaming Language Models with Attention Sinks},
  author    = {Xiao, Guangxuan and Tian, Yuandong and Chen, Beidi and Han, Song and Lewis, Mike},
  booktitle = {International Conference on Learning Representations},
  pages     = {21875--21895},
  year      = {2024},
  volume    = {2024},
  url = {https://proceedings.iclr.cc/paper_files/paper/2024/file/5e5fd18f863cbe6d8ae392a93fd271c9-Paper-Conference.pdf}
}

@article{hu2026centroidkv,
  title   = {Centroid{KV}: Efficient Long-Context {LLM} Inference via {KV} Cache Clustering},
  author  = {Jie Hu and Shengnan Wang and Yutong He and Ping Gong and Jiawei Yi and Juncheng Zhang and Youhui Bai and Renhai Chen and Gong Zhang and Cheng Li and Kun Yuan},
  journal = {Transactions on Machine Learning Research},
  issn    = {2835-8856},
  year    = {2026},
  url     = {https://openreview.net/forum?id=T3EeupQhGj}
}

@inproceedings{hao2025omnikv,
  title = {{OmniKV}: Dynamic Context Selection for Efficient Long-Context {LLM}s},
  author = {Hao, Jitai and Zhu, Yuke and Wang, Tian and Yu, Jun and Xin, Xin and Zheng, Bo and Ren, Zhaochun and Guo, Sheng},
  booktitle = {International Conference on Learning Representations},
  pages = {87443--87464},
  year = {2025},
  volume = {2025},
  url = {https://proceedings.iclr.cc/paper_files/paper/2025/file/da1131a86ac3c70e0b7cae89c3d4df22-Paper-Conference.pdf}
}

@inproceedings{sun2025shadowkv,
  title = 	 {{S}hadow{KV}: {KV} Cache in Shadows for High-Throughput Long-Context {LLM} Inference},
  author =       {Sun, Hanshi and Chang, Li-Wen and Bao, Wenlei and Zheng, Size and Zheng, Ningxin and Liu, Xin and Dong, Harry and Chi, Yuejie and Chen, Beidi},
  booktitle = 	 {Proceedings of the 42nd International Conference on Machine Learning},
  pages = 	 {57355--57373},
  year = 	 {2025},
  editor = 	 {Singh, Aarti and Fazel, Maryam and Hsu, Daniel and Lacoste-Julien, Simon and Berkenkamp, Felix and Maharaj, Tegan and Wagstaff, Kiri and Zhu, Jerry},
  volume = 	 {267},
  series = 	 {Proceedings of Machine Learning Research},
  month = 	 {13--19 Jul},
  publisher =    {PMLR},
  url = 	 {https://proceedings.mlr.press/v267/sun25b.html}
}

@inproceedings{liu2025clusterkv,
  title={Clusterkv: Manipulating llm kv cache in semantic space for recallable compression},
  author={Liu, Guangda and Li, Chengwei and Zhao, Jieru and Zhang, Chenqi and Guo, Minyi},
  booktitle={2025 62nd ACM/IEEE Design Automation Conference (DAC)},
  pages={1--7},
  year={2025},
  organization={IEEE}
}

@inproceedings{ainslie2023gqa,
  title     = "{GQA}: Training Generalized Multi-Query Transformer Models from Multi-Head Checkpoints",
  author    = "Ainslie, Joshua  and Lee-Thorp, James  and de Jong, Michiel  and Zemlyanskiy, Yury  and Lebron, Federico  and Sanghai, Sumit",
  booktitle = "Proceedings of the 2023 Conference on Empirical Methods in Natural Language Processing",
  publisher = "Association for Computational Linguistics",
  year      = "2023",
  month     = dec,
  pages     = "4895--4901",
  url       = "https://aclanthology.org/2023.emnlp-main.298/"
}

@article{su2024rope,
  title     = {{RoFormer}: Enhanced transformer with Rotary Position Embedding},
  author    = {Jianlin Su and Murtadha Ahmed and Yu Lu and Shengfeng Pan and Wen Bo and Yunfeng Liu},
  journal   = {Neurocomputing},
  volume    = {568},
  pages     = {127063},
  year      = {2024},
  issn      = {0925-2312},
  url       = {https://www.sciencedirect.com/science/article/pii/S0925231223011864}
}

@misc{deepseekai2024v2,
  title         = {{DeepSeek-V2}: A Strong, Economical, and Efficient Mixture-of-Experts Language Model},
  author        = {{DeepSeek-AI} and others},
  year          = {2024},
  url           = {https://arxiv.org/abs/2405.04434},
  eprint        = {2405.04434},
  archivePrefix = {arXiv},
  primaryClass  = {cs.CL}
}

@misc{qwen2024qwen25,
  title         = {{Qwen2.5} Technical Report},
  author        = {{Qwen Team}},
  year          = {2024},
  url           = {https://arxiv.org/abs/2412.15115},
  eprint        = {2412.15115},
  archivePrefix = {arXiv},
  primaryClass  = {cs.CL}
}

@inproceedings{hsieh2024ruler,
title={{RULER}: What{\textquoteright}s the Real Context Size of Your Long-Context Language Models?},
author={Cheng-Ping Hsieh and Simeng Sun and Samuel Kriman and Shantanu Acharya and Dima Rekesh and Fei Jia and Boris Ginsburg},
booktitle={First Conference on Language Modeling},
year={2024},
url={https://openreview.net/forum?id=kIoBbc76Sy}
}

@inproceedings{yen2025helmet,
  title     = {{HELMET}: How to Evaluate Long-context Models Effectively and Thoroughly},
  author    = {Yen, Howard and Gao, Tianyu and Hou, Minmin and Ding, Ke and Fleischer, Daniel and Izsak, Peter and Wasserblat, Moshe and Chen, Danqi},
  booktitle = {International Conference on Learning Representations},
  pages     = {98914--98965},
  year      = {2025},
  volume    = {2025},
  url       = {https://proceedings.iclr.cc/paper_files/paper/2025/file/f5332c8273d02729730a9c24dec2135e-Paper-Conference.pdf}
}

@inproceedings{bai2025longbenchv2,
  title     = "{L}ong{B}ench v2: Towards Deeper Understanding and Reasoning on Realistic Long-context Multitasks",
  author    = "Bai, Yushi  and Tu, Shangqing  and Zhang, Jiajie  and Peng, Hao  and Wang, Xiaozhi  and Lv, Xin  and Cao, Shulin  and Xu, Jiazheng  and Hou, Lei  and Dong, Yuxiao  and Tang, Jie  and Li, Juanzi",
  booktitle = "Proceedings of the 63rd Annual Meeting of the Association for Computational Linguistics (Volume 1: Long Papers)",
  pages     = "3639--3664",
  year      = "2025",
  month     = jul,
  publisher = "Association for Computational Linguistics",
  url       = "https://aclanthology.org/2025.acl-long.183/",
}

@inproceedings{lee2026santa,
  title     = {Stochastic Sparse Attention for Memory-Bound Inference},
  author    = {Kyle Lee and Corentin Delacour and Kevin Callahan-Coray and Kyle Jiang and Can Yaras and Samet Oymak and Tathagata Srimani and Kerem Yunus Camsari},
  booktitle = {Forty-third International Conference on Machine Learning},
  year      = {2026},
  url       = {https://openreview.net/forum?id=ptaIjzIuoY}
}

@inproceedings{hooper2025squeezed,
    title = "Squeezed Attention: Accelerating Long Context Length {LLM} Inference",
    author = "Hooper, Coleman Richard Charles  and
      Kim, Sehoon  and
      Mohammadzadeh, Hiva  and
      Maheswaran, Monishwaran  and
      Zhao, Sebastian  and
      Paik, June  and
      Mahoney, Michael W.  and
      Keutzer, Kurt  and
      Gholami, Amir",
    editor = "Che, Wanxiang  and
      Nabende, Joyce  and
      Shutova, Ekaterina  and
      Pilehvar, Mohammad Taher",
    booktitle = "Proceedings of the 63rd Annual Meeting of the Association for Computational Linguistics (Volume 1: Long Papers)",
    month = jul,
    year = "2025",
    address = "Vienna, Austria",
    publisher = "Association for Computational Linguistics",
    url = "https://aclanthology.org/2025.acl-long.1568/",
    doi = "10.18653/v1/2025.acl-long.1568",
    pages = "32631--32652",
    ISBN = "979-8-89176-251-0"
}

@inproceedings{hooper2026multipole,
 author = {Hooper, Coleman and Zhao, Sebastian and Manolache, Luca and Kim, Sehoon and Mahoney, Michael and Shao, Sophia and Keutzer, Kurt and Gholami, Amir},
 booktitle = {Advances in Neural Information Processing Systems},
 doi = {10.52202/085713-1255},
 editor = {D. Belgrave and C. Zhang and H. Lin and R. Pascanu and P. Koniusz and M. Ghassemi and N. Chen},
 pages = {37382--37405},
 publisher = {Curran Associates, Inc.},
 title = {Multipole Attention for Efficient Long Context Reasoning},
 url = {https://proceedings.neurips.cc/paper_files/paper/2025/file/359f442022ef31bcd9e41959576107ef-Paper-Conference.pdf},
 volume = {38, Main Conference},
 year = {2025}
}

@inproceedings{
liu2024kivi,
title={{KIVI}: A Tuning-Free Asymmetric 2bit Quantization for {KV} Cache},
author={Zirui Liu and Jiayi Yuan and Hongye Jin and Shaochen Zhong and Zhaozhuo Xu and Vladimir Braverman and Beidi Chen and Xia Hu},
booktitle={Forty-first International Conference on Machine Learning},
year={2024},
url={https://openreview.net/forum?id=L057s2Rq8O}
}

@inproceedings{hooper2024kvquant,
title={{KVQ}uant: Towards 10 Million Context Length {LLM} Inference with {KV} Cache Quantization},
author={Coleman Richard Charles Hooper and Sehoon Kim and Hiva Mohammadzadeh and Michael W. Mahoney and Sophia Shao and Kurt Keutzer and Amir Gholami},
booktitle={The Thirty-eighth Annual Conference on Neural Information Processing Systems},
year={2024},
url={https://openreview.net/forum?id=0LXotew9Du}
}

@article{child2019generating,
  title={Generating long sequences with sparse transformers},
  author={Child, Rewon and Gray, Scott and Radford, Alec and Sutskever, Ilya},
  journal={arXiv preprint arXiv:1904.10509},
  year={2019}
}

@inproceedings{zaheer2020bigbird,
 author = {Zaheer, Manzil and Guruganesh, Guru and Dubey, Kumar Avinava and Ainslie, Joshua and Alberti, Chris and Ontanon, Santiago and Pham, Philip and Ravula, Anirudh and Wang, Qifan and Yang, Li and Ahmed, Amr},
 booktitle = {Advances in Neural Information Processing Systems},
 editor = {H. Larochelle and M. Ranzato and R. Hadsell and M.F. Balcan and H. Lin},
 pages = {17283--17297},
 publisher = {Curran Associates, Inc.},
 title = {Big Bird: Transformers for Longer Sequences},
 url = {https://proceedings.neurips.cc/paper_files/paper/2020/file/c8512d142a2d849725f31a9a7a361ab9-Paper.pdf},
 volume = {33},
 year = {2020}
}

@article{beltagy2020longformer,
  title={Longformer: The long-document transformer},
  author={Beltagy, Iz and Peters, Matthew E and Cohan, Arman},
  journal={arXiv preprint arXiv:2004.05150},
  year={2020}
}

@article{wang2020linformer,
  title={Linformer: Self-attention with linear complexity},
  author={Wang, Sinong and Li, Belinda Z and Khabsa, Madian and Fang, Han and Ma, Hao},
  journal={arXiv preprint arXiv:2006.04768},
  year={2020}
}

@inproceedings{kitaev2020reformer,
title={Reformer: The Efficient Transformer},
author={Nikita Kitaev and Lukasz Kaiser and Anselm Levskaya},
booktitle={International Conference on Learning Representations},
year={2020},
url={https://openreview.net/forum?id=rkgNKkHtvB}
}

@inproceedings{frantar2023optq,
title={{OPTQ}: Accurate Quantization for Generative Pre-trained Transformers},
author={Elias Frantar and Saleh Ashkboos and Torsten Hoefler and Dan Alistarh},
booktitle={The Eleventh International Conference on Learning Representations },
year={2023},
url={https://openreview.net/forum?id=tcbBPnfwxS}
}

@inproceedings{xiao2025duoattention,
title={DuoAttention: Efficient Long-Context {LLM} Inference with Retrieval and Streaming Heads},
author={Guangxuan Xiao and Jiaming Tang and Jingwei Zuo and Junxian Guo and Shang Yang and Haotian Tang and Yao Fu and Song Han},
booktitle={The Thirteenth International Conference on Learning Representations},
year={2025},
url={https://openreview.net/forum?id=cFu7ze7xUm}
}

@InProceedings{ribar2024sparq,
  title = 	 {{S}par{Q} Attention: Bandwidth-Efficient {LLM} Inference},
  author =       {Ribar, Luka and Chelombiev, Ivan and Hudlass-Galley, Luke and Blake, Charlie and Luschi, Carlo and Orr, Douglas},
  booktitle = 	 {Proceedings of the 41st International Conference on Machine Learning},
  pages = 	 {42558--42583},
  year = 	 {2024},
  editor = 	 {Salakhutdinov, Ruslan and Kolter, Zico and Heller, Katherine and Weller, Adrian and Oliver, Nuria and Scarlett, Jonathan and Berkenkamp, Felix},
  volume = 	 {235},
  series = 	 {Proceedings of Machine Learning Research},
  month = 	 {21--27 Jul},
  publisher =    {PMLR},
  url = 	 {https://proceedings.mlr.press/v235/ribar24a.html}
}

@inproceedings{kim2022fast,
  title={Fast monte-carlo approximation of the attention mechanism},
  author={Kim, Hyunjun and Ko, JeongGil},
  booktitle={Proceedings of the AAAI conference on artificial intelligence},
  volume={36},
  number={7},
  pages={7185--7193},
  year={2022}
}

@inproceedings{gupta2021topk,
    title = "Memory-efficient Transformers via Top-k Attention",
    author = "Gupta, Ankit  and
      Dar, Guy  and
      Goodman, Shaya  and
      Ciprut, David  and
      Berant, Jonathan",
    editor = "Moosavi, Nafise Sadat  and
      Gurevych, Iryna  and
      Fan, Angela  and
      Wolf, Thomas  and
      Hou, Yufang  and
      Marasovi{\'c}, Ana  and
      Ravi, Sujith",
    booktitle = "Proceedings of the Second Workshop on Simple and Efficient Natural Language Processing",
    month = nov,
    year = "2021",
    address = "Virtual",
    publisher = "Association for Computational Linguistics",
    url = "https://aclanthology.org/2021.sustainlp-1.5/",
    doi = "10.18653/v1/2021.sustainlp-1.5",
    pages = "39--52"
}
